\documentclass[letterpaper, 10 pt, conference]{ieeeconf}  

\IEEEoverridecommandlockouts                              

\usepackage{hyperref}
\usepackage{multirow}
\usepackage{romannum}
\usepackage{graphicx}
\usepackage{amsmath}
\usepackage{amssymb}
\usepackage{breqn}
\usepackage{subcaption}
\usepackage{balance}

\usepackage{pgfplots}
\usepackage{booktabs, dcolumn}

\pgfplotsset{compat=newest}
\pgfplotsset{plot coordinates/math parser=false}
\newlength\figureheight
\newlength\figurewidth

\usepackage[font={small}]{caption}
\usepackage{pifont} 
\newcommand{\cmark}{\ding{51}}%
\newcommand{\xmark}{\ding{55}}%
\newcommand{\tabincell}[2]{\begin{tabular}{@{}#1@{}}#2\end{tabular}}  
\usepackage{soul}

\makeatletter
\let\NAT@parse\undefined
\makeatother
\usepackage[numbers,sort,compress]{natbib}
\usepackage{hyperref}
\usepackage{xcolor,colortbl}

\title{\bf
Stereo Matching by Self-supervision of Multiscopic Vision
}

\author{Weihao Yuan, Yazhan Zhang, Bingkun Wu, Siyu Zhu, Ping Tan, Michael Yu Wang, and Qifeng Chen 
\thanks{W. Yuan (\href{mailto:qianmu.ywh@alibaba-inc.com}{qianmu.ywh@alibaba-inc.com}), S. Zhu and P. Tan are with AI Lab, Alibaba Cloud, China.
Y. Zhang, M. Wang and Q. Chen (\href{mailto:cqf@ust.hk}{cqf@ust.hk}) are with the Dept. MAE and Dept. CSE, Hong Kong University of Science and Technology, Hong Kong SAR, China. B. Wu is with the Beihang University, China.}
}

\begin{document}

\maketitle



\begin{abstract}

Self-supervised learning for depth estimation possesses several advantages over supervised learning. The benefits of no need for ground-truth depth, online fine-tuning, and better generalization with unlimited data attract researchers to seek self-supervised solutions. In this work, we propose a new self-supervised framework for stereo matching utilizing multiple images captured at aligned camera positions. A cross photometric loss, an uncertainty-aware mutual-supervision loss, and a new smoothness loss are introduced to optimize the network in learning disparity maps end-to-end without ground-truth depth information. To train this framework, we build a new multiscopic dataset consisting of synthetic images rendered by 3D engines and real images captured by real cameras. After being trained with only the synthetic images, our network can perform well in unseen outdoor scenes. Our experiment shows that our model obtains better disparity maps than previous unsupervised methods on the KITTI dataset and is comparable to supervised methods when generalized to unseen data. Our source code and dataset are available at \url{https://sites.google.com/view/multiscopic}.

\end{abstract}


\section{Introduction}

Accurate depth estimation is essential for numerous applications such as robotic manipulation, autonomous driving, and 3D reconstruction~\cite{geiger2012we, yao2018mvsnet, yuan2019reinforcement, yuan2019end}. Various industrial solutions have employed stereo cameras for dense depth sensing. A vital benefit of a stereo camera is that the two cameras are axis-aligned so that the corresponding pixels between two images can lie on the same row. It appears that adding more axis-aligned cameras can help obtain high-quality depth maps. In this work, we study a novel problem for self-supervised stereo matching: given axis-aligned multiscopic images, can we train a stereo matching model for better depth sensing?

Recent deep-learning-based methods for depth estimation have dominated most benchmarks~\cite{kendall2017end,chang2018pyramid,zhang2019ga, Xie2020}. However, these learning-based methods do not perform stably on unseen scenarios since the methods rely heavily on the training data. The ground-truth depth is hard to collect in the real world so that the size of the training data with ground-truth depth is usually small~\cite{geiger2012we}. In this case, the network trained with limited data is prone to overfitting the training data rather than understanding the underlying structure of the task. Thus, most deep-learning-based approaches do not have stable generalization ability, which restricts their applications in the industry.

To this end, researchers are exploring unsupervised learning for depth estimation~\cite{godard2017unsupervised,wang2019unos}. Compared to supervised approaches, unsupervised learning does not need the ground-truth depth, so it can be trained with a large amount of data. Also, since there is no ground-truth depth, the network needs to find the underlying pattern and geometry structure of images for stereo matching rather than simply overfitting the real depth value. Another advantage of unsupervised learning is that online fine-tuning becomes plausible~\cite{tonioni2019learning,tonioni2019real,zhong2017self}. As the ground truth is not needed for training a model, the model can be updated when processing the new inputs, which can lead to various practical applications.

The idea of exploiting more input data to assist the learning has been recently explored by some researchers~\cite{godard2017unsupervised,wang2019unos,liu2020flow2stereo}. For example, a model for unsupervised monocular depth estimation can be trained with stereo images~\cite{godard2017unsupervised}. Inspired by this idea, we wonder if the unsupervised stereo matching can be improved with three or more images during training.


\begin{figure}[]
\centering
  \includegraphics[width=0.7\columnwidth, trim={0cm 0cm 0cm 0cm}, clip]{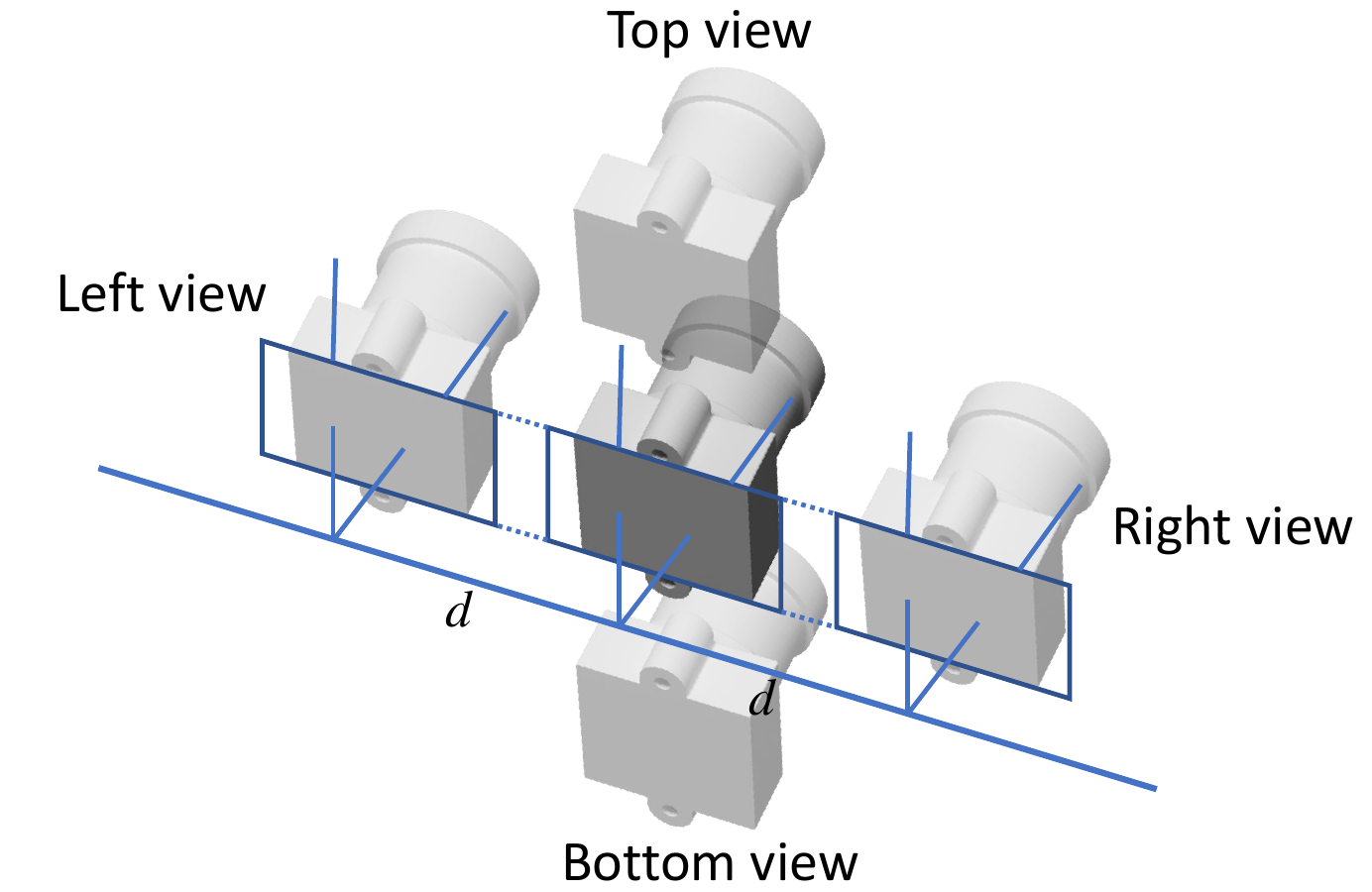}
\caption{Our multiscopic vision structure. Images are taken at parallel, co-planar, and same-parallax camera positions, such that the pixel disparities between adjacent images are the same.}
\label{fig:camera}
\vspace{-0.5cm}
\end{figure}

\begin{figure*}[t]
\centering
  \includegraphics[width=1.8\columnwidth, trim={0cm 0cm 0cm 0cm}, clip]{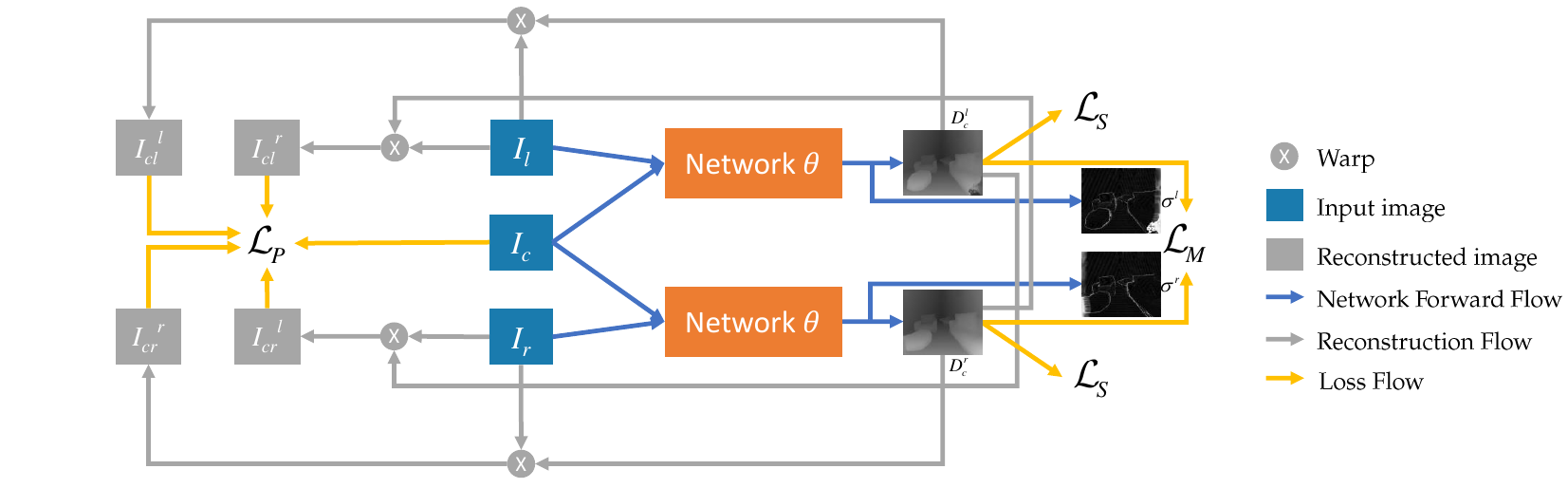}
\caption{Self-supervised learning framework for three-image multiscopic structure. Two networks sharing the same parameters compute the disparity $D_c^l,D_c^r$ and uncertainty map $\sigma^l,\sigma^r$ end-to-end, which are then used to cross warp the left and right images $I_l,I_r$ back to the center view, generating four reconstructed images $I_{cl}^l, I_{cl}^r, I_{cr}^l, I_{cr}^r$. The cross photometric loss $\mathcal{L}_P$ is built between them and the real center image $I_c$, the uncertainty-aware mutual-supervision loss $\mathcal{L}_M$ is built between $D_c^l$ and $D_c^r$ conditioned on $\sigma^l,\sigma^r$, and the smoothness loss $\mathcal{L}_S$ is built for each disparity map.}
\label{fig:framework}
\vspace{-0.5cm}
\end{figure*}

In this paper, we explore self-supervised learning depth estimation utilizing multiple images taken at aligned camera positions as input during training. We assume that multiple images are captured at horizontally or vertically aligned positions. As the stereo structure brings some benefits to depth estimation, the multiscopic structure introduces more constraints that can supervise the learning~\cite{yuan2020mfusenet}. These cameras are parallel, co-planar, and with the same baselines, as shown in Fig.~\ref{fig:camera}, such that the pixel disparities between adjacent images are the same. In this case, the disparity maps from the center view to any surrounding view should be exactly the same. Furthermore, for the left and right views, the pixel shift exists only in the horizontal axis, and for the bottom and top view, the pixel shift exists only in the vertical axis.

By exploiting these constraints, we can train a network to predict a better disparity map end-to-end without ground-truth depth information, as shown in Fig.~\ref{fig:framework}. Uncertainty-aware mutual supervision between different disparity maps serves as the self-supervising signal to optimize the network. Since the learning is self-supervised, it needs to find the underlying relationship of the multiscopic images to compute the disparity. Our experiment shows that the learned network can be generalized to unseen scenarios well, as there is no chance to overfit the ground-truth depth. After being trained with only synthetic images, our model performs well in real-world scenarios.

To make our unsupervised learning framework work and to promote more works in multiscopic vision, we propose a new dataset with 1300 scenes of multiscopic images as there is no available large dataset for multiscopic vision. There are a large number of synthetic images rendered by 3D engines~\cite{McCormac:etal:ICCV2017,Matterport3D} and a small number of real images taken by our multiscopic cameras. For the synthetic images, ground-truth depths are also provided.

Our main contributions concerning the self-supervised learning depth estimation framework and the multiscopic dataset are summarized as follows:

1) We exploit multiple images in a multiscopic structure for training a network to predict the disparity maps without the need for ground-truth depth. Optimization is built from the mutual supervision between confident areas of disparities obtained from different center-surround image pairs.

2) We build a new dataset of multiscopic images of 1200 scenes rendered in 3D engines and 100 scenes captured in the real world with our multiscopic cameras. Trained on only synthetic images, our model can be generalized to perform well on real-world data.

3) The experiment shows that our model outperforms other unsupervised approaches for stereo matching by a large margin and is comparable to supervised methods when generalized to unseen data.



\begin{figure*}[t]
\centering
\begin{subfigure}{0.65\columnwidth}
  \centering
  \includegraphics[height=4.5cm, trim={0cm 0cm 0cm 0cm}, clip]{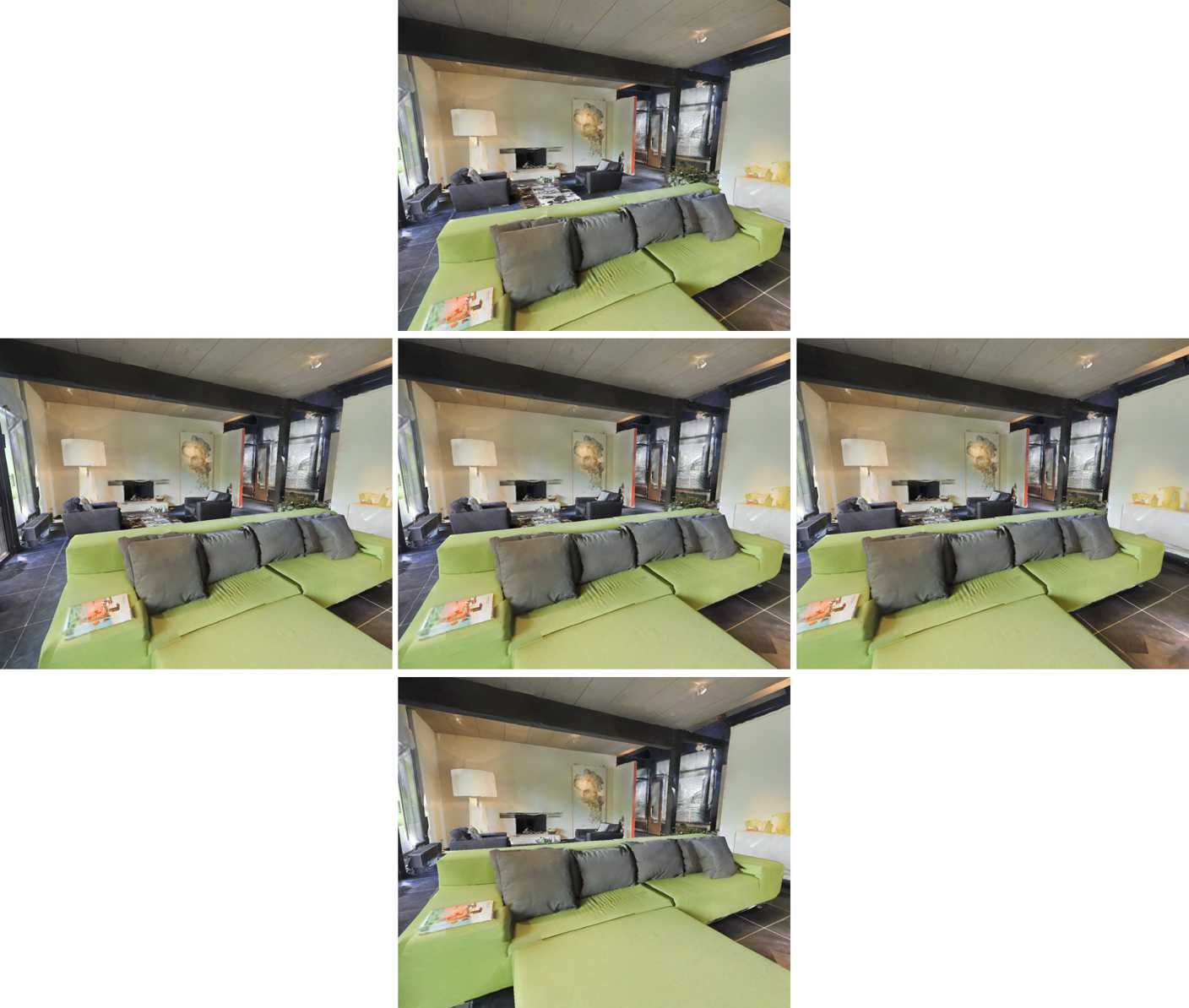}
  \caption{Color images}
\end{subfigure}
%
\begin{subfigure}{0.65\columnwidth}
  \centering
  \includegraphics[height=4.5cm, trim={0cm 0cm 0cm 0cm}, clip]{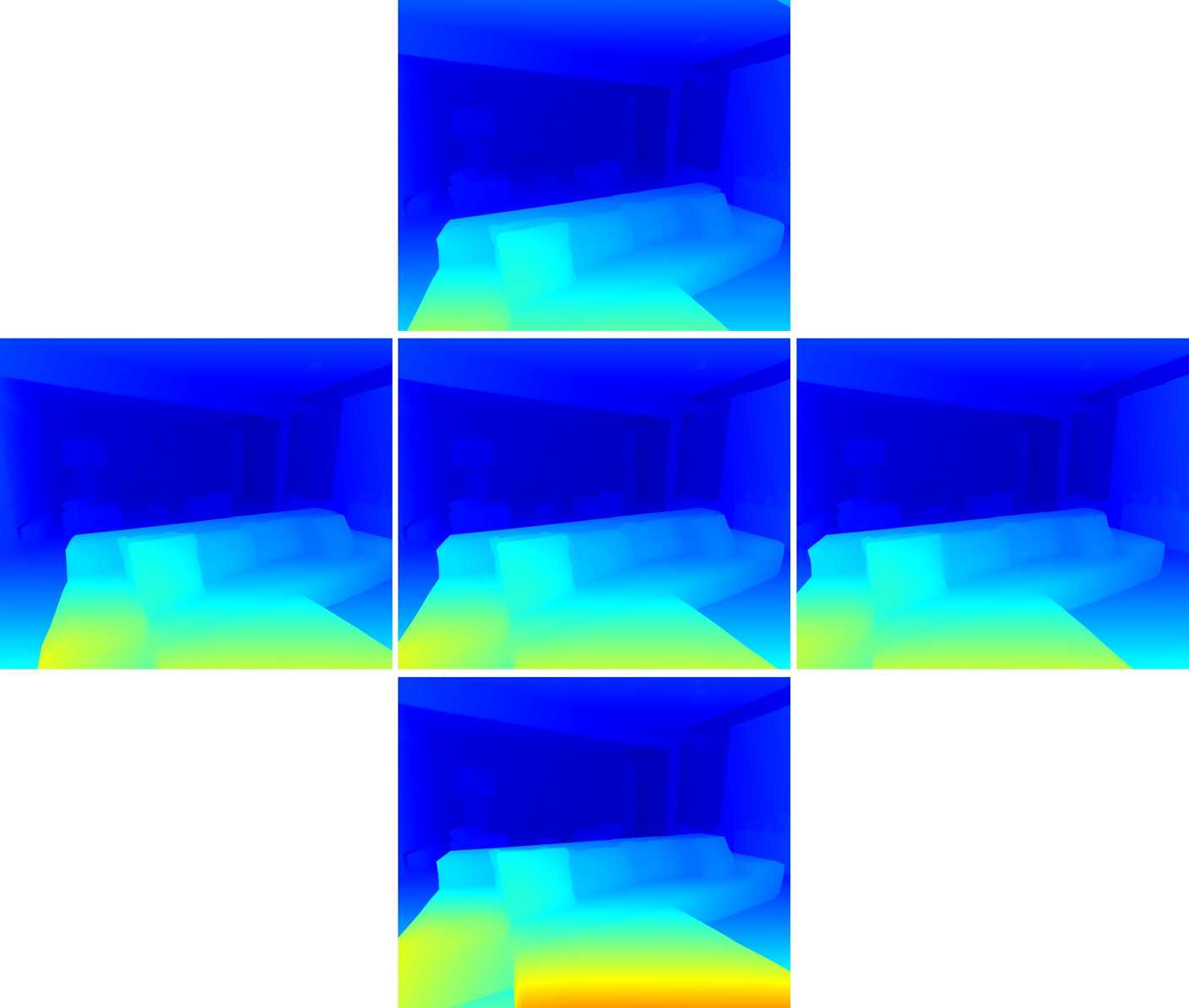}
  \caption{Ground-truth disparity maps}
\end{subfigure}
%
\begin{subfigure}{0.3\columnwidth}
  \centering
  \includegraphics[height=4.5cm, trim={0cm 0cm 0cm 0cm}, clip]{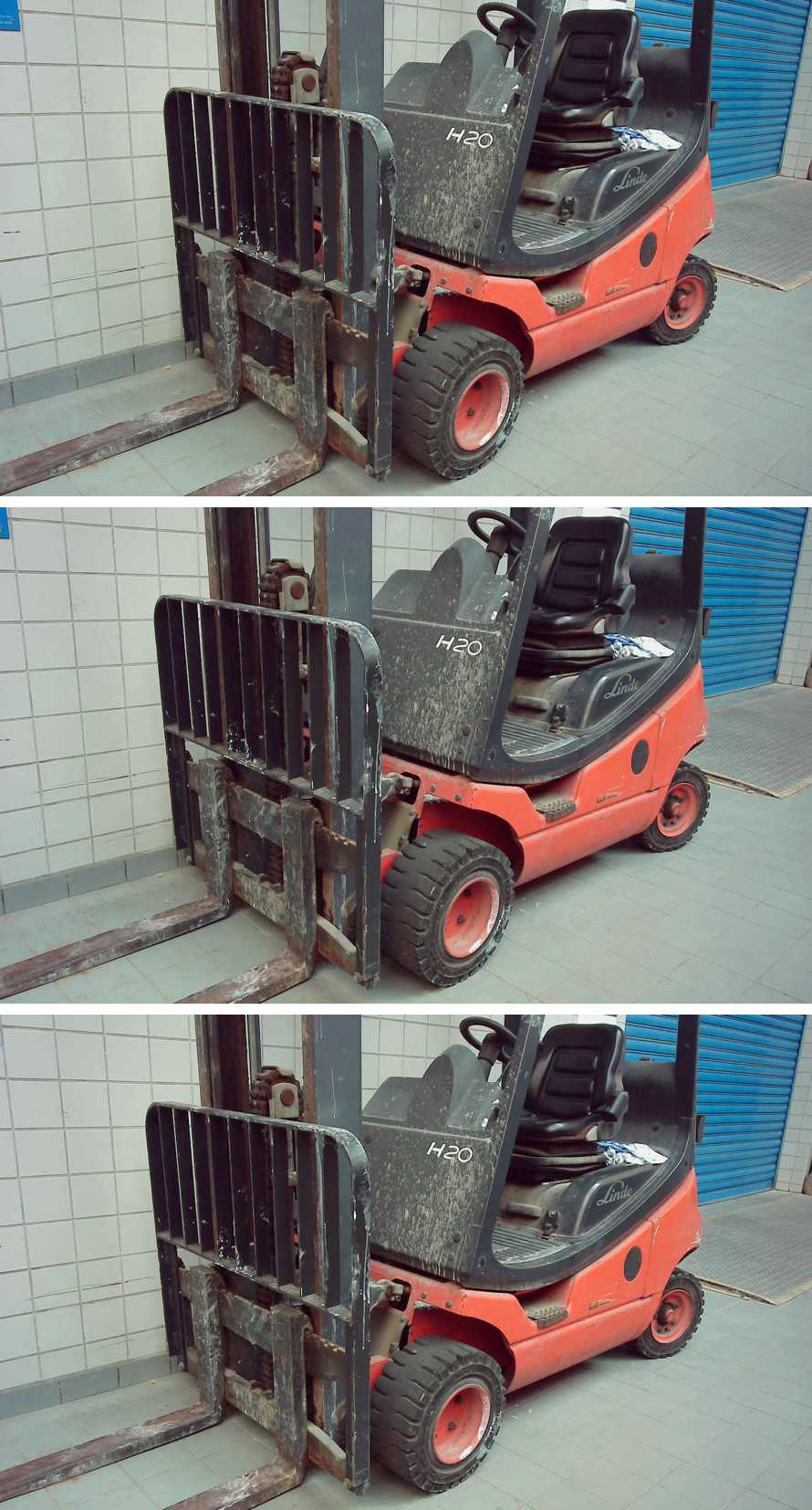}
  \caption{Real images}
\end{subfigure}
\caption{Example scenes of multiscopic images in our dataset.
The synthetic color images (a) and corresponding perfect disparity maps (b) displayed in Jet colormap are presented. (c) is one set of images obtained by the multiscopic system in the real world.}
\label{fig:dataset}
\vspace{-0.5cm}
\end{figure*}

\begin{figure}[t]
\centering
\begin{subfigure}{0.4\columnwidth}
  \centering
  \includegraphics[height=2cm, trim={0cm 0cm 0cm 0cm}, clip]{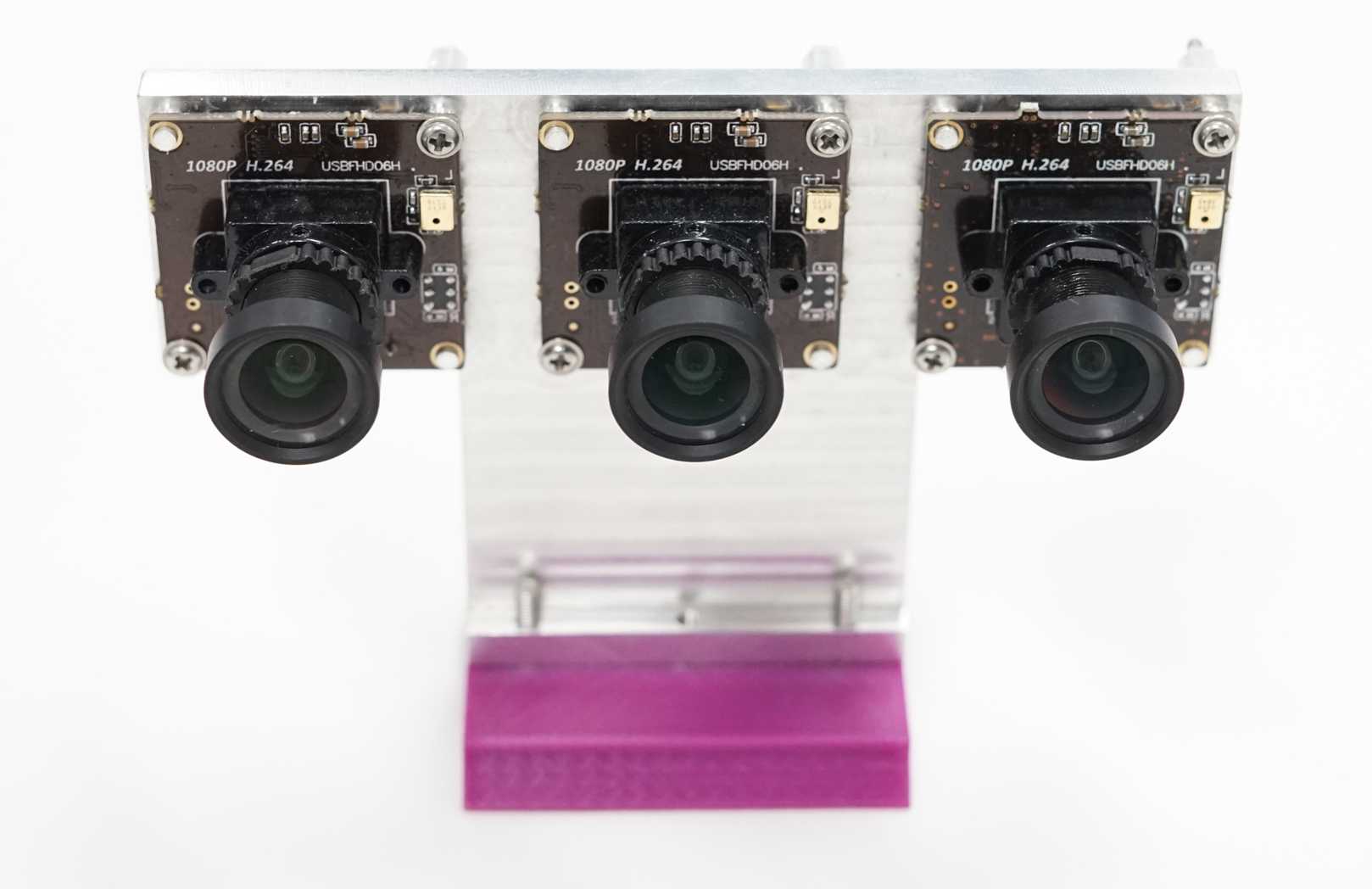}
  \caption{Multi-eye}
\end{subfigure}
\begin{subfigure}{0.5\columnwidth}
  \centering
  \includegraphics[height=2cm, trim={0cm 0cm 0cm 0cm}, clip]{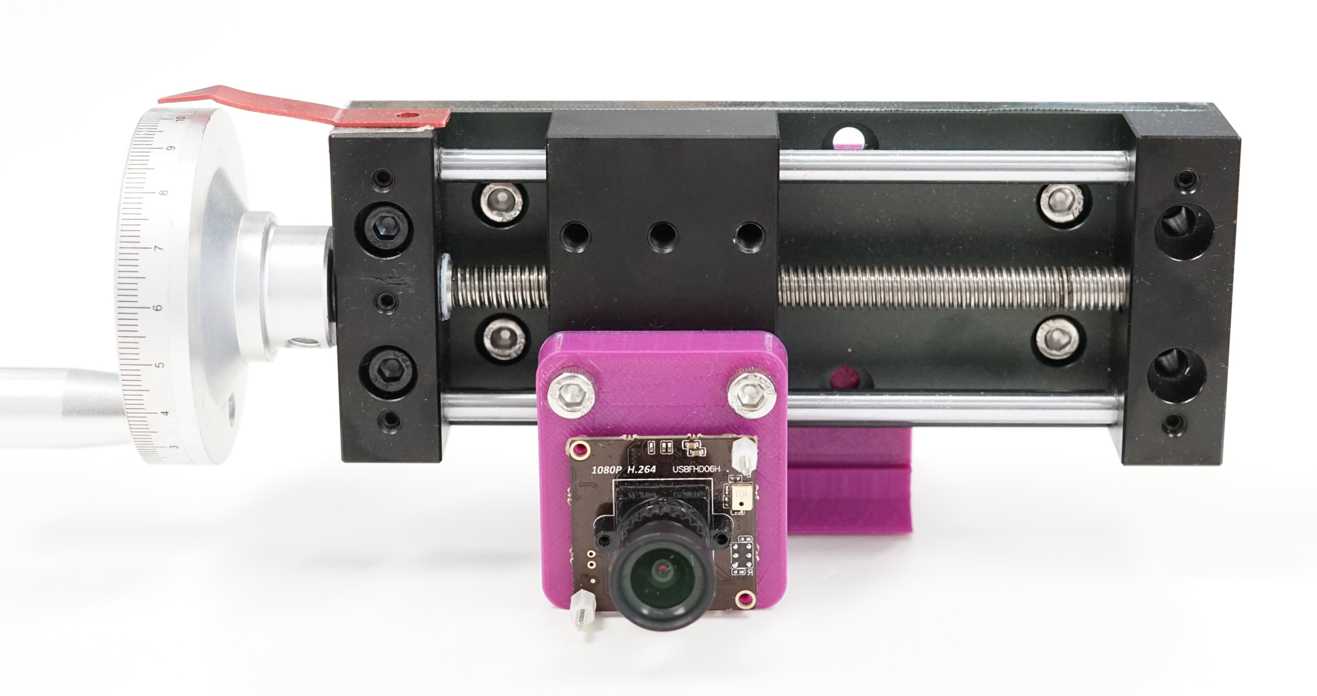}
  \caption{Moving-eye}
\end{subfigure}

\caption{Multiscopic systems designed to capture real multiscopic images. Multi-eye design (a) is using multiple identical cameras while moving-eye design (b) is using one camera sliding on a precision linear rail.}
\label{fig:realcam}
\vspace{-0.5cm}
\end{figure}

\section{Related Work}

\textbf{Multiscopic depth estimation.} All previous depth estimation works using multiscopic images are based on traditional matching methods \cite{kolmogorov2002multi,maitre2008symmetric,wei2005asymmetrical,lee2014multi}, since there is no available large dataset to train a network. Various methods take advantage of some constraints in multiscopic structure to improve the estimation results and only evaluate it on a small synthetic dataset, the Tsukuba images \cite{nakamura1996occlusion}.

\textbf{Deep depth estimation.} Deep learning based stereo methods explore using neural networks to replace one \cite{zbontar2016stereo} or all steps \cite{kendall2017end,chang2018pyramid,zhang2019ga} in traditional stereo matching to compute the disparity map. Exploiting the network to compute the matching cost of two image patches was first proposed in MC-CNN \cite{zbontar2016stereo}. To further take advantage of the neural networks, end-to-end approaches are proposed to remove all traditional matching steps by directly using the network \cite{kendall2017end,chang2018pyramid,zhang2019ga}.  Additionally, optimization methods in traditional stereo matching, such as semi-global matching, are also introduced to assist the deep methods in more meaningful ways \cite{zhang2019ga}. Heretofore, end-to-end learning approaches have dominated most benchmarks and show significant advantages. 

\textbf{Unsupervised depth estimation.} Despite the success of learning-based approaches in some specific datasets, they need plenty of ground-truth training data, which is expensive to acquire. To overcome this, more researchers are seeking assistance from unsupervised learning. Some unsupervised frameworks were first proposed for monocular depth estimation. The monocular depth network is supervised by the reconstruction loss between the stereo image pairs \cite{garg2016unsupervised,xie2016deep3d,guo2018learning}. The reconstructed images are synthesized by warping one view to the other view using the disparity maps. This is enhanced by introducing the consistency of disparity maps between the left view and the right view \cite{godard2017unsupervised}.

Instead of relying on the stereo images, some works combine the ego-motion and optical flow estimation to help the unsupervised depth learning from videos \cite{zhou2017unsupervised,yin2018geonet}. The loss is built by warping adjacent views with the computed depth, pose, and optical flow. Also, some geometric consistency losses are introduced to improve the estimation \cite{yin2018geonet}.

For stereo depth estimation, there are some losses similar to monocular approaches and more constraints that can further supervise the learning. Zhou et al.~\cite{zhou2017unsupervised} evaluate the matching result and pick the confident patch pairs as the new training data, in which way the network is iteratively optimized. Godard et al.~\cite{godard2017unsupervised} realize the stereo matching using the same reconstruction loss and left-right consistency loss as the monocular depth task. This is further improved by introducing occlusion-aware loss \cite{luo2018unsupervised} or heuristic loss \cite{zhong2017self}. Then more works try to explore supervision from the consistency in stereo videos \cite{zhong2018open, wang2019unos, liu2020flow2stereo, kim2020loop}, by utilizing the consistency in stereo depth, optical flow, and ego-motion. Some other works also introduce parallax-attention \cite{wang2020parallax} or seek the assistance from other networks \cite{aleotti2020reversing}.

So far, a limited number of works have been dedicated to unsupervised deep depth estimation in the multiscopic structure that provides more constraints to assist unsupervised learning \cite{yuan2020mfusenet,Zhang2020}. Similar to supervising the monocular depth estimation with stereo images, we show that we can supervise the stereo matching with multiscopic images to obtain a better disparity map by exploiting the inner shape of the multiscopic structure. To the best of our knowledge, our work is the first deep stereo depth estimation method utilizing multiscopic images.


\section{Multiscopic Dataset}

In this section, we introduce the details of our multiscopic vision system and describe how we build our multiscopic dataset in virtual environments and the real world.

In a multiscopic vision structure, multiple images are captured at axis-aligned camera positions, such that these images are taken with parallel optical axes on the same image plane, and with the same parallaxes, as shown in Fig.~\ref{fig:camera}. Multiscopic vision structure brings clear benefits to depth estimation~\cite{yuan2020active}. The disparity of a pixel from the center view to any surrounding view should be the same, which is a strong constraint to optimize the result. Additionally, the occluded regions in one view can be perceived from other views, in which case there are no occlusion problems for the entire multiscopic system \cite{wei2005asymmetrical}.

Some interesting research can be conducted in this setup. For instance, multiscopic images can be utilized for better image enhancement since there are more constraints to be enforced, such as super-resolution with multiple images \cite{li2010multi}. 
Also, novel view synthesis \cite{zhou2018stereo} can be done in a multiscopic structure, such as predicting the right view provided with only the left and center views.

Since there is no available large dataset for multiscopic vision, we build a new dataset composed of 1200 scenes of synthetic multiscopic data rendered by 3D engines and 100 scenes of real images taken with real-world cameras.

\subsection{Synthetic Data}

The synthetic images are rendered using the 3D render engine Habitat-sim \cite{savva2019habitat} and SceneNet RGB-D \cite{McCormac:etal:ICCV2017}, with 3D models from Matterport3D \cite{Matterport3D}, Gibson \cite{xiazamirhe2018gibsonenv}, and ShapeNet \cite{chang2015shapenet}. Habitat-sim is used to render Matterport3D as well as Gibson, and SceneNet RGB-D is used to render ShapeNet. The models from Matterport3D and Gibson are collected from real indoor spaces by reconstruction from data acquired using real-world RGB cameras, depth cameras, and 3D scanning sensors. Most models within them are ordinary houses with daily-life objects like furniture, decorations, and instruments. The models from ShapeNet are some virtual indoor space models in which we can randomly place all kinds of 3D object models, like toys, plants, cars, or even airplanes. So when we render the scenes using models from ShapeNet, we adjust the lighting condition randomly while we keep the light the same as the real world when we render the models from Matterport3D and Gibson.

There are more than 20K raw scenes of images. We remove the similar scenes and keep 1200 scenes in total that are unlike each other. This makes these 1200 scenes representative of visual variation.
For each scene in our dataset, there are five images taken with multiscopic vision structure, the center, left, right, bottom, and top view, as shown in Fig.~\ref{fig:dataset}. More scenes can be seen in the supplementary materials. The resolution of all images is $1280\times1080$, and every color image has its corresponding ground-truth disparity map. As a multiscopic structure, the baselines between the four surrounding images and the center image are the same and vary from 0.05 to 0.2 meters. The value range of the disparity maps is from 0 to 255.

\subsection{Real-world Data}

Real-world multiscopic data are taken with one or multiple RGB cameras. Multiscopic images require aligned camera locations, which is not easy to control. Therefore, to capture images in this structure, we design two architecture: 
The first architecture is composed of three identical cameras with Sony IMX322 inside, whose resolution is $1920\times1080$. These three cameras are aligned on a horizontal line, as displayed in Fig.~\ref{fig:realcam}(a). They are parallel, co-planar, and with the same baseline of 4 centimeters.
The second design is a moving-eye system. Although we utilize the same cameras, a multi-eye structure still suffers from the problem that the cameras' parameters are not exactly the same. To overcome this, we place one camera on a precision linear sliding rail and slide it horizontally on its image plane, as displayed in Fig.~\ref{fig:realcam}(b). In this system, three images are taken at three locations to obtain multiscopic images. The baselines of adjacent locations vary from 3 to 4 centimeters.

Images obtained by a moving-eye system have strictly identical parameters, but sliding a camera needs some time so that it cannot take images of dynamic scenes. Thus, we combine these two systems to build a real-world image dataset. There are 100 scenes of real-world images in our dataset. For every scene, there are the left, center, and right images, as shown in Fig.~\ref{fig:dataset}(c). More scenes are shown in the supplementary materials. Some images are cropped to reduce the vignette influence and remove messy environments, after which the resolutions vary from $1400\times1000$ to $1920\times1080$. For the real data, there is no ground-truth depth information.

\section{Self-supervised Multiscopic Matching}


\subsection{Learning Framework}
In rectified stereo matching, pixels shift along the horizontal axis such that the searching for pixels correspondence is simplified. The depth map can be constructed from a computed disparity map that establishes pixel correspondences between the left image $I_l$ and the right image $I_r$. With a disparity map, the right image can be warped back to the left image to produce a synthetic left image $I_l'$.
%
Ideally, if the disparity map $D_l$ is correctly estimated, the synthetic left image $I_l'$ should be the same as $I_l$ except for occluded regions. Then the discrepancy between them can be used as a photometric loss in training a self-supervised model \cite{godard2017unsupervised}.

\begin{figure}[t]
\centering
\begin{subfigure}{1\columnwidth}
  \centering
  \includegraphics[width=1\columnwidth, trim={0cm 0cm 0cm 0cm}, clip]{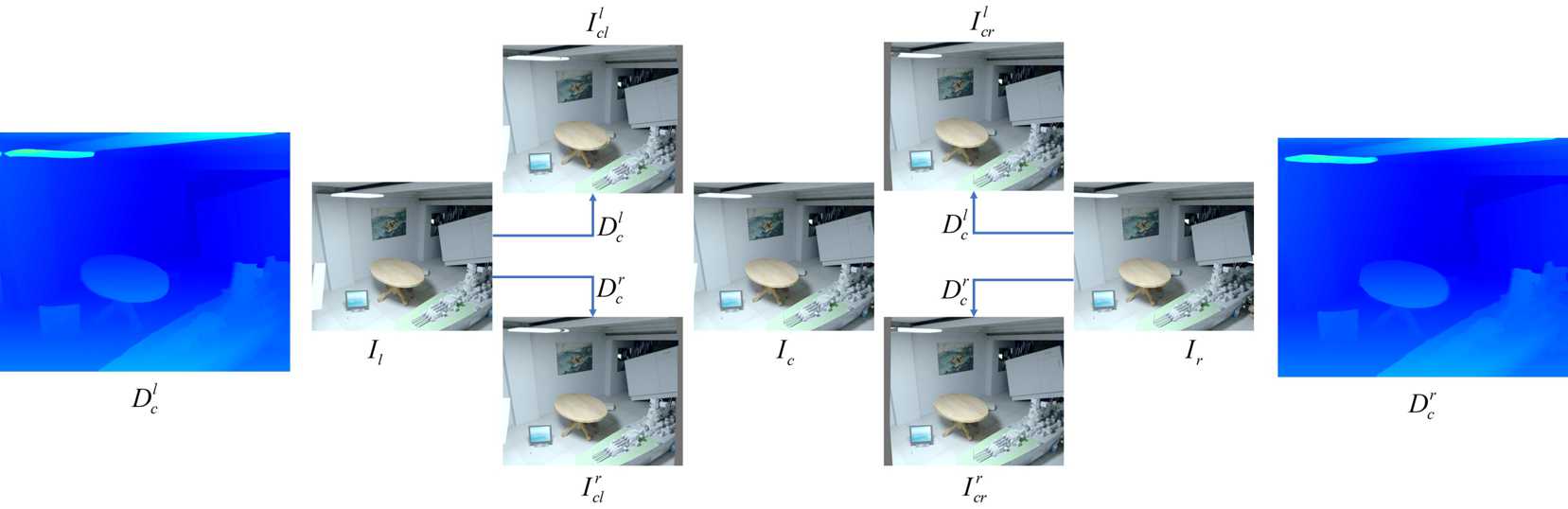}
\end{subfigure}
\caption{Cross warping between $I_l,I_c,I_r$ using disparity maps $D_c^l, D_c^r$. The cross photometric loss is then established between the warping results $I_{cl}^{l},I_{cl}^{r},I_{cr}^{r},I_{cr}^{l}$ and the center image $I_c$.}
\label{fig:warp}
\vspace{-0.5cm}
\end{figure}

In multiscopic vision, a similar photometric loss can be enforced between any two axis-aligned images. Since the baselines between adjacent camera positions are the same, more constraints could be established in a constrained way. As shown in Fig.~\ref{fig:framework}, we deal with a multiscopic system with three images without loss of generality. Benefiting from the multiscopic architecture, the disparities between the center image and any surrounding image should be the same, with the center image as the reference. Therefore, one disparity map can be used to warp all surrounding images back to the center image to work as a cross photometric loss. The disparity $D_c^r$ between the center image $I_c$ and the right image $I_r$ can be used to not only warp $I_r$ to produce a synthetic center image $I_{cr}^{r}$, but also warp $I_l$ to produce $I_{cl}^{r}$, as shown in Fig.~\ref{fig:warp}. On the one hand, this cross warping consistency can strengthen the photometric supervision. On the other hand, it can address the occlusion problem which troubles stereo vision, e.g., the invisible pixel in the right view would cause a large loss between $I_{cl}^{r}$ and $I_c$.

In a multiscopic structure, another constraint is that the disparity maps between the center image and all surrounding images should be the same. The discrepancy in disparity values indicates inaccurate disparity estimation. Thus the difference between multiple disparities could work as mutual supervision, as shown in Fig.~\ref{fig:framework}. This also helps handle the occlusion problem, which is usually problematic in stereo matching. The occluded regions of the center image in one surrounding image can be seen in other images. For example, the occluded areas between the left image and the center image are hard to be estimated correctly in $D_c^l$, but these parts are estimated well in $D_c^r$. In this case, we estimate an uncertainty map for each disparity map. In the uncertain areas like occluded pixels in $D_c^l$, we take the confident pixels in $D_c^r$ as supervision labels and optimize the estimation network for $D_c^l$.

\begin{figure*}[t]
\centering
\begin{subfigure}{0.4\columnwidth}
  \centering
  \includegraphics[width=1\columnwidth, trim={0cm 0cm 0cm 1cm}, clip]{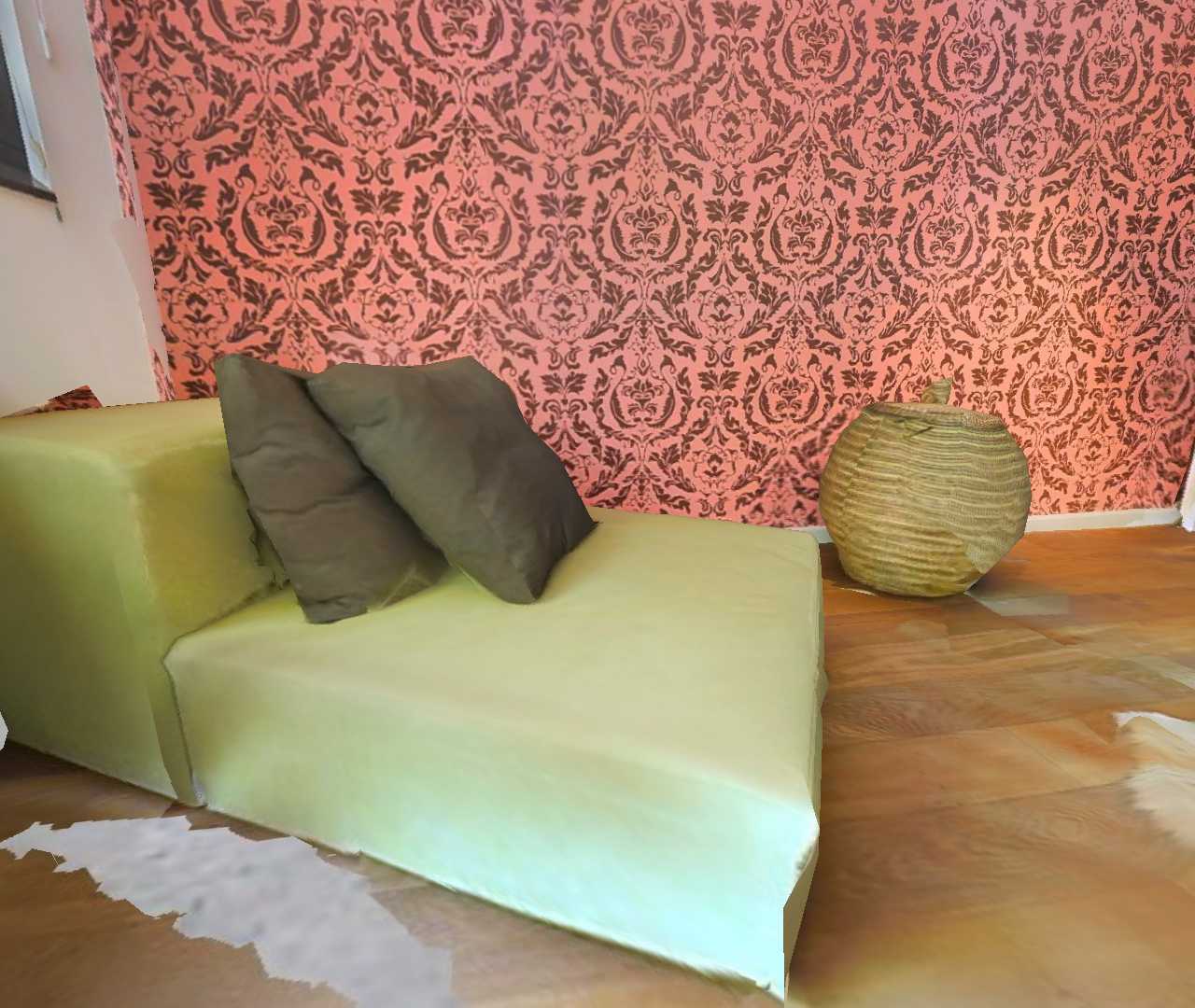}
\end{subfigure}
\begin{subfigure}{0.4\columnwidth}
  \centering
  \includegraphics[width=1\columnwidth, trim={0cm 0cm 0cm 1cm}, clip]{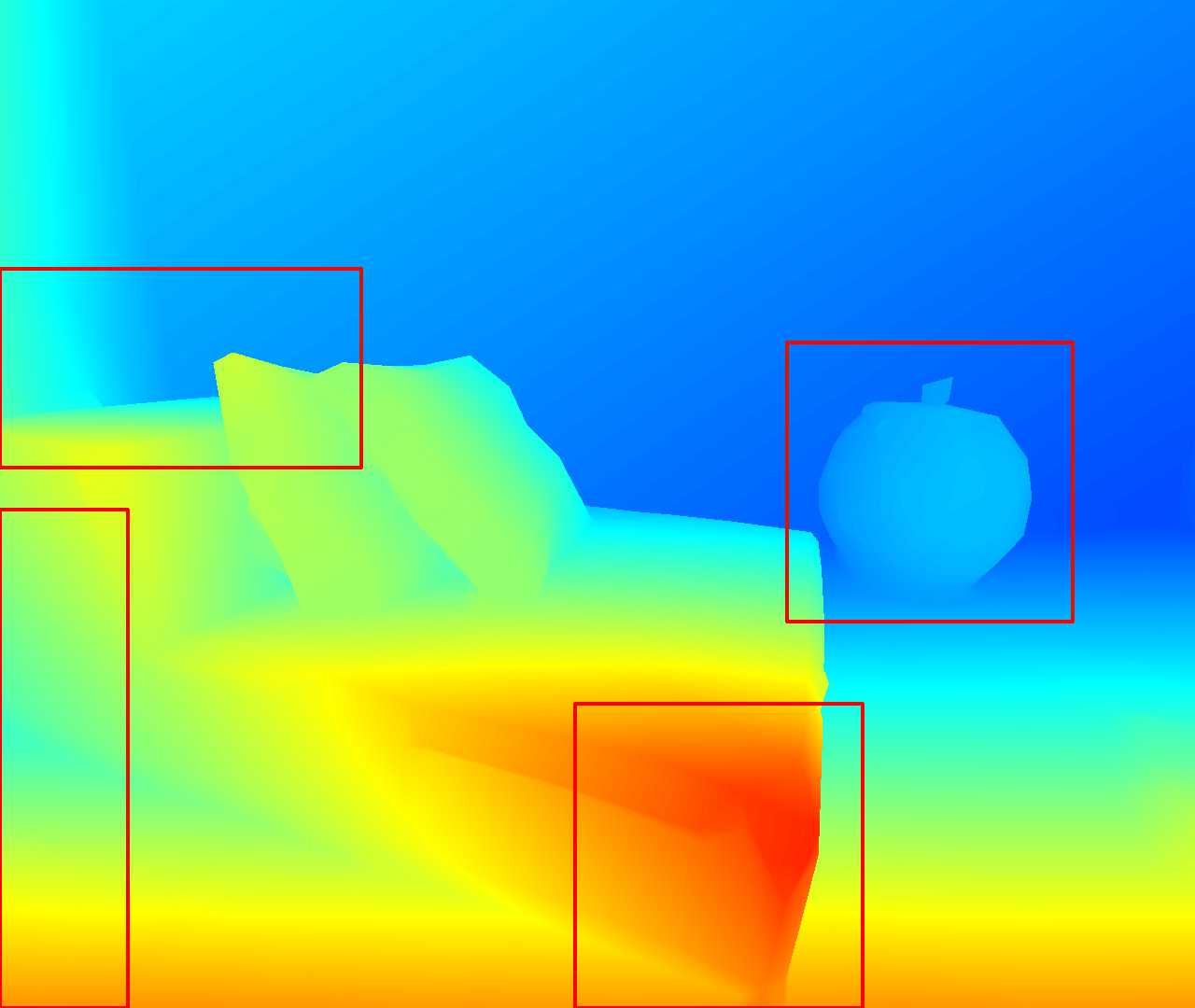}
\end{subfigure}
\begin{subfigure}{0.4\columnwidth}
  \centering
  \includegraphics[width=1\columnwidth, trim={0cm 0cm 0cm 1cm}, clip]{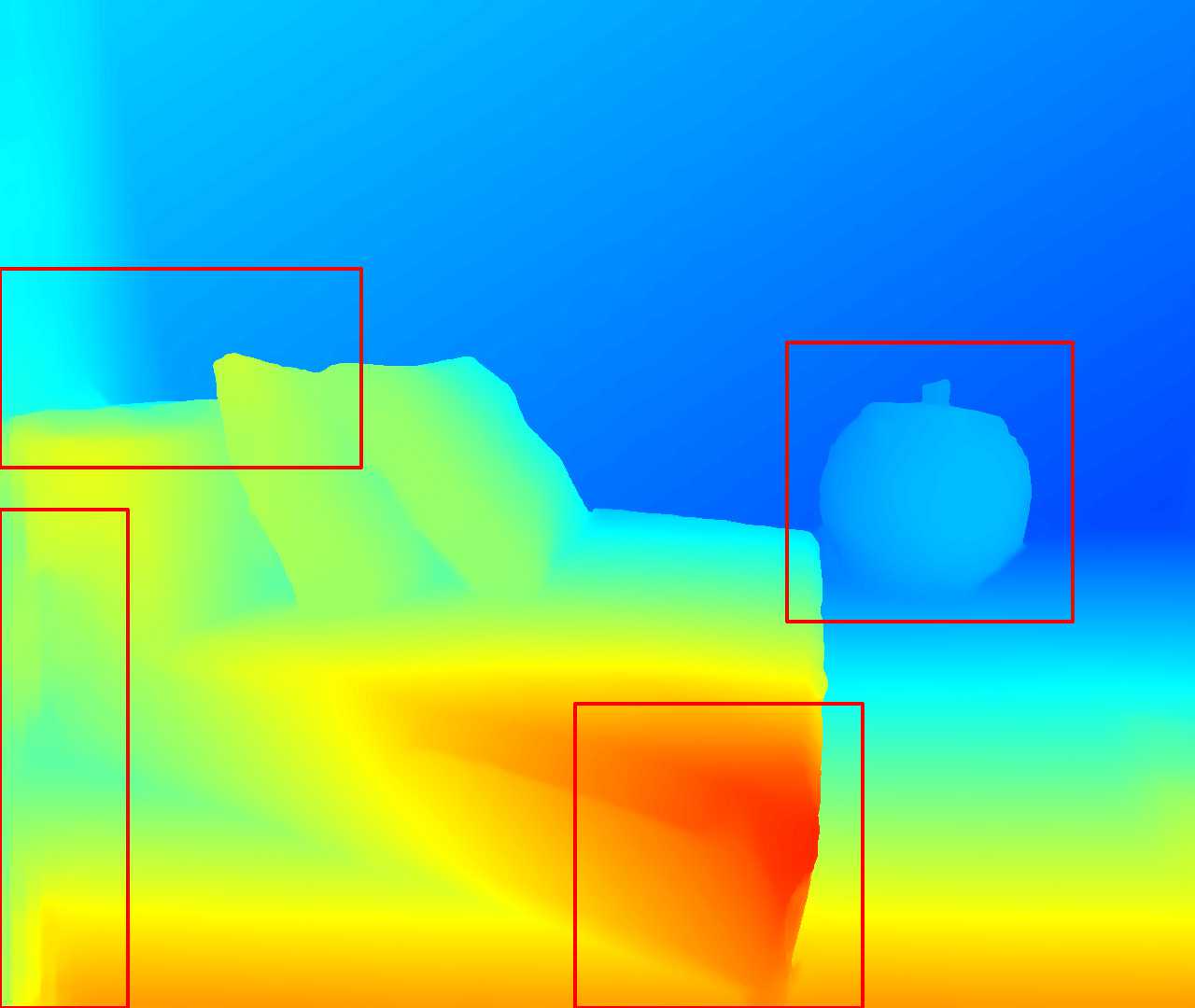}
\end{subfigure}
\begin{subfigure}{0.4\columnwidth}
  \centering
  \includegraphics[width=1\columnwidth, trim={0cm 0cm 0cm 1cm}, clip]{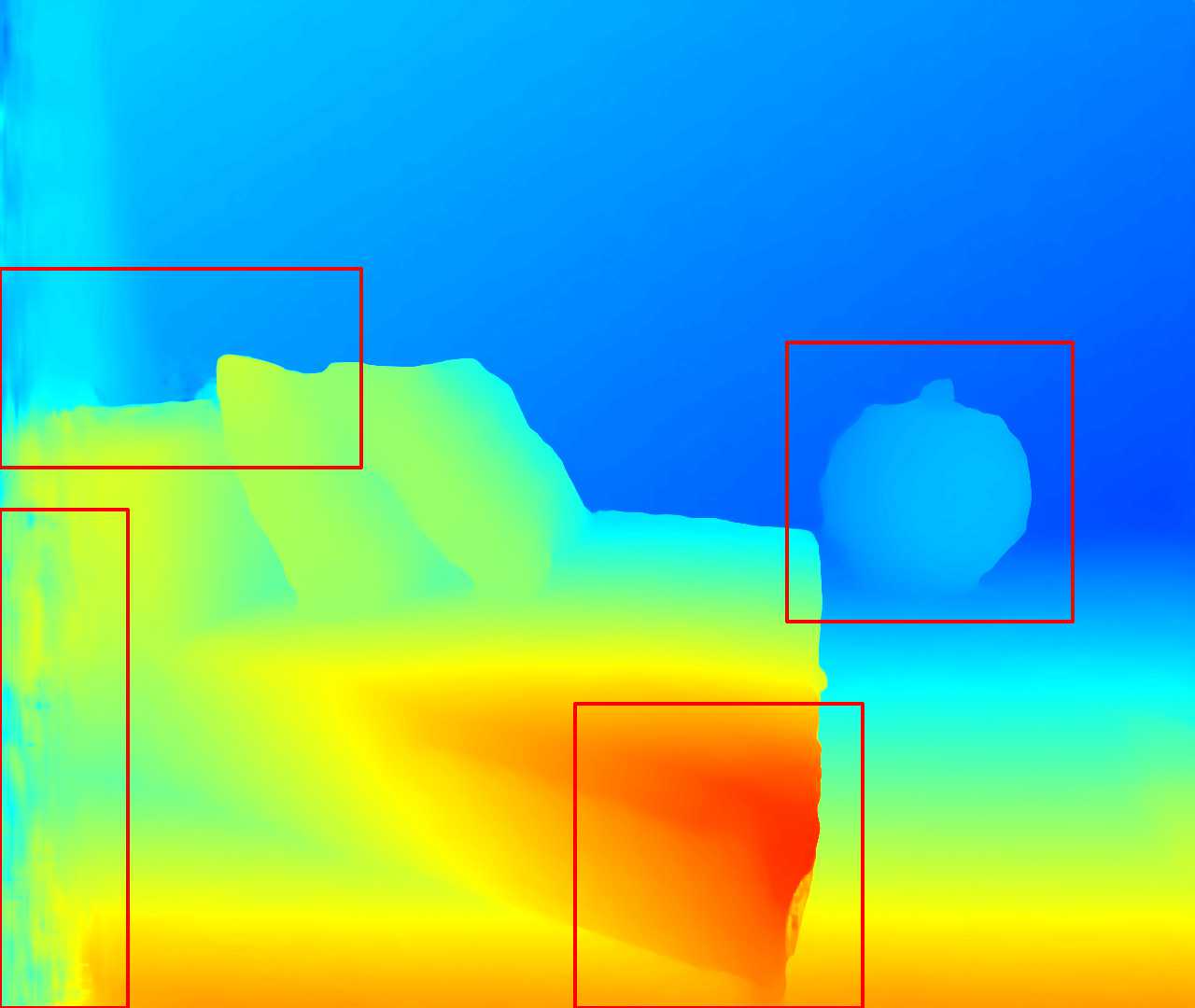}
\end{subfigure}

\begin{subfigure}{0.4\columnwidth}
  \centering
  \includegraphics[width=1\columnwidth, trim={0cm 0cm 0cm 0cm}, clip]{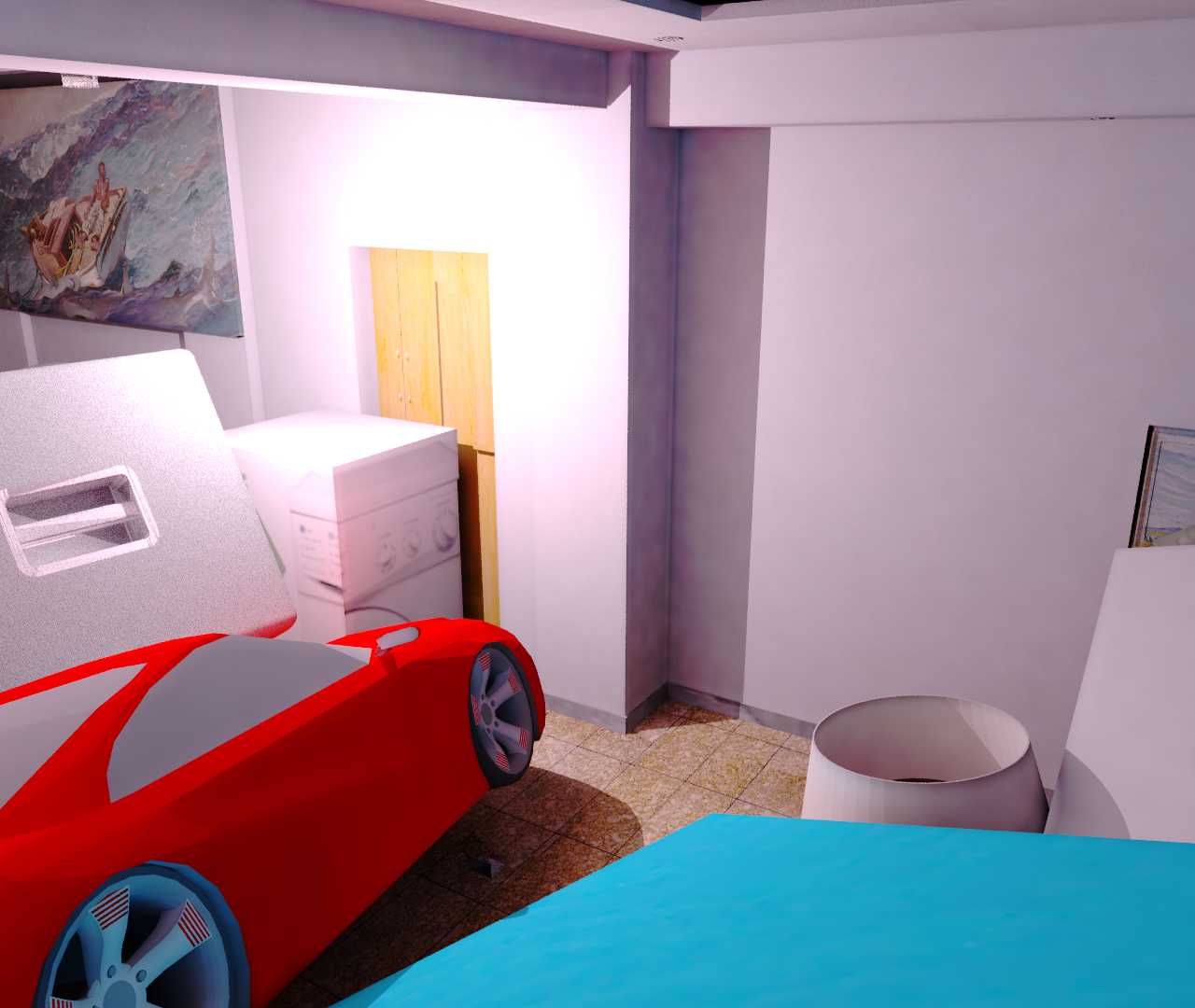}
  \caption*{Center view}
\end{subfigure}
\begin{subfigure}{0.4\columnwidth}
  \centering
  \includegraphics[width=1\columnwidth, trim={0cm 0cm 0cm 0cm}, clip]{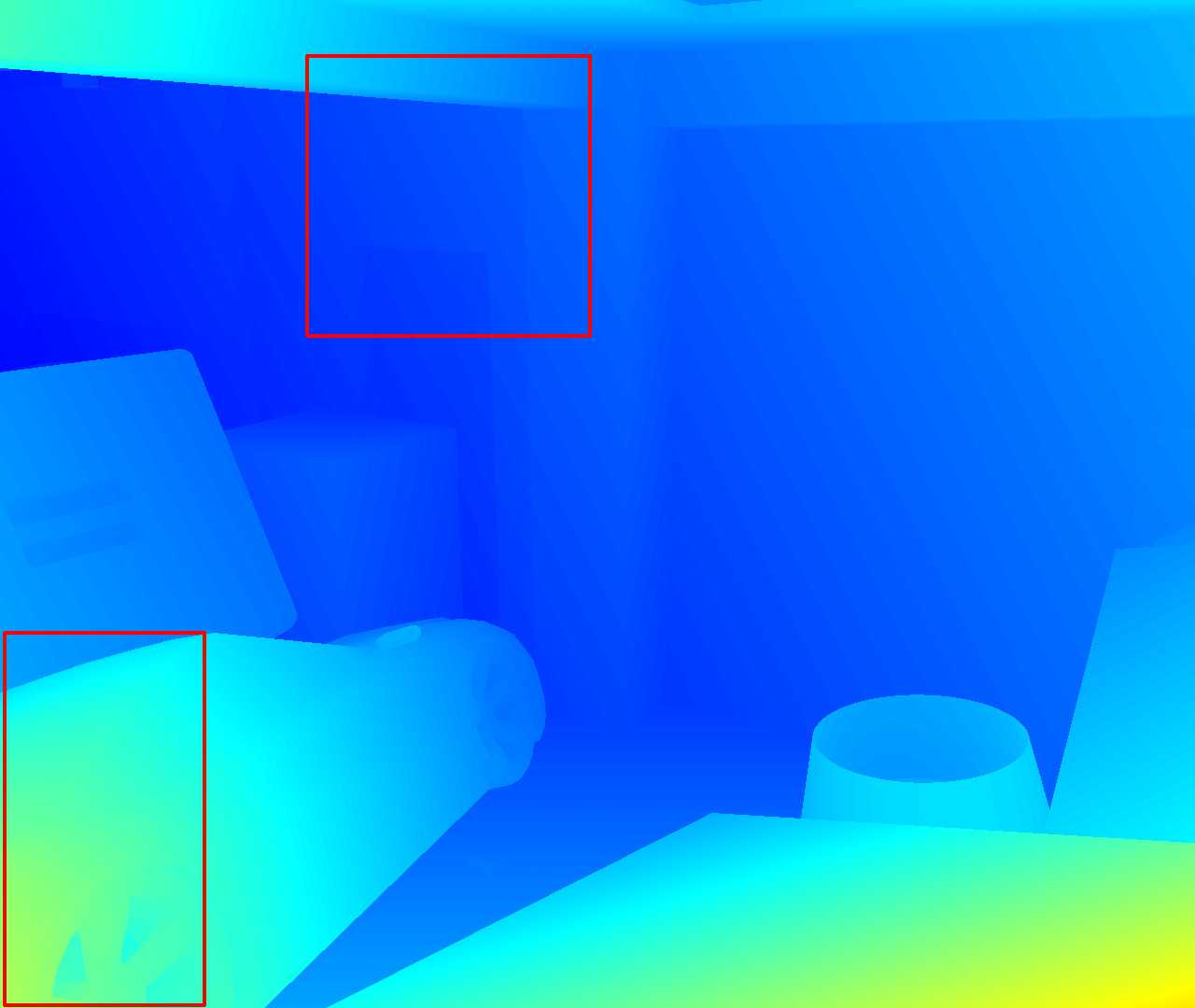}
  \caption*{Ground truth}
\end{subfigure}
\begin{subfigure}{0.4\columnwidth}
  \centering
  \includegraphics[width=1\columnwidth, trim={0cm 0cm 0cm 0cm}, clip]{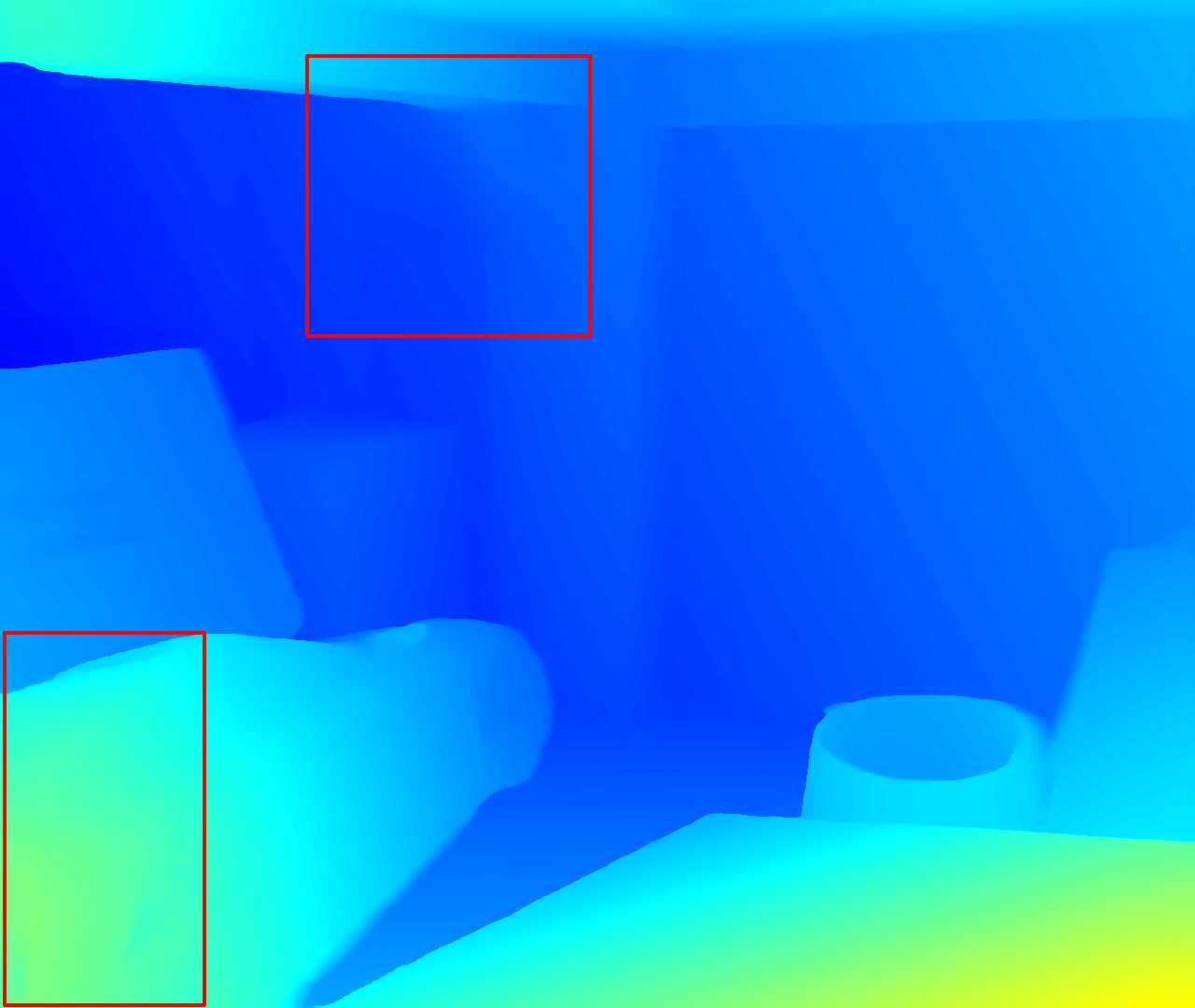}
  \caption*{Our model}
\end{subfigure}
\begin{subfigure}{0.4\columnwidth}
  \centering
  \includegraphics[width=1\columnwidth, trim={0cm 0cm 0cm 0cm}, clip]{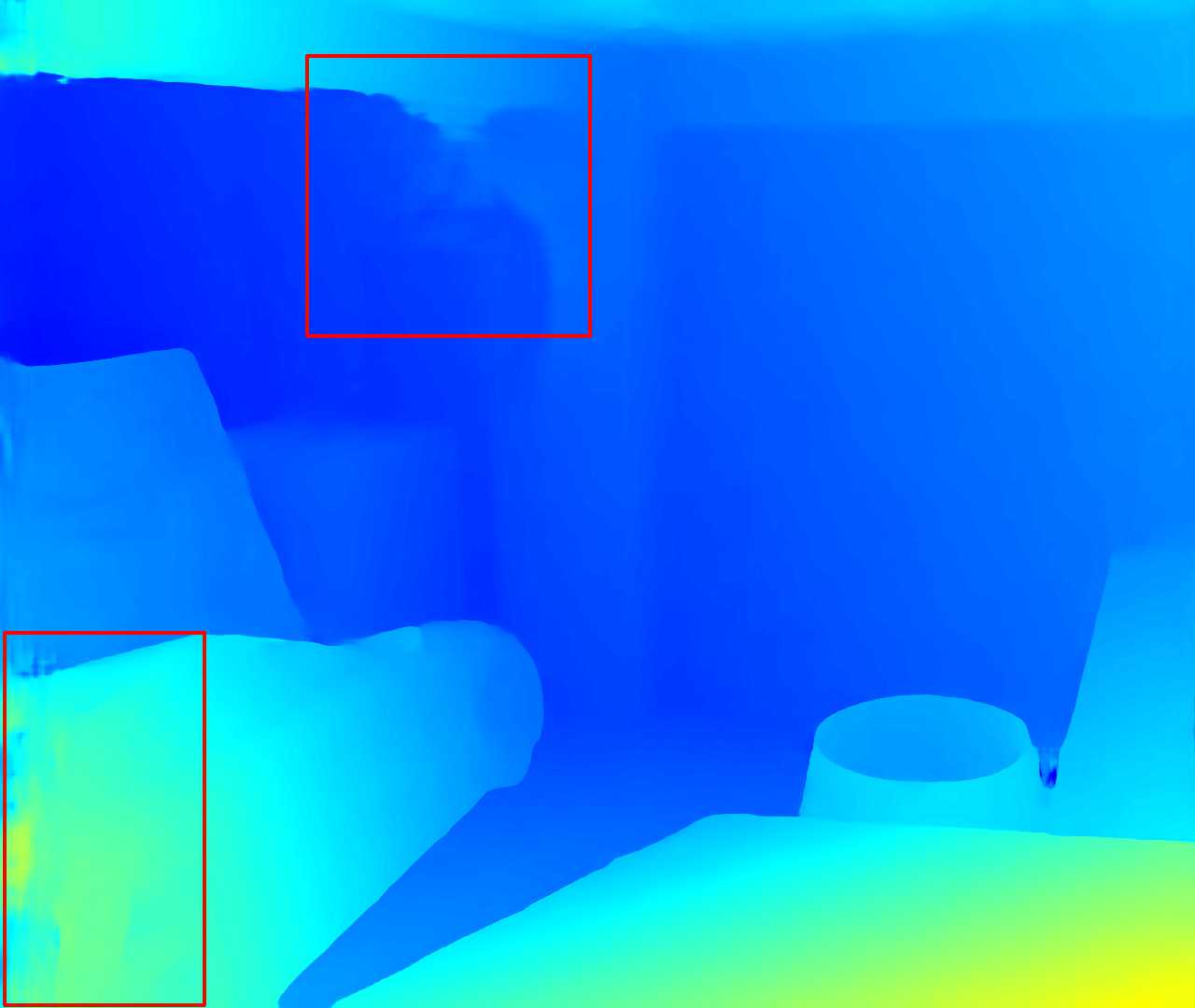}
  \caption*{PSMNet}
\end{subfigure}
\caption{Example outputs on synthetic test images. The center-view color images, the ground-truth disparity maps, the output disparity maps of our self-supervised model, and the output of supervised method PSMNet are shown.}
\label{fig:res-synthe}
\vspace{-0.2cm}
\end{figure*}

\setlength{\tabcolsep}{3pt}
\begin{table*}[]
\small
\centering
\begin{tabular}{c c c c c c c c c}
\toprule
Method & Training set & {Gound truth} & EPE  & Bad$0.5$ & Bad$1$ & Bad$2$  \\
\midrule
Godard et al.~\cite{godard2017unsupervised} & Synthetic & \xmark &
$2.16$ & $35.22\%$ & $21.67\%$ & $10.83\%$ \\
Ours & Synthetic & \xmark & $1.19$ & $\mathbf{16.54\%}$ & $\mathbf{9.46\%}$ & $\mathbf{6.07\%}$  \\
\midrule
PSMNet \cite{chang2018pyramid} & Synthetic & \cmark & $\mathbf{1.07}$  & $23.01\%$ & $11.69\%$ & ${6.36\%}$ \\
\midrule
MC-CNN \cite{zbontar2016stereo} & KITTI & \cmark & $3.55$ & $42.88\%$ & $27.18\%$ & $20.05\%$ \\
PSMNet \cite{chang2018pyramid} & Cityscapes + KITTI & \cmark & $2.94$ & $75.53\%$ & $50.12\%$ & $18.09\%$ \\
\bottomrule
\end{tabular}
\caption{Results on the synthetic test set.}
\label{tab:synthetic}
\vspace{-0.5cm}
\end{table*}
\setlength{\tabcolsep}{3pt}

\subsection{Network Structure}
While the design of the network structure is not our focus in this work, we use PSMNet \cite{chang2018pyramid} as our network backbone, which is an end-to-end disparity regression model. We trim the structure by reducing some layers for efficiency and add an uncertainty head. Details of the network are provided in the supplementary materials. The generated cost volume is used to regress a disparity map by the disparity head and is used to regresses an uncertainty map by the uncertainty head.

In our three-image multiscopic system, there are two matchings. One is between the center image and the left image, and the other is between the center image and the right image. We use two matching networks sharing the same parameters to calculate the disparity maps $D_c^l$ and $D_c^r$.

\subsection{Loss Function}

In our multiscopic self-supervised framework, we utilize the cross photometric loss $\mathcal{L}_P$ and the uncertainty-aware mutual-supervision loss $\mathcal{L}_\sigma, \mathcal{L}_M$ to guide the learning process. Additionally, a regularization term $\mathcal{L}_S$ is also introduced to improve the smoothness of the output disparity. Our overall loss is the sum of these terms with weighting factors $\lambda_1, \lambda_2, \lambda_3, \lambda_4$:
\begin{equation}
    \mathcal{L} = \lambda_1\mathcal{L}_P + \lambda_2\mathcal{L}_\sigma + \lambda_3\mathcal{L}_M + \lambda_4\mathcal{L}_S.
\end{equation}

\subsubsection{Cross Photometric Loss}
$\mathcal{L}_P$ is used to evaluate how similar the reconstructed images $I_{cl}^{l},I_{cl}^{r},I_{cr}^{r},I_{cr}^{l}$ are to the real center image $I_c$, where
\begin{equation}
\begin{aligned}
    I_{cl}^{l}(u,v) &= I_l(u+D_c^l(u,v), v),\\
    I_{cl}^{r}(u,v) &= I_l(u+D_c^r(u,v), v).
\end{aligned}
\end{equation}
$I_{cr}^{r},I_{cr}^{l}$ are calculated in a similar way. The warping process is illustrated in Fig.~\ref{fig:warp}.

We use structural similarity (SSIM) \cite{wang2004image} loss to calculate the image discrepancy as
\begin{equation}
    \mathcal{L}_P(I_{cl}^l, I_c) = \frac{1-SSIM(I_{cl}^l, I_c)}{2},
\end{equation}
where the SSIM window size is set to $5$.
The other three parts are calculated in a similar way, and the final cross photometric loss is computed by taking the average:
\begin{equation}
    \mathcal{L}_P=\frac{\mathcal{L}_P(I_{cl}^l)+
                        \mathcal{L}_P(I_{cl}^r)+
                        \mathcal{L}_P(I_{cr}^l)+
                        \mathcal{L}_P(I_{cr}^r)}{4}.
\end{equation}

\begin{figure*}[t]
\centering
\begin{subfigure}{0.6\columnwidth}
  \centering
  \includegraphics[width=1\columnwidth, trim={0cm 0cm 0cm 0cm}, clip]{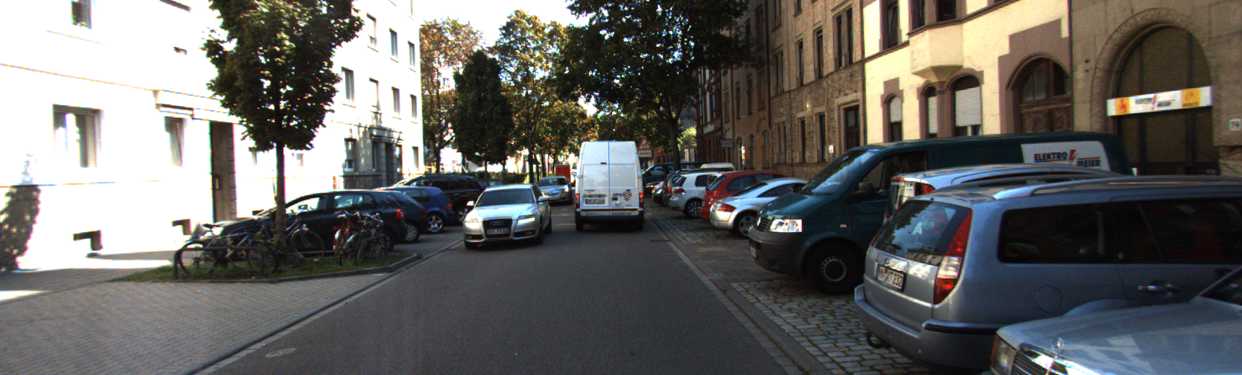}
\end{subfigure}
\begin{subfigure}{0.6\columnwidth}
  \centering
  \includegraphics[width=1\columnwidth, trim={0cm 0cm 0cm 0cm}, clip]{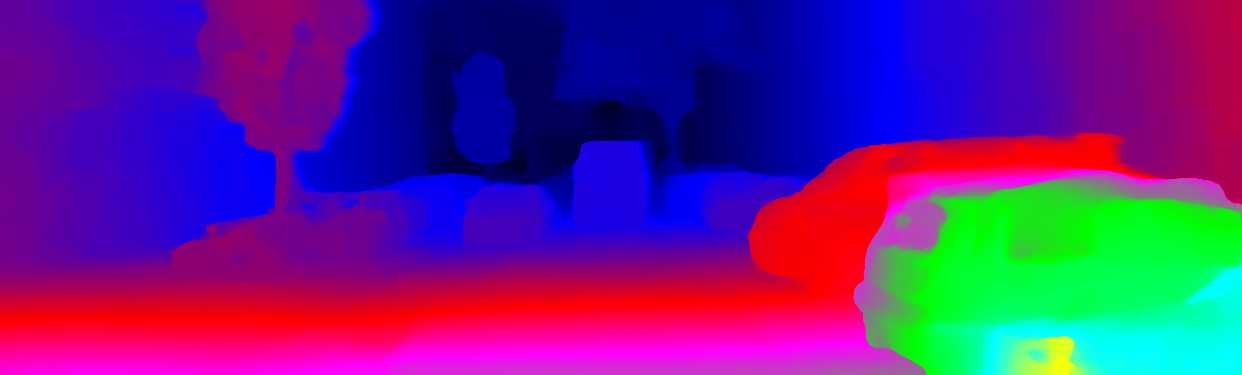}
\end{subfigure}
\begin{subfigure}{0.6\columnwidth}
  \centering
  \includegraphics[width=1\columnwidth, trim={0cm 0cm 0cm 0cm}, clip]{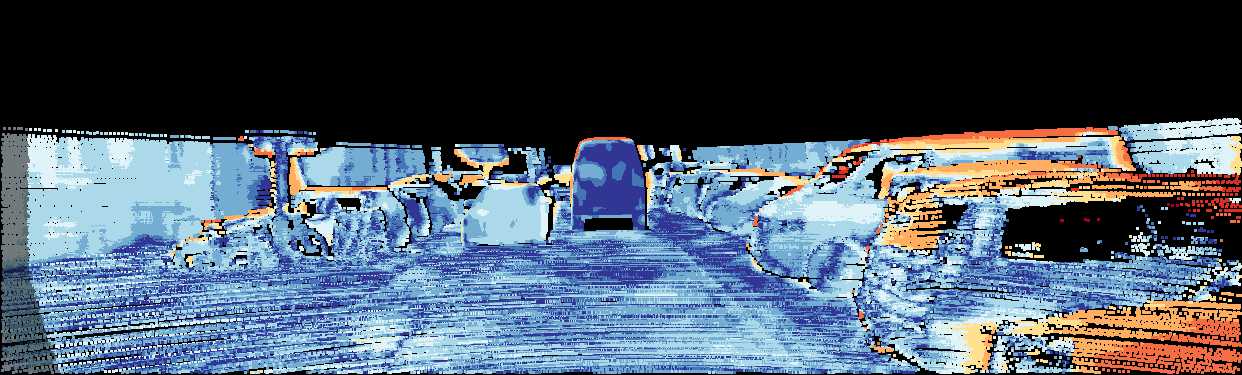}
\end{subfigure}

\begin{subfigure}{0.6\columnwidth}
  \centering
  \includegraphics[width=1\columnwidth, trim={0cm 0cm 0cm 0cm}, clip]{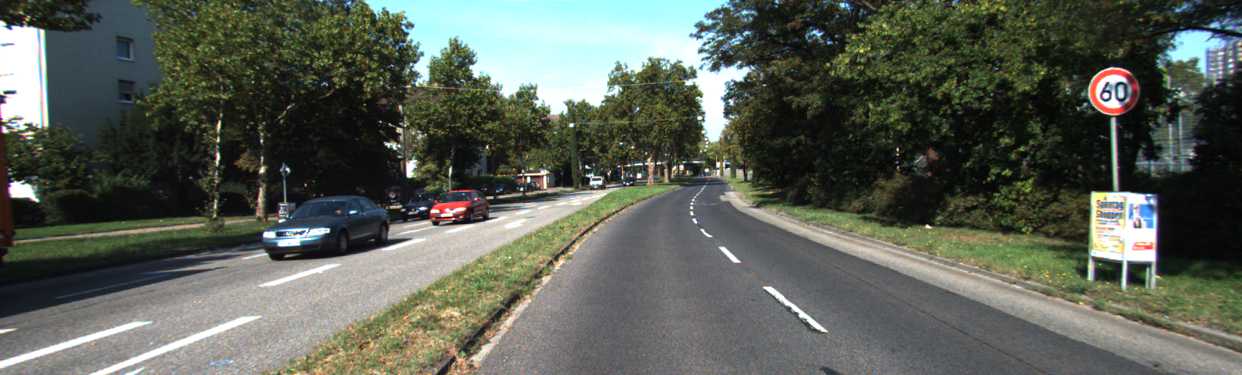}
  \caption*{Left image}
\end{subfigure}
\begin{subfigure}{0.6\columnwidth}
  \centering
  \includegraphics[width=1\columnwidth, trim={0cm 0cm 0cm 0cm}, clip]{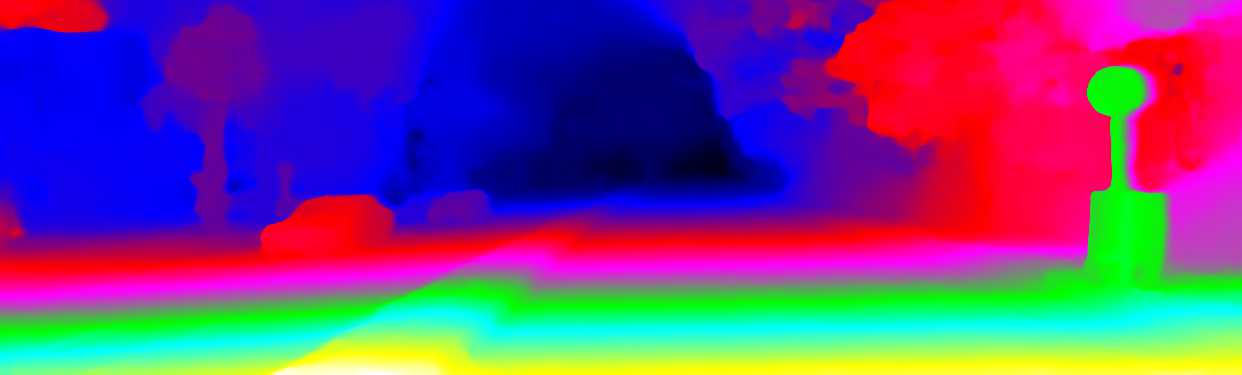}
  \caption*{Disparity map}
\end{subfigure}
\begin{subfigure}{0.6\columnwidth}
  \centering
  \includegraphics[width=1\columnwidth, trim={0cm 0cm 0cm 0cm}, clip]{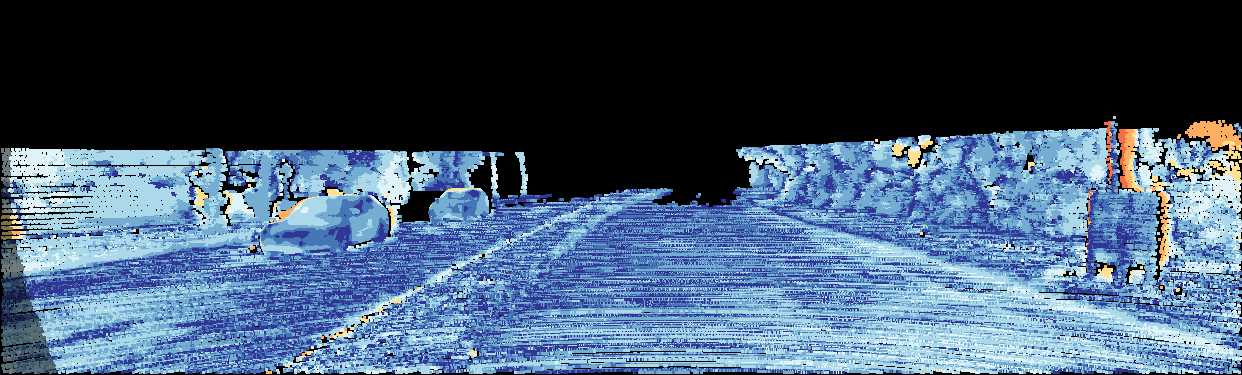}
  \caption*{Error map}
\end{subfigure}
\caption{Qualitative results on KITTI 2015 test set.}
\label{fig:res-kitti}
\vspace{-0.3cm}
\end{figure*}

\subsubsection{Uncertainty-aware Mutual-supervision Loss}
$\mathcal{L}_M$ utilizes the output of one network to supervise the other network. The discrepancy between the disparity output $D_c^l$ generated by the center-left network and $D_c^r$ generated by the center-right network means erroneous matching. Thus we learn an uncertainty map for each disparity output. In uncertain areas, we supervise the uncertain output with the other output, and in confident areas, we calculate the consistency loss between two outputs.

Following the heteroscedastic aleatoric uncertainty setup in \cite{kendall2017uncertainties}, we model the reconstruction task with uncertain L1 loss as
\begin{equation}
\begin{aligned}
    \mathcal{L}_\sigma(I_{cl}^l) = \sum_{u,v} \frac{\sqrt{2}}{\sigma^l(u,v)}|I_{cl}^l(u,v)-I_c(u,v)| + \log \sigma^l(u,v) ,
\end{aligned}
\end{equation}
where $\sigma^l$ is the estimated uncertainty map generated by center-left image pair. Then $\mathcal{L}_\sigma$ is computed by taking the average of $\mathcal{L}_\sigma(I_{cl}^l), \mathcal{L}_\sigma(I_{cl}^r), \mathcal{L}_\sigma(I_{cr}^l), \mathcal{L}_\sigma(I_{cr}^r)$.

With the uncertainty map $\sigma_l, \sigma_r$, we can do the mutual-supervising as

\begin{equation}
\begin{aligned}
\mathcal{L}_M=
  \begin{cases}
  ||D_c^l-D_c^r|| , \quad & \text{if} \ \sigma_l, \sigma_r < T_\sigma \\
  ||D_c^l-\text{label}(D_c^r)|| , \quad & \text{if} \ \sigma_l \geq T_\sigma, \sigma_r < T_\sigma \\
  ||D_c^r-\text{label}(D_c^l)|| , \quad & \text{if} \ \sigma_r \geq T_\sigma, \sigma_l < T_\sigma .
  \end{cases}
\end{aligned}
\label{equ:lossm}
\end{equation}
where $T_\sigma=e$ is a threshold over which the disparity estimation is regarded as uncertain, and the label operation converts the confident estimation to supervision signal that is used to optimize the estimation of the other network. In regions where both estimations are uncertain, we do not calculate the loss. This loss could be especially useful in occluded regions. When the matching between the center-left images is wrong due to occlusion, it can be corrected by the label from the center-right matching.


\subsubsection{Smoothness Loss}
$\mathcal{L}_S$ encourages the adjacent pixels to have similar disparities, especially for those with a similar color. The punishment is enforced for discontinuous disparities and determined by the disparity variation and color difference, as
\begin{equation}
\begin{split}
    \mathcal{L}_S=\
    &f_x(D_c^l)e^{-\gamma||\partial_x I_c||}+f_y(D_c^l)e^{-\gamma||\partial_y I_c||}+\\
    &f_x(D_c^r)e^{-\gamma||\partial_x I_c||}+f_y(D_c^r)e^{-\gamma||\partial_y I_c||},
\end{split}
\end{equation}
where $\gamma$ is a scale factor related to the image intensity normalization and is set to $10$ in our work. $f_{x/y}$ is composed of a discontinuous loss $K$ plus the disparity differential as
\begin{equation}
f_x(D)=
\begin{cases}
    K+||\partial_x D|| , \quad & \text{if} \ ||\partial_x D||>0.5\\
    ||\partial_x D||  , \quad & \text{otherwise}.
\end{cases}
\end{equation}
The disparity difference larger than $0.5$ is considered discontinuity and is punished by an additional $K=10$.

\setlength{\tabcolsep}{5pt}
\begin{table}[]
\centering
\begin{tabular}{c c c c}
\toprule
Method & \tabincell{c}{\#Infer \\ input} & \tabincell{c}{Train set\\D1-all}  & \tabincell{c}{Test set\\D1-all}  \\
\midrule
Zhou et al. 2017~\cite{zhou2017unsupervised} & $2$ & $9.91\%$ & $-$ \\
Godard et al. 2017~\cite{godard2017unsupervised} & $2$ & $9.194\%$ & $-$ \\
Li et al. 2018~\cite{li2018occlusion} & $2$ & $-$ & $8.98\%$ \\
Guo et al. 2018~\cite{guo2018learning} & $2$ & $7.06\%$ & $-$ \\
UnOS 2019~\cite{wang2019unos} & $2$ & $7.073\%$ & $-$ \\
UnOS 2019~\cite{wang2019unos} & $4$ & $5.943\%$ & $6.67\%$ \\
Wang et al. 2020~\cite{wang2020parallax} & $2$ & $-$ & $7.23\%$ \\
Flow2Stereo 2020~\cite{liu2020flow2stereo} & $2$ & $6.13\%$ & $6.61\%$ \\
Ours & $2$ & $\mathbf{3.82\%}$ & $\mathbf{4.43\%}$  \\
\midrule
PSMNet & $2$ & $1.83\%$ & $2.32\%$  \\
PSMNet* & $2$ & $5.90\%$ & $-$  \\
\bottomrule
\end{tabular}%
\caption{Quantitative evaluation on the KITTI 2015 dataset. D1-all is the percentage of stereo disparity outliers in all regions of the first frame. End-point error smaller than $3$ pixels or $5\%$ is considered correct.}
\label{tab:kitti}
\vspace{-0.5cm}
\end{table}
\setlength{\tabcolsep}{3pt}


\section{Experiments}

In this section, we first present the details of our experimental setup. Afterward, we evaluate our model on the synthetic test set, KITTI 2015 dataset, and real-world images. Also, ablation studies are performed to provide insight into the effect of each module in our framework.

\subsection{Implementation Details}

The self-supervised learning framework is implemented in Pytorch and trained on two Nvidia GTX 1080 Ti GPUs. The batch size is set to 4 and the weights $\lambda_1, \lambda_2, \lambda_3, \lambda_4$ are set to $1,0.01,0.03,0.03$. 
Images are randomly cropped to $512\times256$ for every epoch during training to reduce the size and work as data augmentation. The network is optimized end-to-end with Adam ($\beta_1=0.9$, $\beta_1=0.999$) and the learning rate is set to $0.0001$.

Our network is only trained using 1,000 scenes of synthetic data. The performance trained with additional real multiscopic images is similar since our synthetic images are quite photo-realistic.
During training, three multiscopic images are used as the input of our learning framework because the performance of using five images is similar, and our real-world system captures three images. For evaluation, we only use two stereo images for depth estimation. Examples of the output $D_c^r$ between the right view and the center view of test images are shown in Fig.~\ref{fig:res-synthe}. The output disparity maps are very close to the ground truth disparity maps.
{For the computation efficiency, since the backbone of our network is based on PSMNet}~\cite{chang2018pyramid}{, the network forward time is similar at around 0.5~s for one KITTI image pair. By choosing the backbone, the balance between accuracy and efficiency can be achieved.}

\subsection{Evaluation}

\subsubsection{Synthetic data} We build a synthetic test set consisting of 200 scenes of multiscopic images. After the training, we evaluate our model on these images to examine its performance on synthetic images. The center view and the right view are used as the input to the network. Four metrics are applied, and the results are summarized in Table~\ref{tab:synthetic}. The EPE is the average endpoint error, Bad$0.5$, Bad$1$, and Bad$2$ is the percentage of pixels whose error is higher than $0.5$, $1$, and $2$, respectively.

We also compare our method with some unsupervised and supervised methods when trained on our synthetic dataset, as presented in Table~\ref{tab:synthetic}. The results show that our method is better than the previous unsupervised method \cite{godard2017unsupervised} and comparable to the supervised method PSMNet. Qualitative results are displayed in Fig.~\ref{fig:res-synthe}.
Then, we test the performance of the original supervised models of the MC-CNN and PSMNet trained on the KITTI dataset. The errors of their outputs are high, which illustrates that the generalization ability of these supervised methods is not satisfactory.

\subsubsection{KITTI 2015}

To compare our method with other unsupervised or self-supervised approaches, we evaluate the trained model on the KITTI 2015 dataset \cite{geiger2012we}, which includes 200 training stereo images pairs and 200 test pairs. The ground truth for the test set is withheld for evaluation. From the quantitative results shown in Table~\ref{tab:kitti}, our method outperforms all previous unsupervised learning methods by a large margin even without fine-tuning on the KITTI data. In other words, our model is not trained on outdoor images but can perform well on outdoor images in the KITTI dataset. We believe our model could achieve a lower error if trained on KITTI-style multiscopic images. The D1-all error of Godard~\cite{godard2017unsupervised} trained with our synthetic images is $11.36\%$ while its error reduces to $9.194\%$ after the fine-tuning on KITTI images.
Also, the second-best performing method \cite{wang2019unos} utilizes two consecutive stereo pairs while we only use one pair.

Some qualitative results are displayed in Fig.~\ref{fig:res-kitti} for visualization. The results demonstrate the generalization ability of our method, which benefits from self-supervised learning. In self-supervised learning, since there is no ground truth depth information, there is no chance to overfit the depth value. Instead, the network needs to find the inner pattern and geometry shape underneath.


We also compare our results with the supervised method PSMNet when trained on our synthetic dataset without fine-tuning on the KITTI images, denoted as ``PSMNet*". When the model is trained with KITTI data, the D1 error on KITTI 2015 can reach $1.83\%$. However, the error is $5.90\%$ when trained on our synthetic training set. This means if not trained on the same dataset, the performance of this supervised method does not have advantages over our self-supervised approach. Generalization between different datasets suffers from a large performance decrease, which is the drawback of supervised learning methods.

\begin{figure}[t]
\centering
\begin{subfigure}{0.32\columnwidth}
  \centering
  \includegraphics[width=1\columnwidth, trim={0cm 0cm 0cm 0cm}, clip]{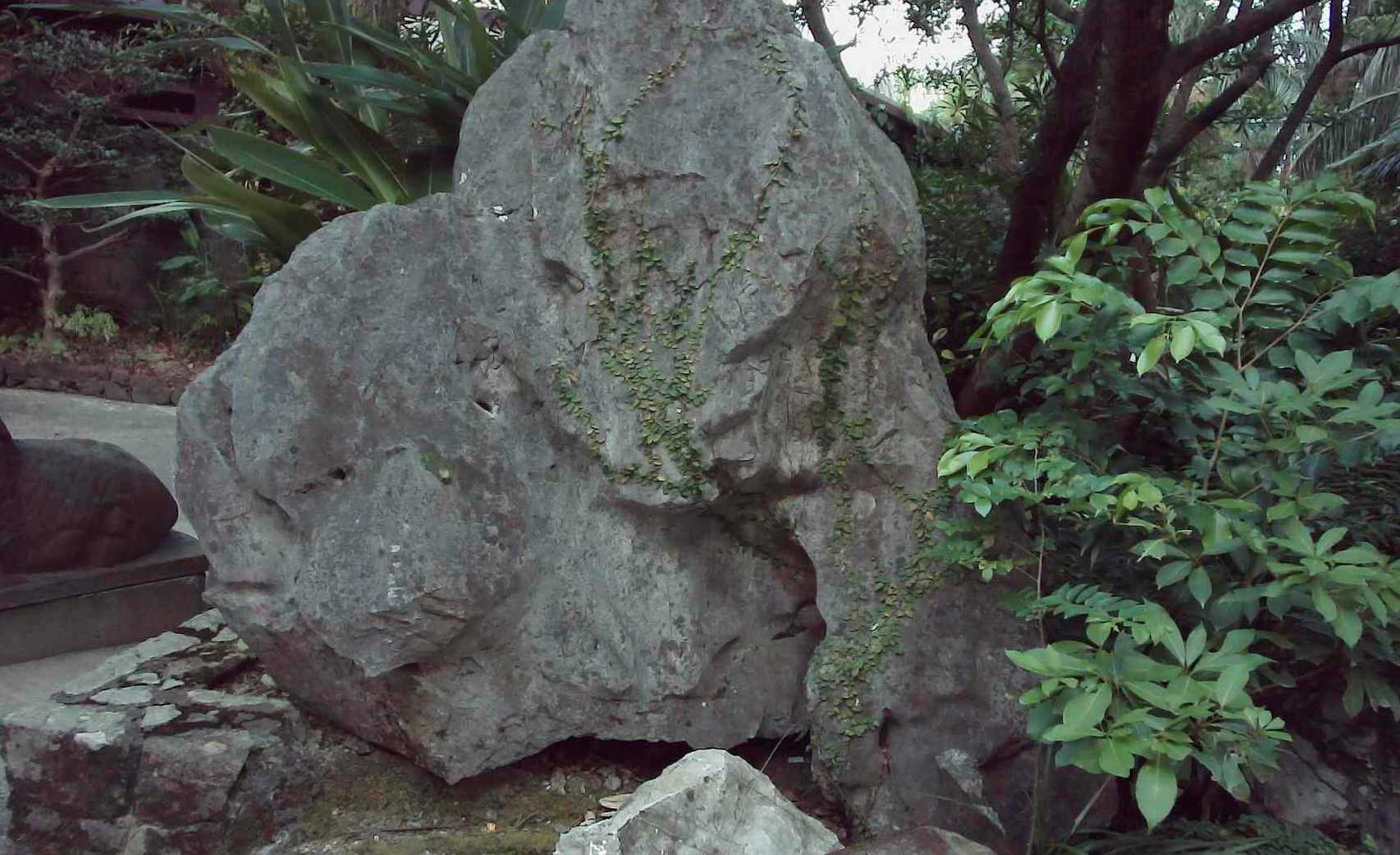}
\end{subfigure}
\begin{subfigure}{0.32\columnwidth}
  \centering
  \includegraphics[width=1\columnwidth, trim={0cm 0cm 0cm 0cm}, clip]{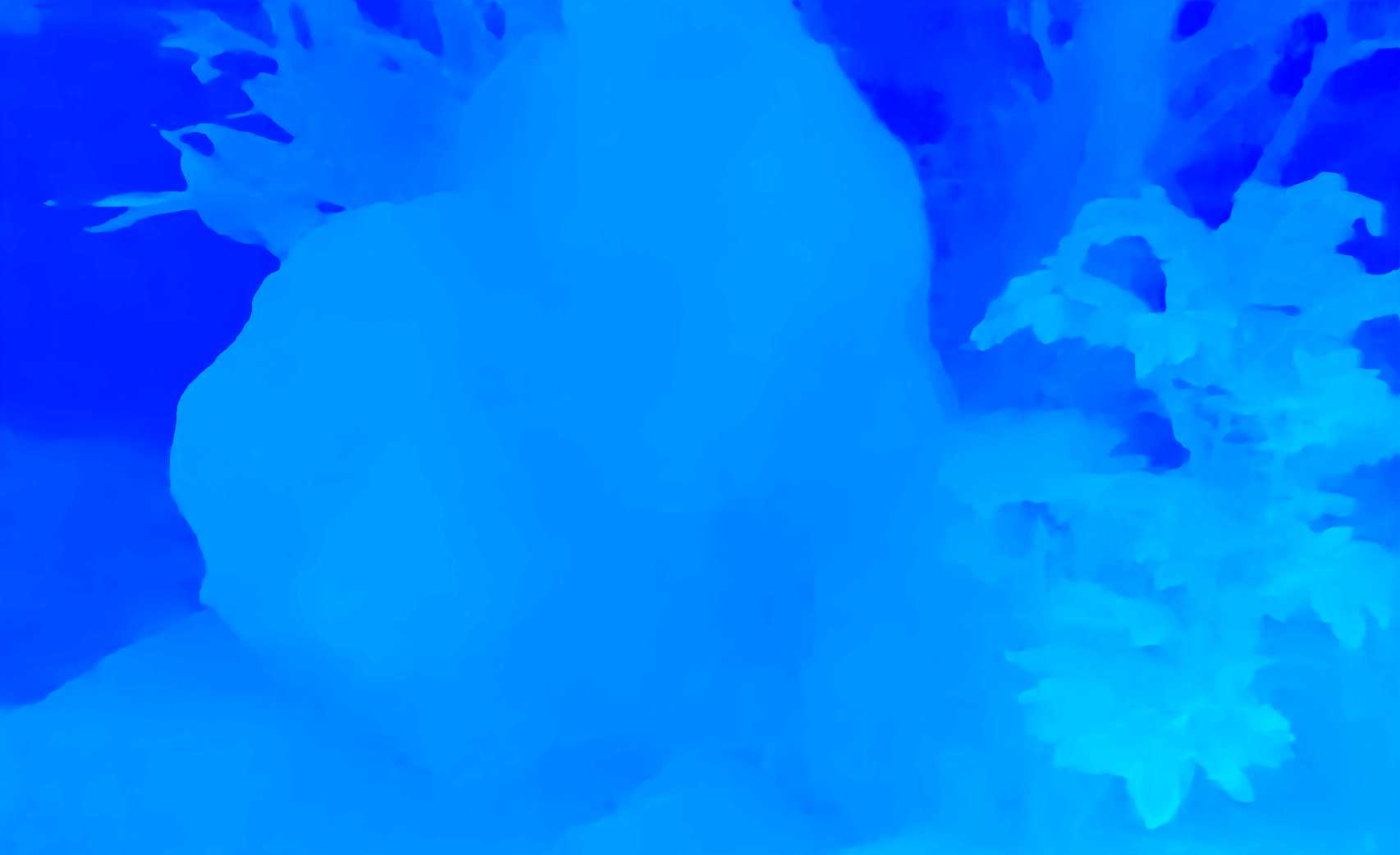}
\end{subfigure}
\begin{subfigure}{0.32\columnwidth}
  \centering
  \includegraphics[width=1\columnwidth, trim={0cm 0cm 0cm 0cm}, clip]{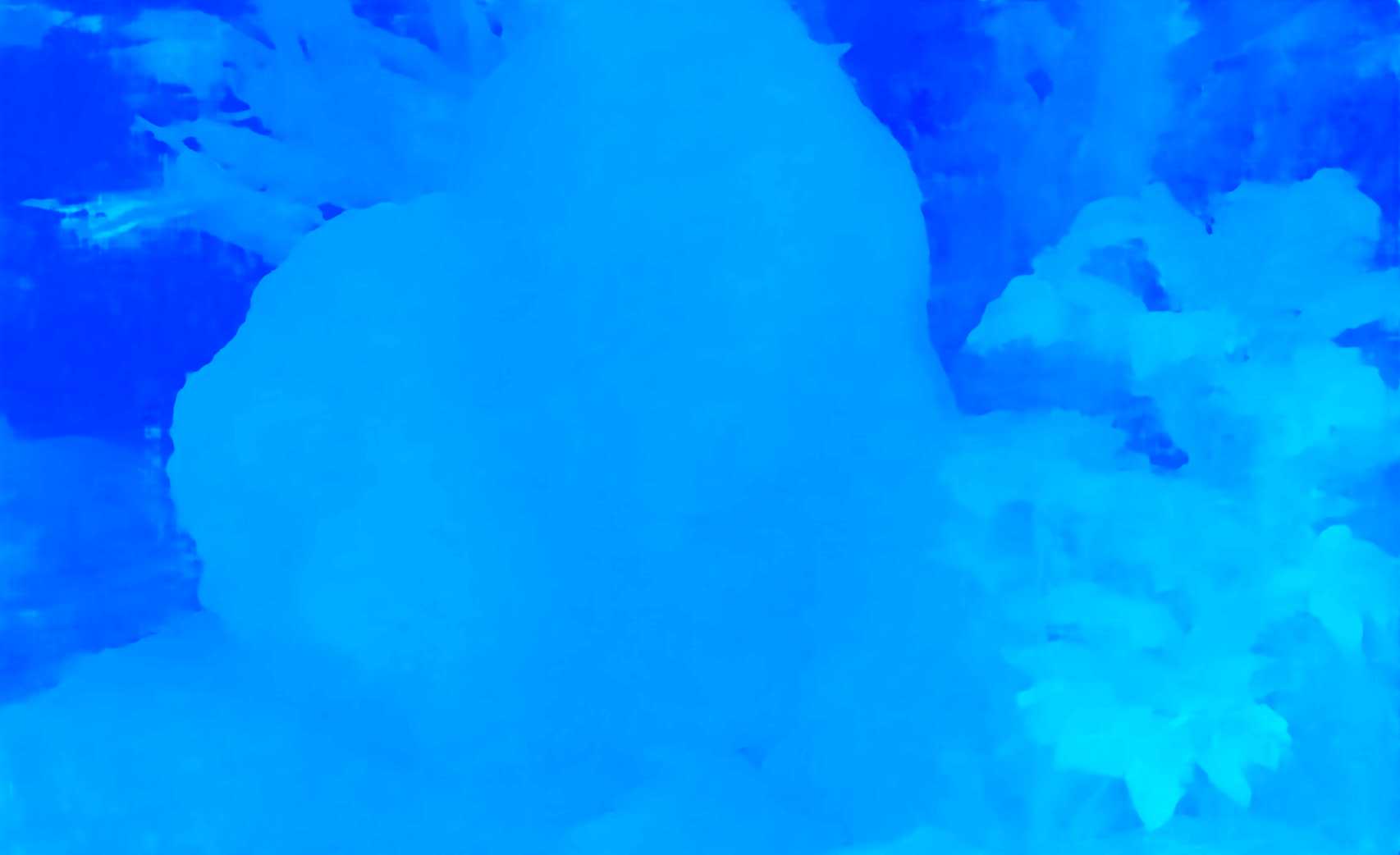}
\end{subfigure}

\begin{subfigure}{0.32\columnwidth}
  \centering
  \includegraphics[width=1\columnwidth, trim={0cm 0cm 0cm 0cm}, clip]{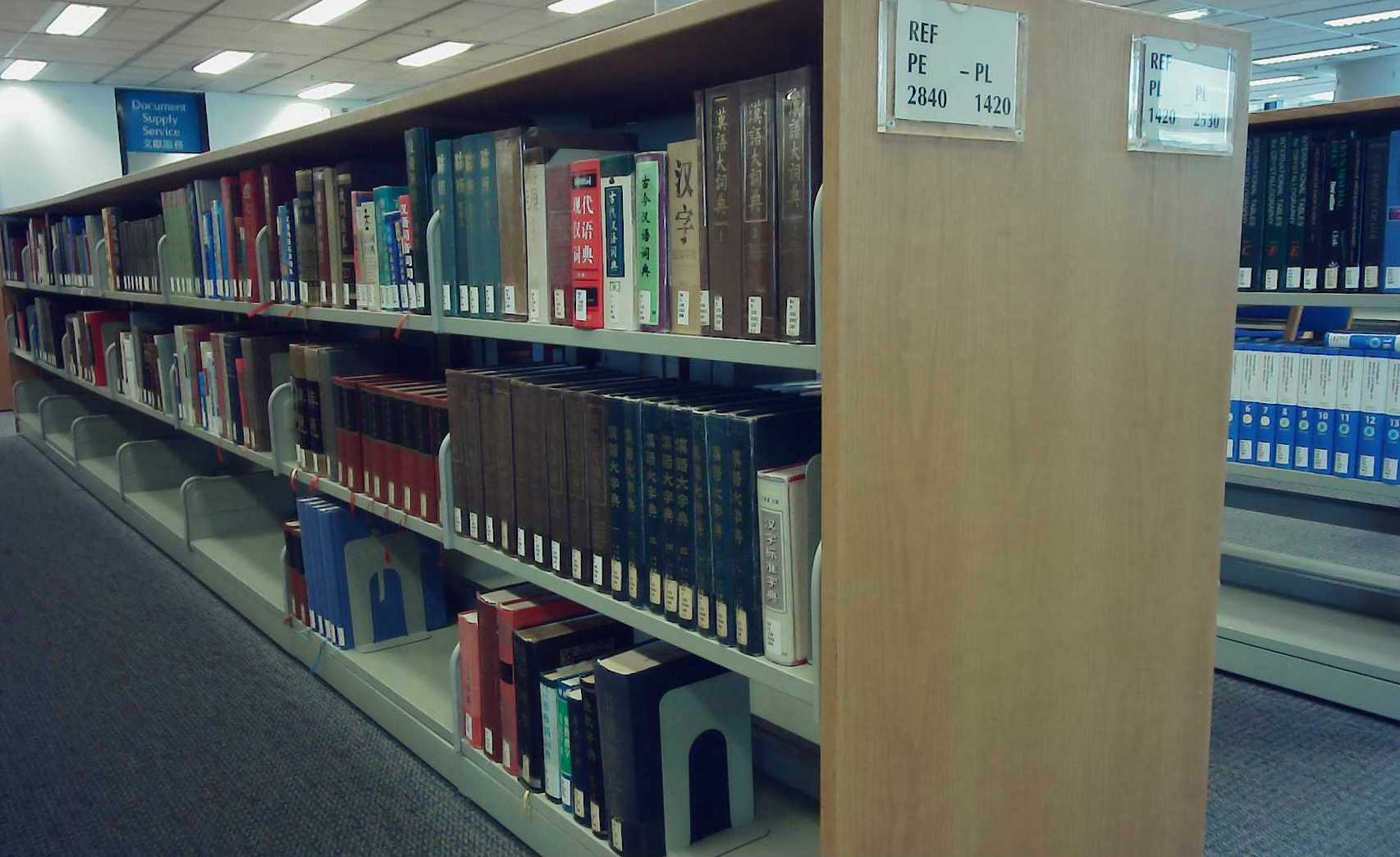}
  \caption*{Color image}
\end{subfigure}
\begin{subfigure}{0.32\columnwidth}
  \centering
  \includegraphics[width=1\columnwidth, trim={0cm 0cm 0cm 0cm}, clip]{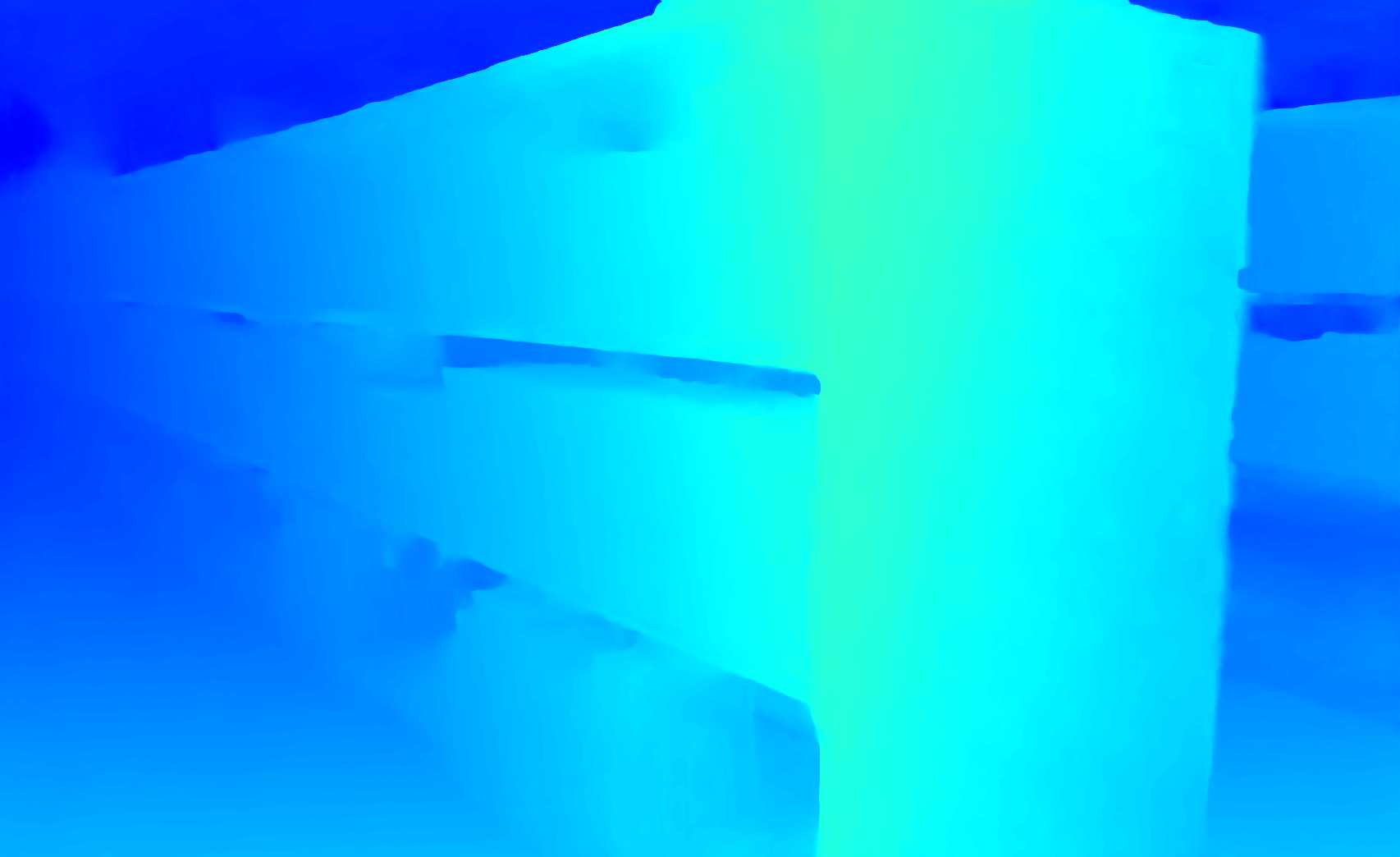}
  \caption*{Our model}
\end{subfigure}
\begin{subfigure}{0.32\columnwidth}
  \centering
  \includegraphics[width=1\columnwidth, trim={0cm 0cm 0cm 0cm}, clip]{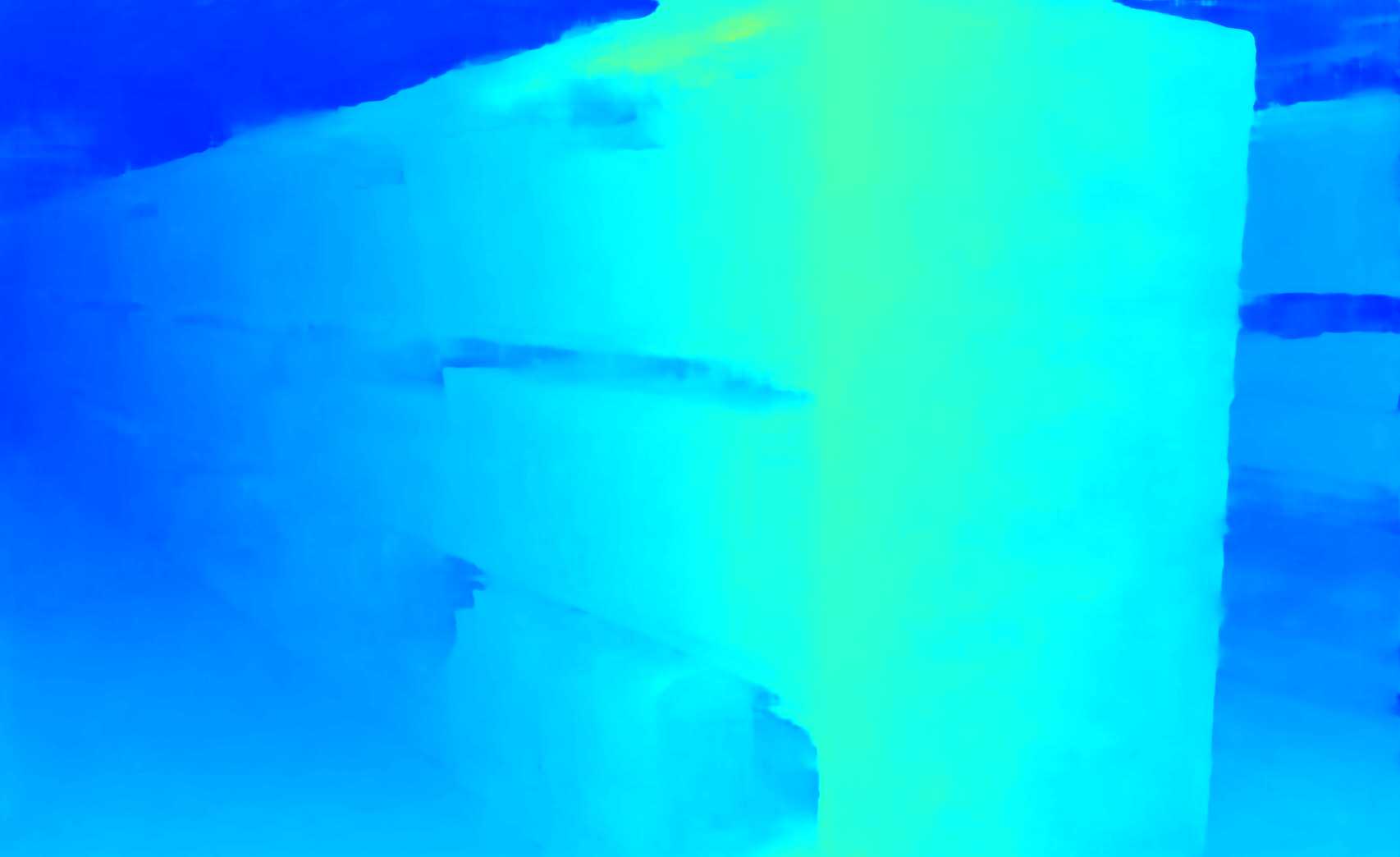}
  \caption*{PSMNet}
\end{subfigure}
%
%
\caption{Test on real images in our dataset. The output of our model and the supervised method PSMNet are presented.}
\label{fig:res-real}
\vspace{-0.5cm}
\end{figure}

\subsubsection{Real-world Data and Online Fine-tuning}

To further inspect the generalization ability of our method, we test the network on the images captured by the real cameras in our dataset. The center-view color image and the output disparity map between the center view and the right view are shown in Fig.~\ref{fig:res-real}. From the qualitative results, we can see the network still performs stably and obtains better disparity maps than the supervised method PSMNet after being trained on our synthetic dataset. This indicates that after the training, our self-supervised model can perform more stably in practical applications.
Another advantage of the self-supervised method is that online fine-tuning becomes feasible. Therefore, we fine-tune the network using the newly captured images when being employed in the real world. The results show the estimated disparity maps are further improved, which are presented in supplementary materials. This demonstrates the practical usefulness of our self-supervised multiscopic framework.

\begin{figure}[t]
\centering
\begin{subfigure}{0.4\columnwidth}
  \centering
  \includegraphics[width=1\columnwidth, trim={8cm 0cm 10cm 17cm}, clip]{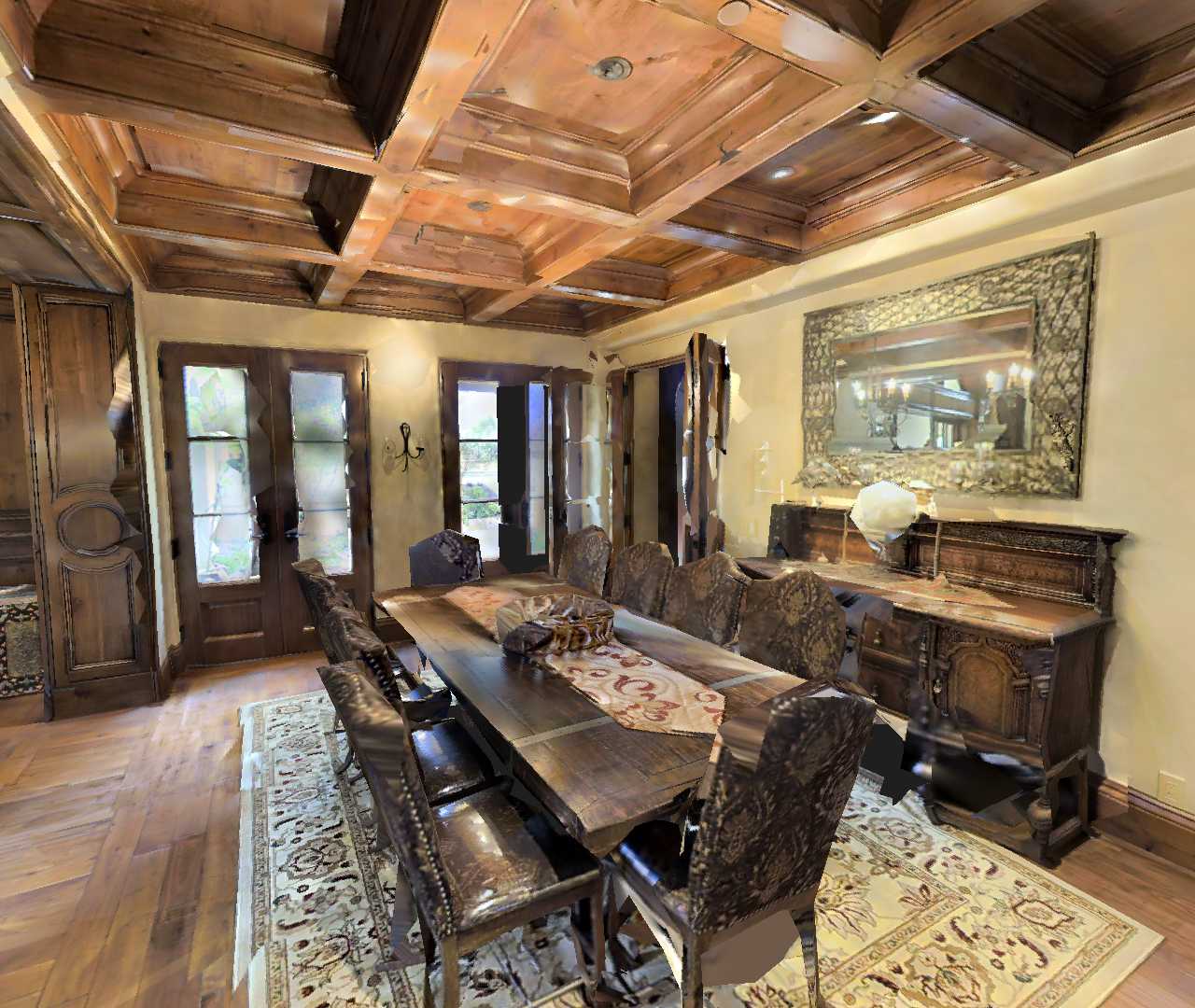}
\end{subfigure}
\begin{subfigure}{0.4\columnwidth}
  \centering
  \includegraphics[width=1\columnwidth, trim={8cm 0cm 10cm 17cm}, clip]{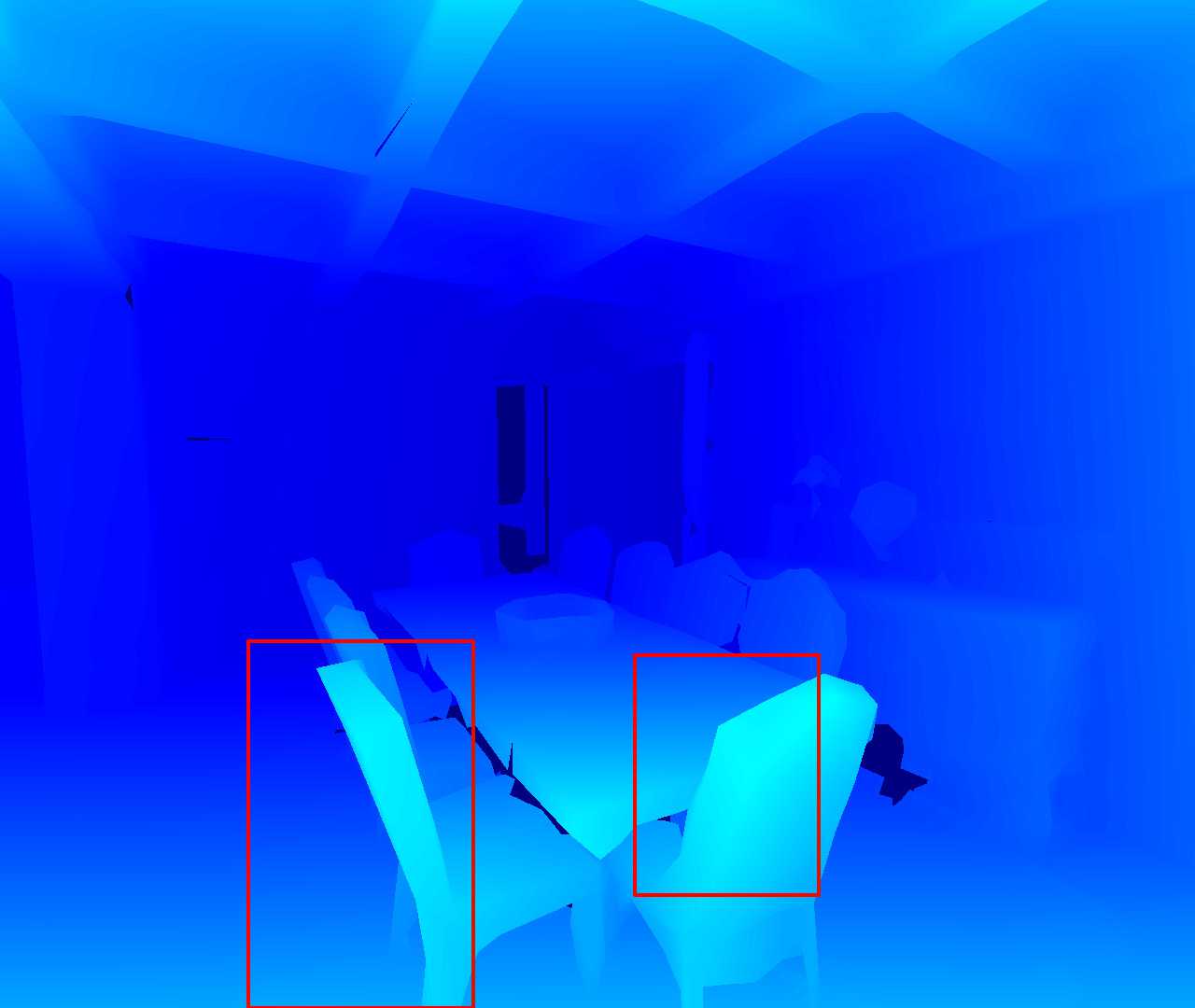}
\end{subfigure}
\begin{subfigure}{0.4\columnwidth}
  \centering
  \includegraphics[width=1\columnwidth, trim={8cm 0cm 10cm 17cm}, clip]{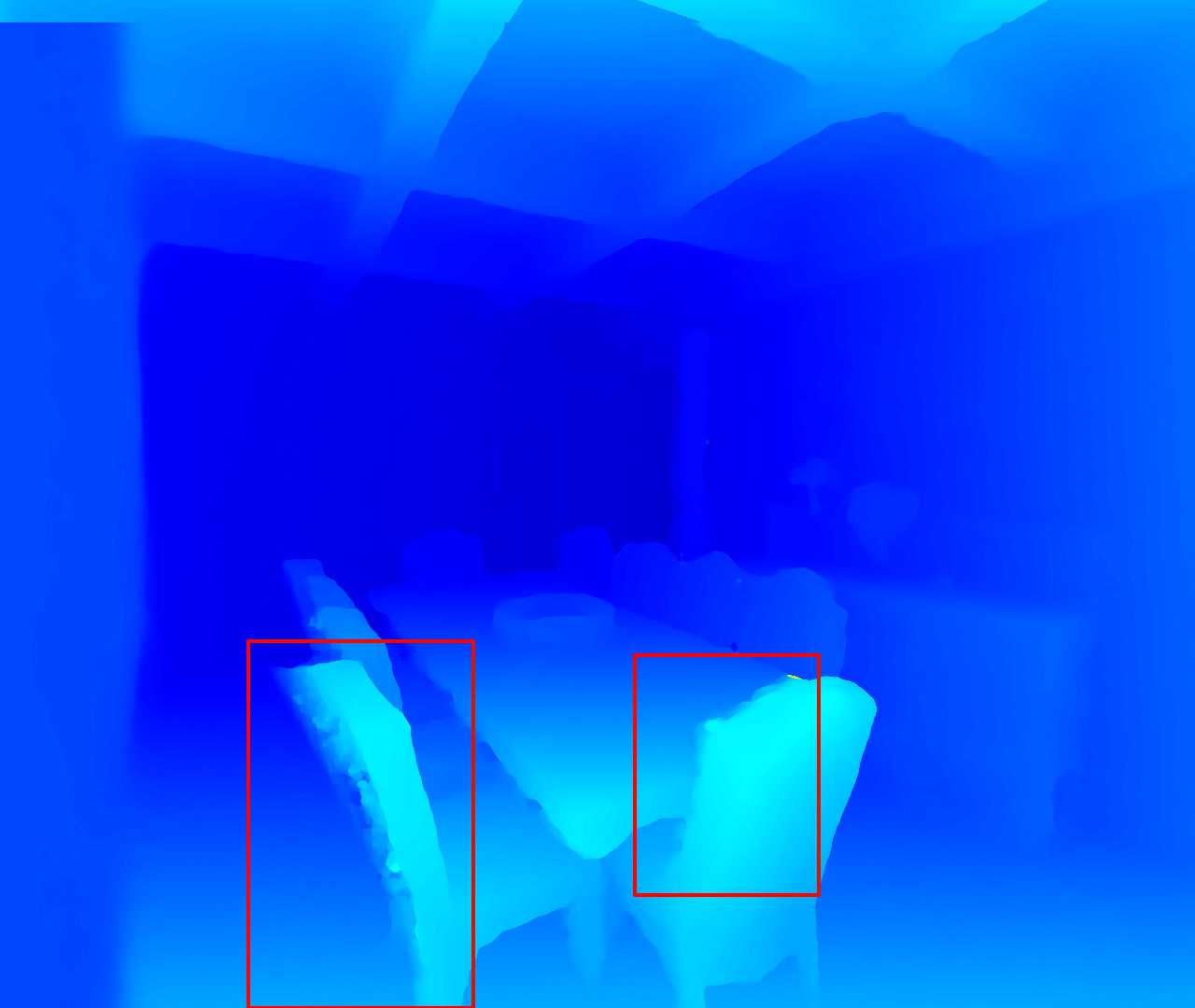}
  \caption*{Stereo}
\end{subfigure}
\begin{subfigure}{0.4\columnwidth}
  \centering
  \includegraphics[width=1\columnwidth, trim={8cm 0cm 10cm 17cm}, clip]{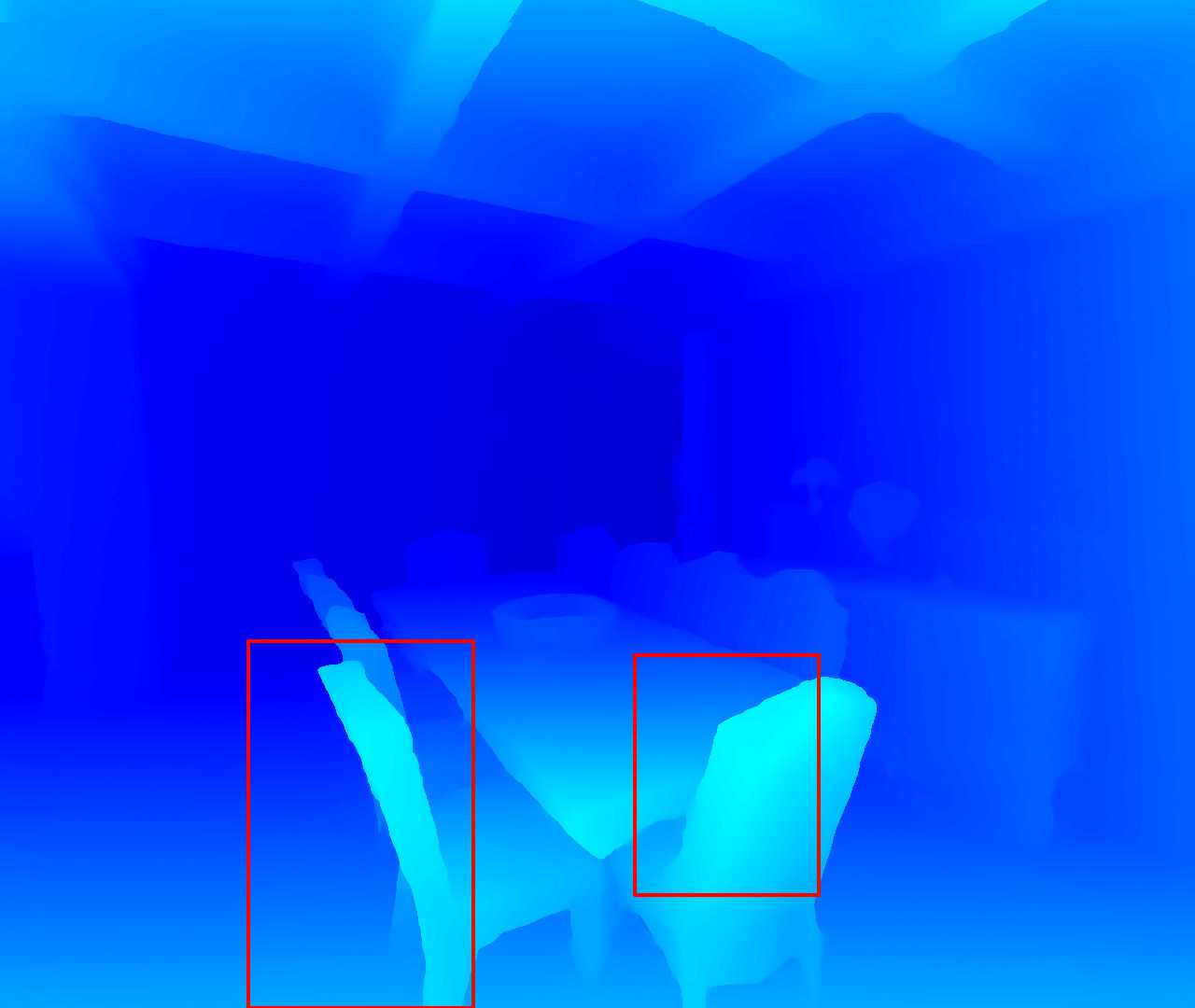}
  \caption*{Multiscopic}
\end{subfigure}
\caption{Comparison between stereo mode and our multiscopic mode. The first row displays the center view image ground-truth disparity. The second row displays the disparity map obtained by self-supervised learning using 2 images and that using 3 images. Occluded areas are much better with our multiscopic approach.}
\label{fig:res-ablation}
\vspace{-0.2cm}
\end{figure}

\setlength{\tabcolsep}{4pt}
\begin{table}[t]
\centering
\begin{tabular}{c c c c c c}
\toprule
{Image} & $\mathcal{L}_P$ & $\mathcal{L}_M$  & D1-all & D1-occ   \\
\midrule
$2$ & w/o cross & \xmark & $7.43\%$ & $91.66\%$ \\
$3$ & w/o cross & \xmark & $7.40\%$ & $91.54\%$ \\
$3$ & Cross & \xmark & $5.05\%$ & $38.71\%$ \\
$3$ & Cross & w/o uncertainty & $4.32\%$ & $15.29\%$ \\
$3$ & Cross & uncertainty-aware & $3.82\%$ & $10.87\%$ \\
\bottomrule
\end{tabular}
\caption{Ablation study on KITTI 2015. ``w/o cross" means only using $\mathcal{L}_P(I_{cl}^l, I_c)$, $\mathcal{L}_P(I_{cr}^r, I_c)$. ``w/o uncertainty" means only using the first equation of Equ.~\ref{equ:lossm} for $\mathcal{L}_M$. D1-occ is the D1 error of occluded pixels.}
\label{tab:ablation}
\vspace{-0.3cm}
\end{table}
\setlength{\tabcolsep}{4pt}

\subsection{Ablation Study}

To inspect the effect of our framework, we evaluate each module of our model on the KITTI 2015 training set, as displayed in Table~\ref{tab:ablation}. For the stereo self-supervised mode, we use only two images where the photometric loss is built by warping between the center image and the right image. Then we add modules proposed in our framework one by one.
The result shows that a better disparity map can be obtained with our cross-photometric loss as well as the uncertainty-aware mutual-supervision loss, especially in occluded areas. A qualitative comparison is shown in Fig.~\ref{fig:res-ablation}, from where we can see the occluded parts in stereo mode cannot be reconstructed correctly but are estimated much better in multiscopic mode. This is as expected. The loss $\mathcal{L}_P$ and $\mathcal{L}_M$ in multiscopic mode can both assist the estimation in occluded areas since they link the left and the right view, e.g., the invisible pixel in the left view would cause a large loss in $\mathcal{L}_M$ and $\mathcal{L}_P(I_{cr}^{l},I_c)$.

\section{Conclusion}

In this paper, we exploit the image shape of multiscopic images to learn an end-to-end network to estimate the disparity map without ground-truth depth information. 
In a multiscopic structure, the images are captured with parallel, co-planar, and same-parallax camera angles, such that the pixel disparities are the same between the center image and any surrounding image. Self-supervision is obtained by taking advantage of this constrain.
The cross photometric loss, uncertainty-aware mutual-supervision loss, and smoothness loss are introduced to guide self-supervised learning. 
As there is no public large multiscopic dataset, we build a new dataset composed of synthetic images rendered by 3D engines and real-world images captured by the multiscopic systems designed by ourselves. 
After being trained with only these indoor synthetic images, our network can obtain better disparity maps in the unseen outdoor KITTI dataset compared to previous unsupervised methods.

\balance



{\small
\bibliographystyle{IEEEtranN}
\bibliography{ref}
}
\end{document}


\maketitle




\section{Network Structure}

The structure of our network is based on PSMNet \cite{chang2018pyramid}. The network structure and parameters are displayed in Table~\ref{tab:network}.
We trim the model by reducing some layers for efficiency, and add an uncertainty head. The spatial pyramid pooling module and 3D convolution module incorporate both the local and global information to produce a cost volume, which is then converted to a probability volume to regress the final disparity prediction. The uncertainty head regresses an uncertainty map from the cost volume.

\makeatletter
\renewcommand*\env@matrix[1][\arraystretch]{%
  \edef\arraystretch{#1}%
  \hskip -\arraycolsep
  \let\@ifnextchar\new@ifnextchar
  \array{*\c@MaxMatrixCols c}}
\makeatother

\setlength{\tabcolsep}{8pt}
\begin{table}[]
\scriptsize
\centering
\renewcommand{\arraystretch}{1.6}
\begin{tabular}{ | c | c | c | }
\hline
Layer name & Layer structure & Output dimension \\
\hline
Input & & $H \times W \times 3$ \\
\hline
\multicolumn{3}{|c|}{Feature extraction} \\
\hline
Conv0 &
$ 3\times3, 32 $ &
$\frac12H \times \frac12 W \times 32$\\
\hline
Conv1 &
$\left[ 3\times3, 32 \right] * 3$ &
$\frac12H \times \frac12 W \times 32$\\
\hline
Conv2 &
$\left[ 3\times3, 64 \right] * 16$ &
$\frac14H \times \frac14 W \times 64$\\
\hline
Conv3 &
$\left[ 3\times3, 128 \right] * 3, \text{dila}=2$&
$\frac14H \times \frac14 W \times 128$\\
\hline
Conv4 &
$\left[ 3\times3, 128 \right] * 3, \text{dila}=4$&
$\frac14H \times \frac14 W \times 128$\\
\hline
\multicolumn{3}{|c|}{Spatial pyramid pooling} \\
\hline
Branch1 &
$\begin{matrix}[1.2] 64 \times 64 \ \text{avg. pool} \\ 3\times 3, 32 \end{matrix}$ &
$\frac14H \times \frac14 W \times 32$\\
\hline
Branch2 &
$\begin{matrix}[1.2] 32 \times 32 \ \text{avg. pool} \\ 3\times 3, 32 \end{matrix}$ &
$\frac14H \times \frac14 W \times 32$\\
\hline
Branch3 &
$\begin{matrix}[1.2] 16 \times 16 \ \text{avg. pool} \\ 3\times 3, 32 \end{matrix}$ &
$\frac14H \times \frac14 W \times 32$\\
\hline
Branch4 &
$\begin{matrix}[1.2] 8 \times 8 \ \text{avg. pool} \\ 3\times 3, 32 \end{matrix}$ &
$\frac14H \times \frac14 W \times 32$\\
\hline
\multicolumn{2}{|c|}{
Concat $\begin{bmatrix}[1.4] \text{Conv}2, \text{Conv}4, \text{Branch}1, \\ \text{Branch}2, \text{Branch}3, \text{Branch}4 \end{bmatrix}$
} &
$\frac14H \times \frac14 W \times 320$\\
\hline
Fusion &
$ \begin{matrix}[1.2] 3\times3, 128 \\ 1\times1, 32 \end{matrix} $ &
$\frac14H \times \frac14 W \times 32$\\
\hline
\multicolumn{2}{|c|}{Concat left and shifted right} &
$\frac14D \times \frac14H \times \frac14 W \times 32$\\
\hline
\multicolumn{3}{|c|}{3D convolution} \\
\hline
3Dconv &
$\left[ 3\times3\times3, 32 \right] * 4$ &
$\frac14D \times \frac14H \times \frac14W \times 32$\\
\hline
3Dstack1 &
$\begin{matrix}[1.2]
[3\times3\times3, 64 ] * 4 \\
\text{deconv}\ 3\times3\times3, 64 \\
\text{deconv}\ 3\times3\times3, 32 \\
\end{matrix}$ &
$\frac1{4}D \times \frac1{4}H \times \frac1{4}W \times 32$\\
\hline
Output1 &
$\begin{matrix}[1.2]
3\times3\times3, 32\\
3\times3\times3, 1\\
\end{matrix}$ &
$\frac1{4}D \times \frac1{4}H \times \frac1{4}W \times 1$\\
\hline
3Dstack2 &
$\begin{matrix}[1.2]
[3\times3\times3, 64 ] * 4 \\
\text{deconv}\ 3\times3\times3, 64 \\
\text{deconv}\ 3\times3\times3, 32 \\
\end{matrix}$ &
$\frac1{4}D \times \frac1{4}H \times \frac1{4}W \times 32$\\
\hline
Output2 &
$\begin{matrix}[1.2]
3\times3\times3, 32\\
3\times3\times3, 1\\
\text{add Output}1
\end{matrix}$ &
$\frac1{4}D \times \frac1{4}H \times \frac1{4}W \times 1$\\
\hline
Upsampling & Bilinear interpolation & $D \times H \times W$\\
\hline
Disparity & Disparity regression & $ H \times W $\\
\hline
Uncertainty & $3 \times 3, 1$ & $H \times W$ \\
\hline
\end{tabular}
\caption{Network structure.}
\label{tab:network}
\end{table}
\setlength{\tabcolsep}{8pt}

\section{Dataset Examples}

\begin{figure*}[]
\centering
\begin{subfigure}{1\columnwidth}
  \centering
  \includegraphics[width=1\columnwidth, trim={0cm 0cm 0cm 0cm}, clip]{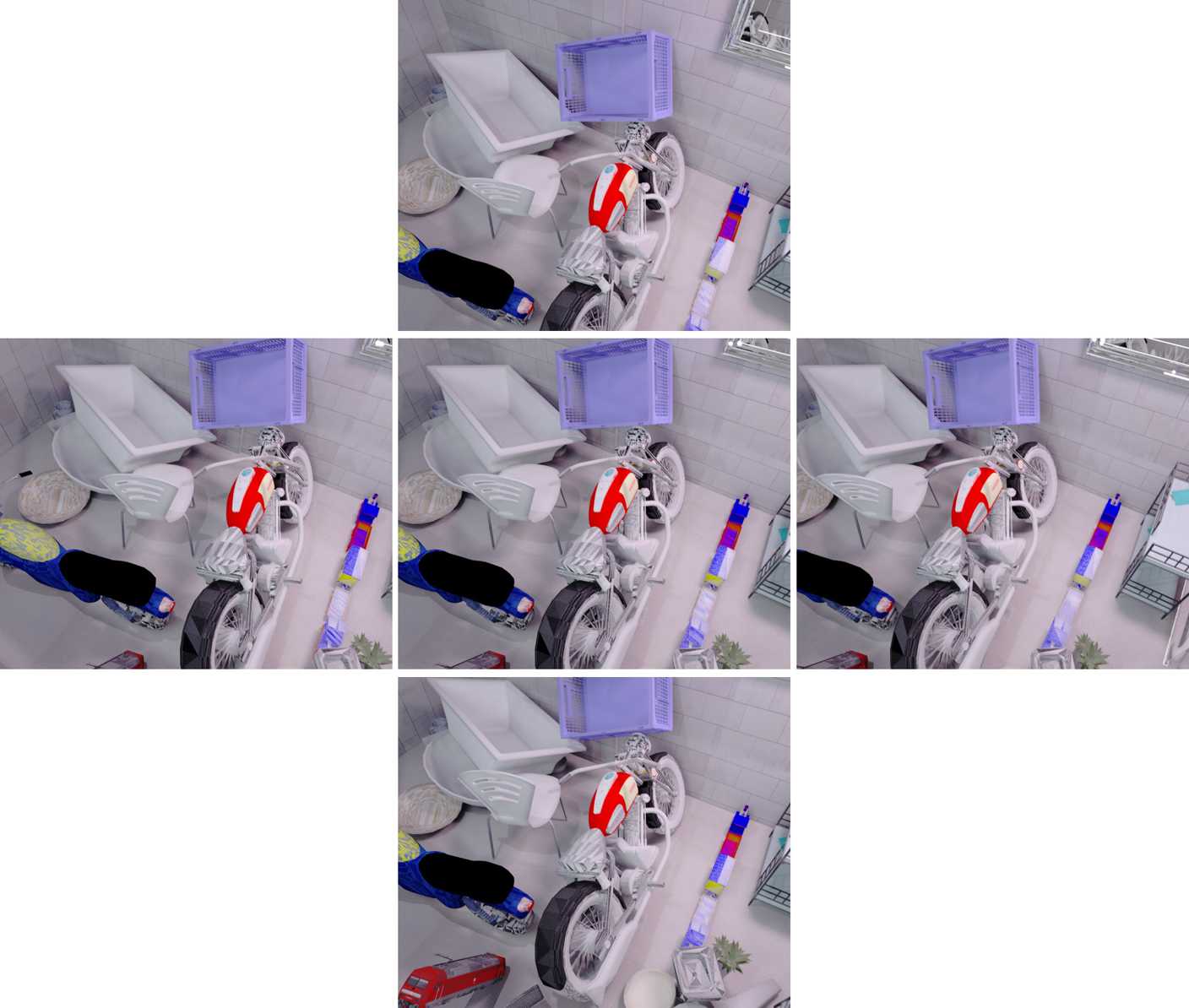}
\end{subfigure}
\begin{subfigure}{1\columnwidth}
  \centering
  \includegraphics[width=1\columnwidth, trim={0cm 0cm 0cm 0cm}, clip]{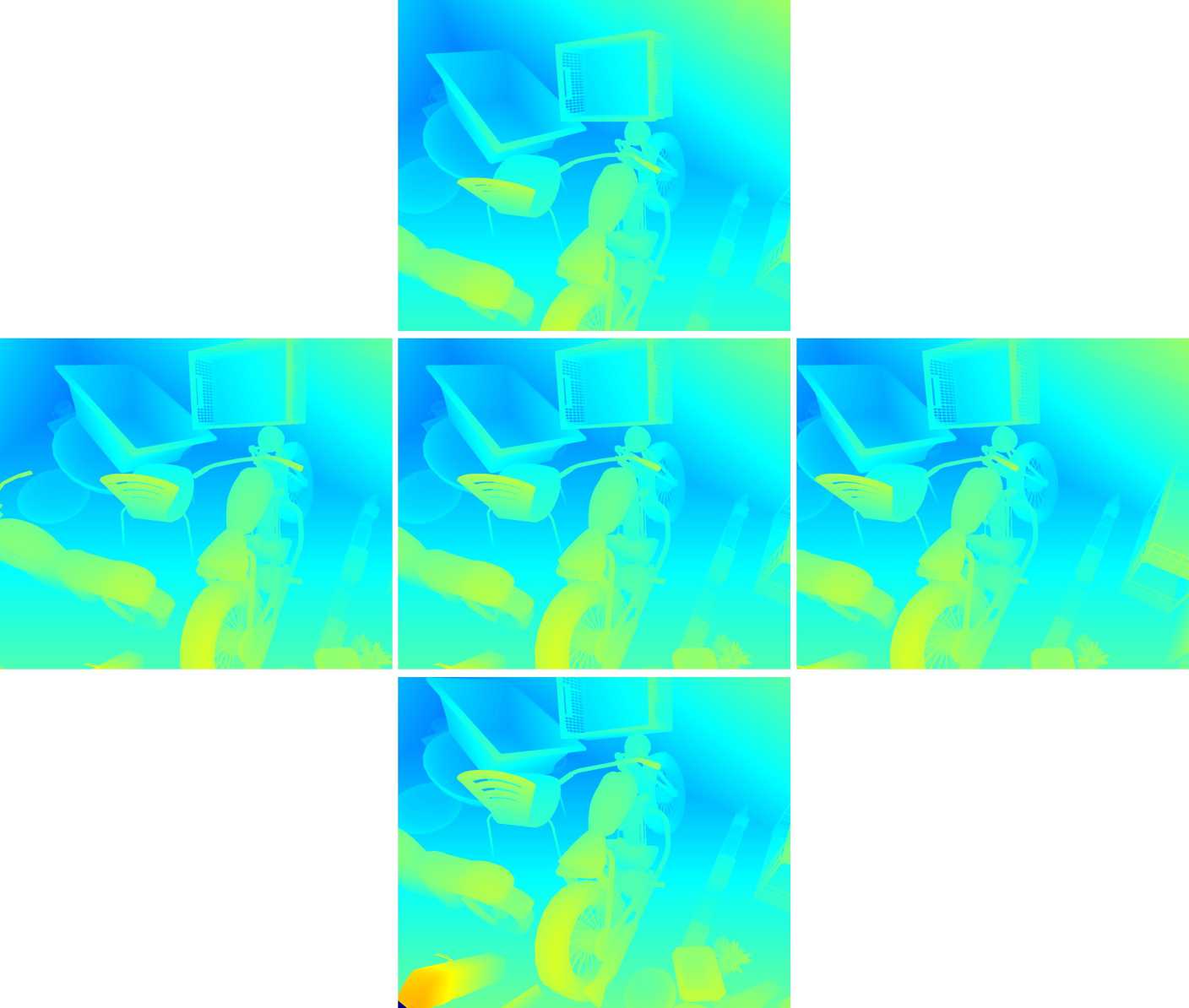}
\end{subfigure}

\begin{subfigure}{1\columnwidth}
  \centering
  \includegraphics[width=1\columnwidth, trim={0cm 0cm 0cm 0cm}, clip]{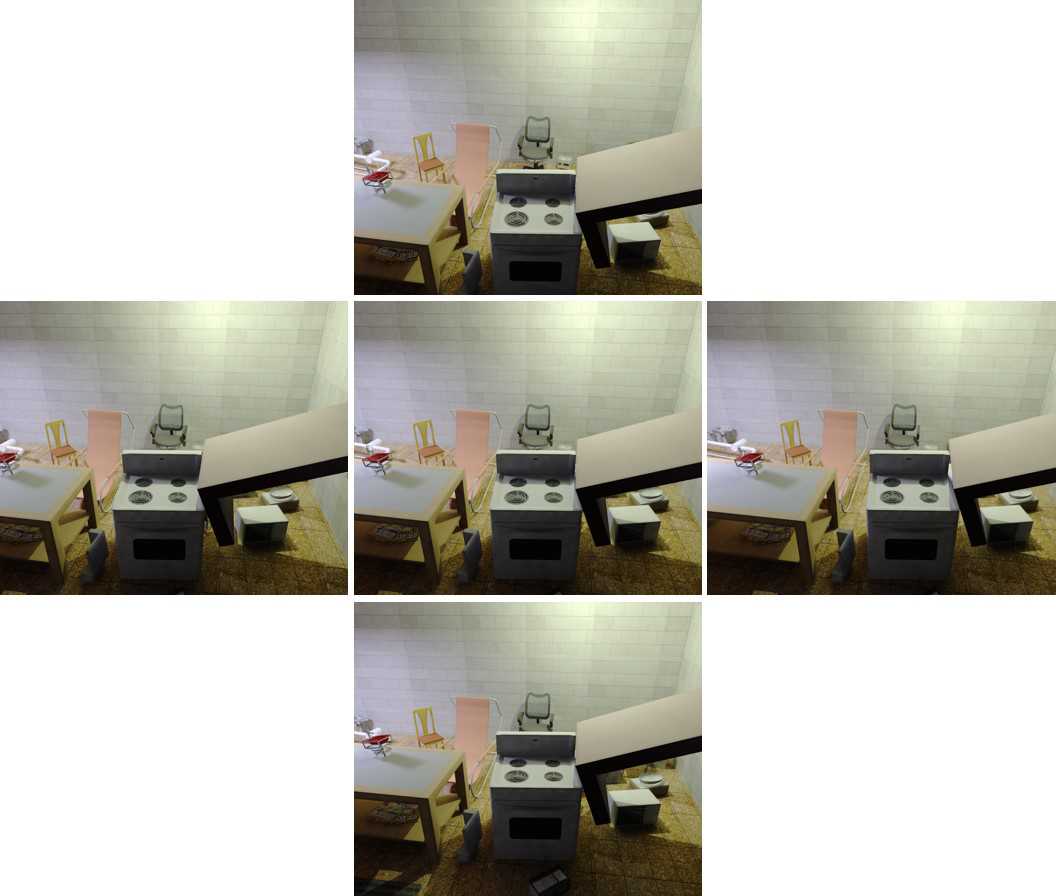}
\end{subfigure}
\begin{subfigure}{1\columnwidth}
  \centering
  \includegraphics[width=1\columnwidth, trim={0cm 0cm 0cm 0cm}, clip]{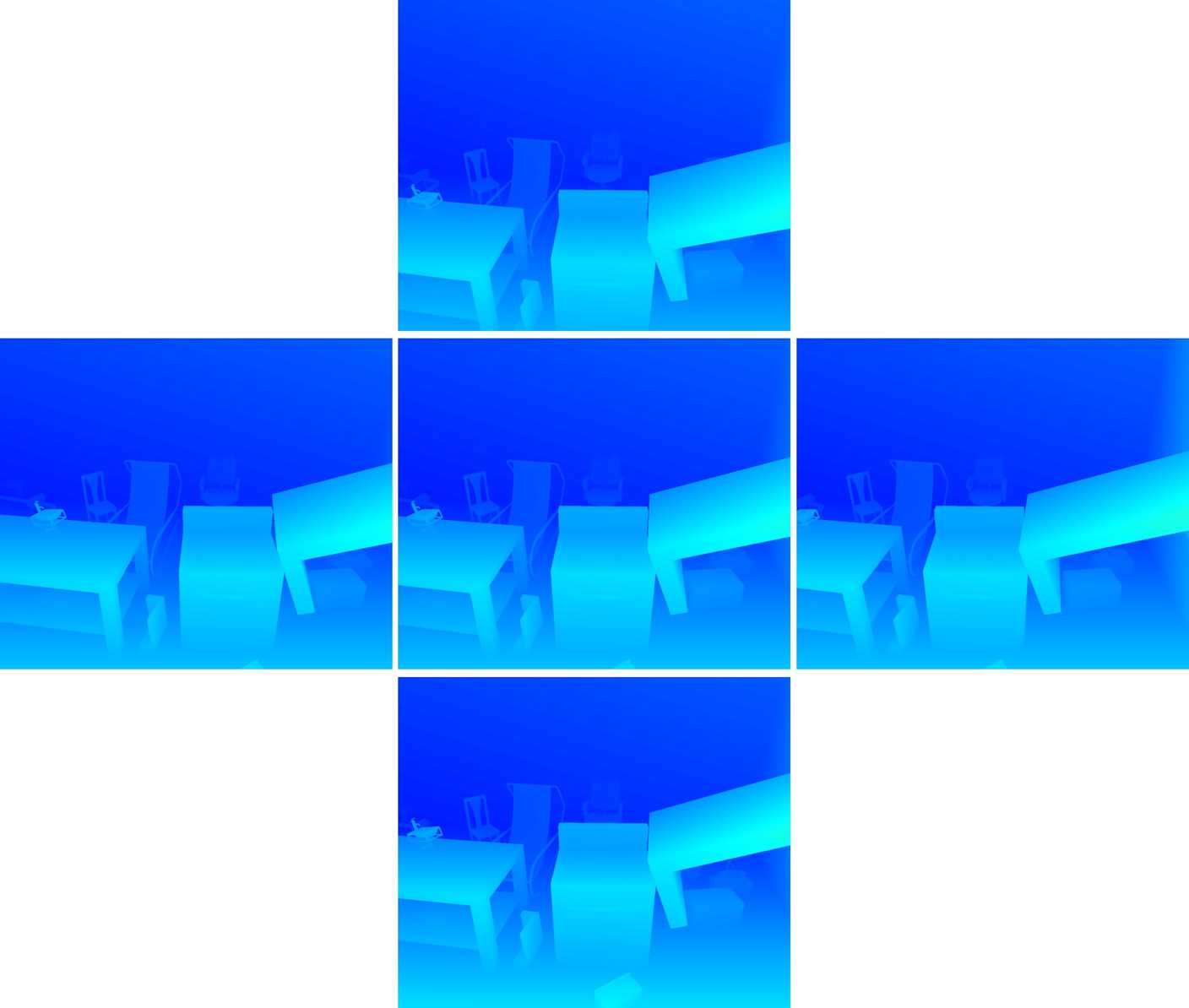}
\end{subfigure}

\begin{subfigure}{1\columnwidth}
  \centering
  \includegraphics[width=1\columnwidth, trim={0cm 0cm 0cm 0cm}, clip]{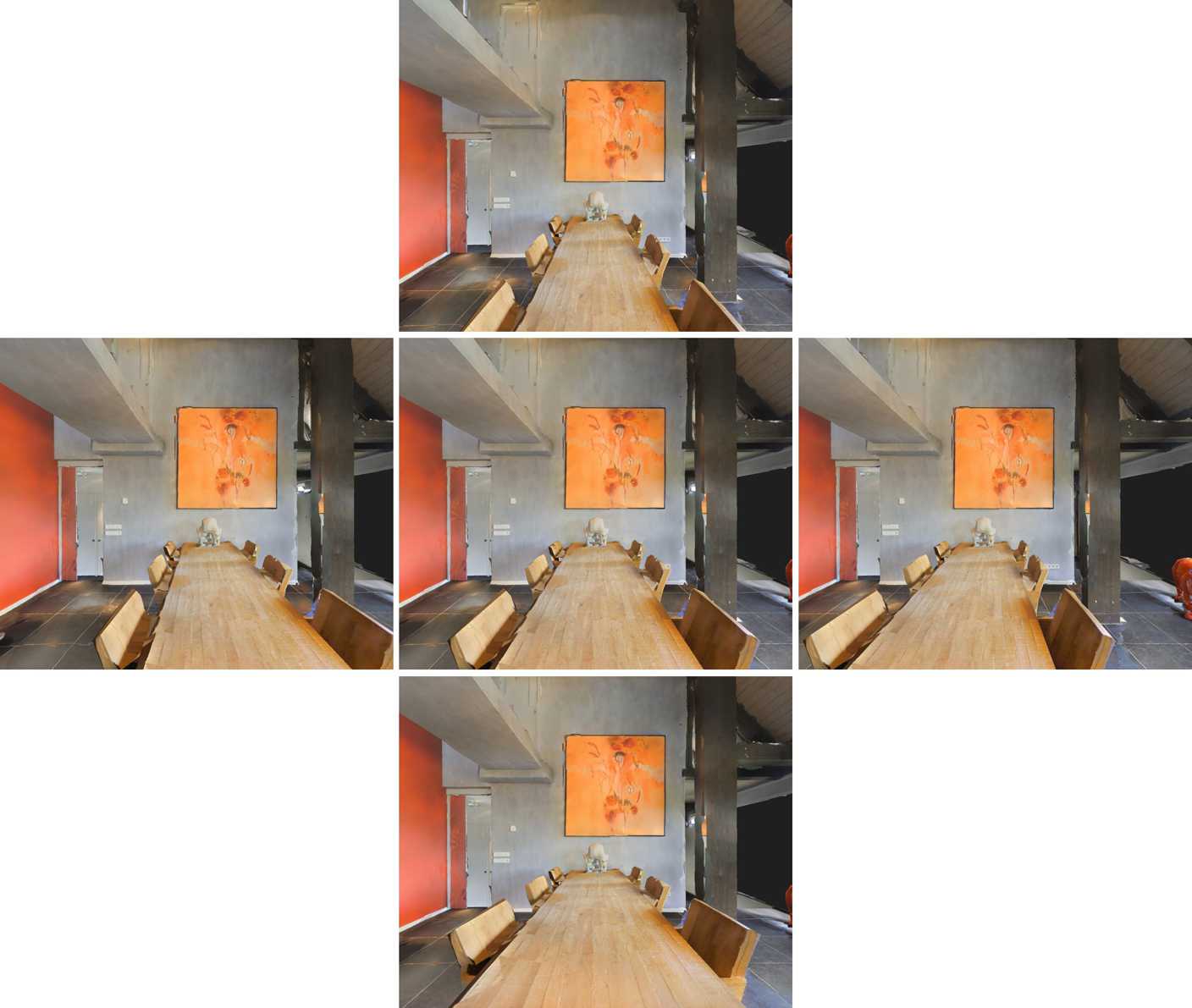}
\end{subfigure}
\begin{subfigure}{1\columnwidth}
  \centering
  \includegraphics[width=1\columnwidth, trim={0cm 0cm 0cm 0cm}, clip]{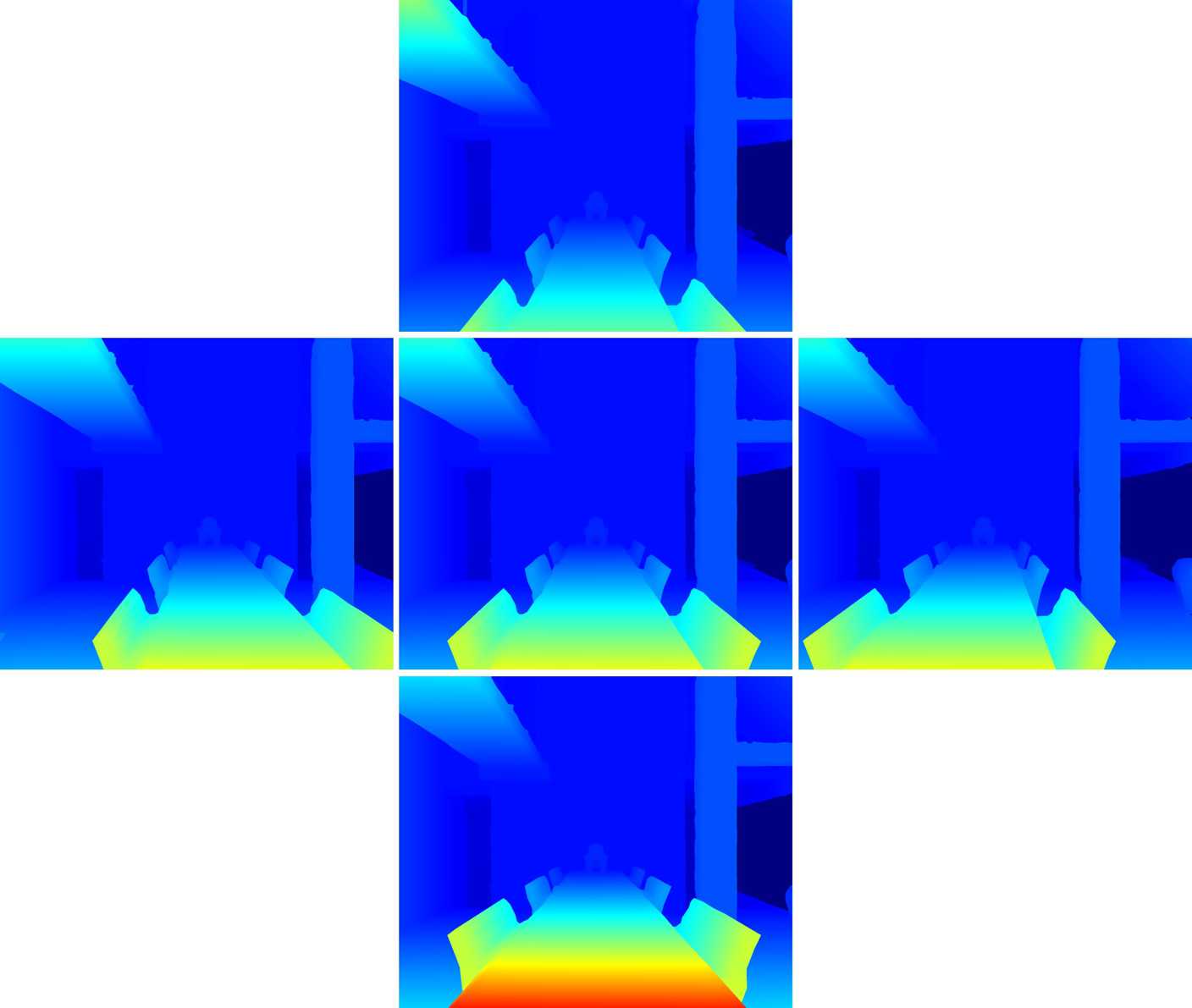}
\end{subfigure}
\caption{Example scenes of synthetic multiscopic images in our dataset. The first two rows are rendered by SceneNet RGB-D, and the third row is rendered by Habitat-sim.}
\label{fig:app-dataset}
\end{figure*}

\begin{figure*}[]
\centering
\begin{subfigure}{1\columnwidth}
  \centering
  \includegraphics[width=1\columnwidth, trim={0cm 0cm 0cm 0cm}, clip]{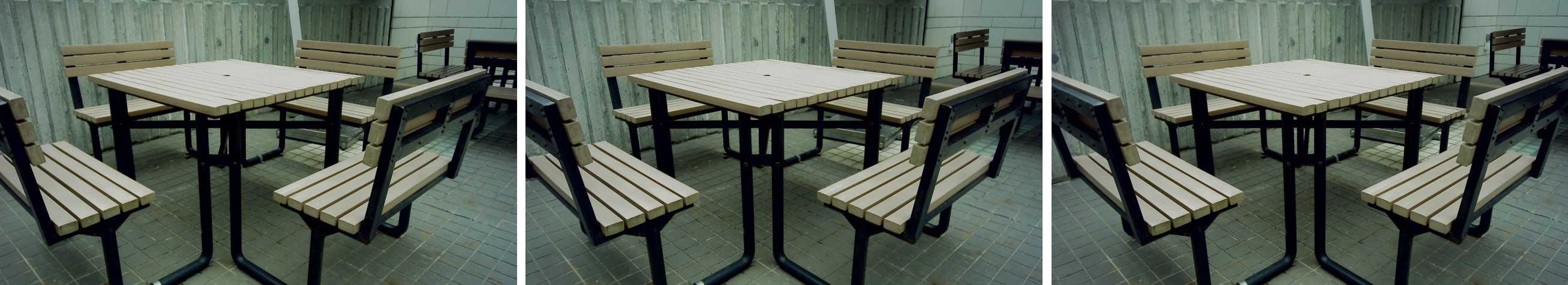}
\end{subfigure}
\begin{subfigure}{1\columnwidth}
  \centering
  \includegraphics[width=1\columnwidth, trim={0cm 0cm 0cm 0cm}, clip]{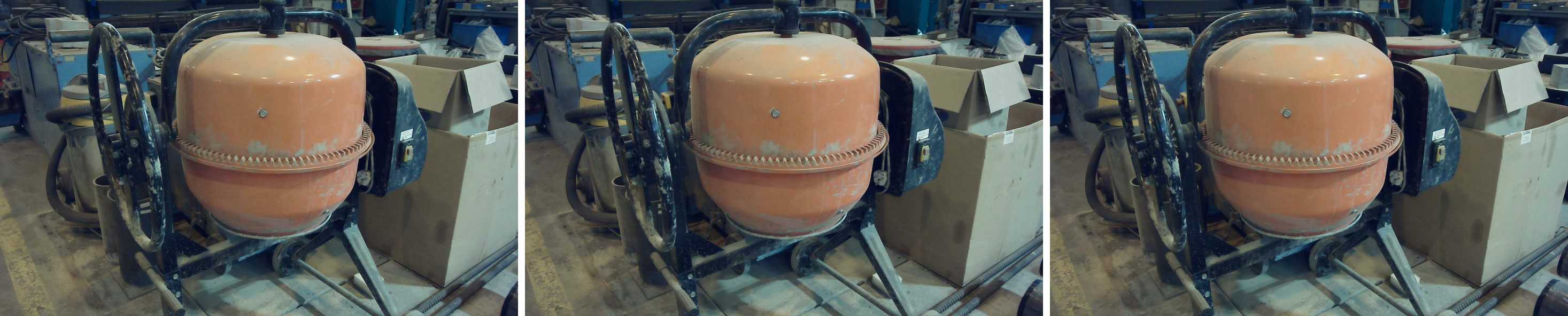}
\end{subfigure}
\begin{subfigure}{1\columnwidth}
  \centering
  \includegraphics[width=1\columnwidth, trim={0cm 0cm 0cm 0cm}, clip]{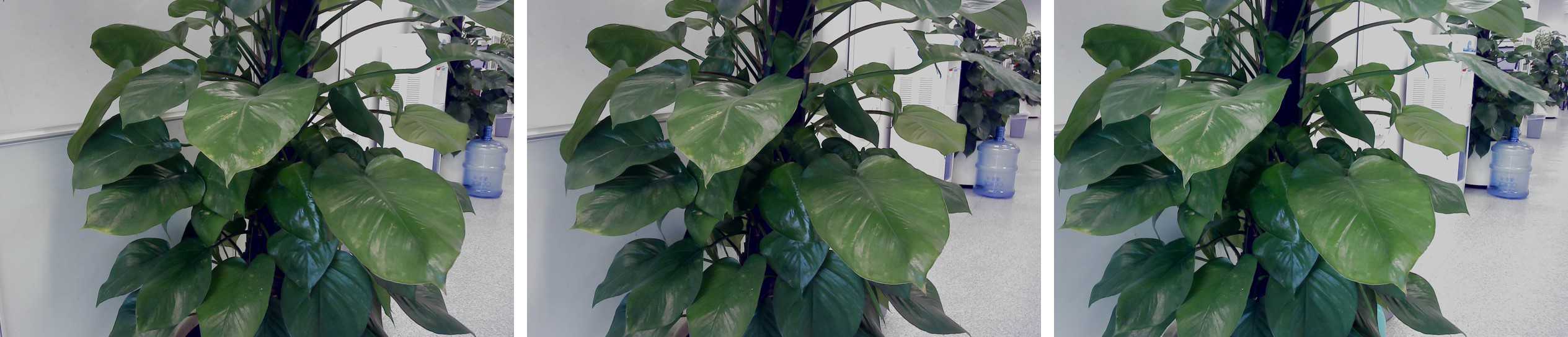}
\end{subfigure}
\begin{subfigure}{1\columnwidth}
  \centering
  \includegraphics[width=1\columnwidth, trim={0cm 0cm 0cm 0cm}, clip]{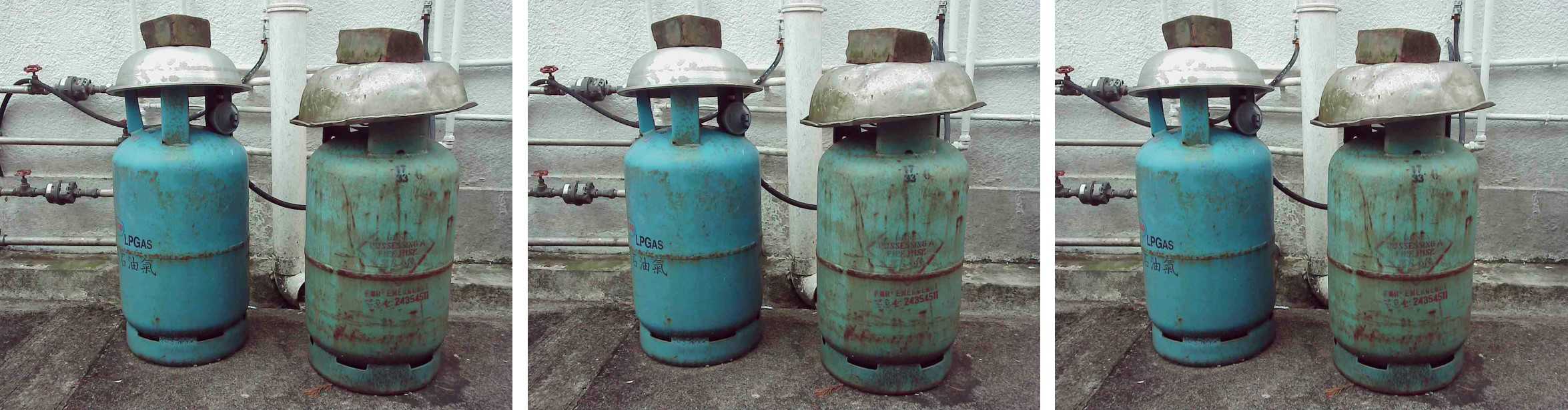}
\end{subfigure}
\caption{Example scenes of real multiscopic images in our dataset obtained by real cameras.}
\label{fig:app-dataset2}
\end{figure*}

In Fig.~\ref{fig:app-dataset}, we present more scenes of synthetic multiscopic images in our dataset, and in Fig.~\ref{fig:app-dataset2} we show real images obtained by our multiscopic system.

\section{More Qualitative Results}

\begin{figure*}[]
\centering
\begin{subfigure}{0.48\columnwidth}
  \centering
  \includegraphics[width=1\columnwidth, trim={0cm 0cm 0cm 0cm}, clip]{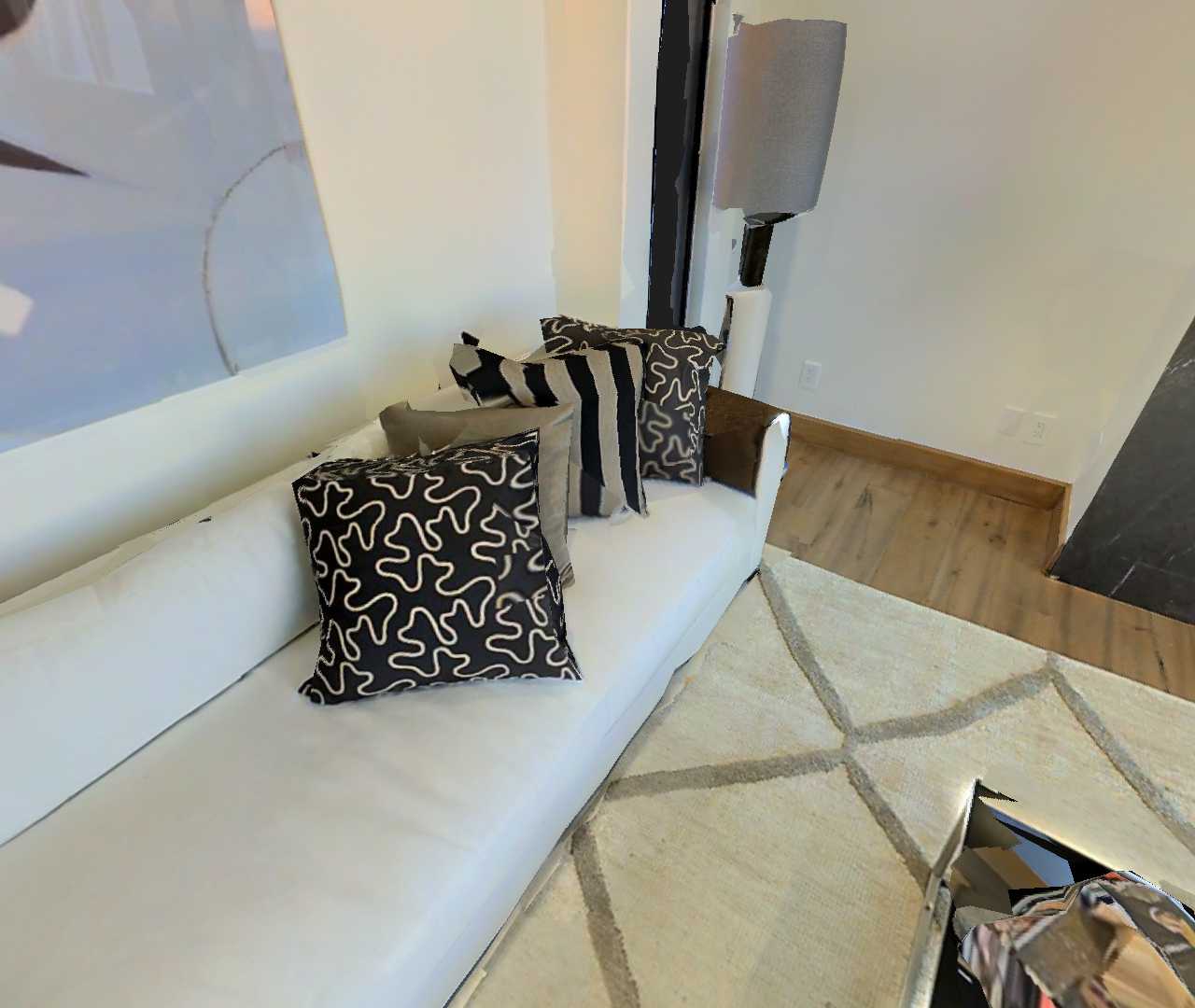}
\end{subfigure}
\begin{subfigure}{0.48\columnwidth}
  \centering
  \includegraphics[width=1\columnwidth, trim={0cm 0cm 0cm 0cm}, clip]{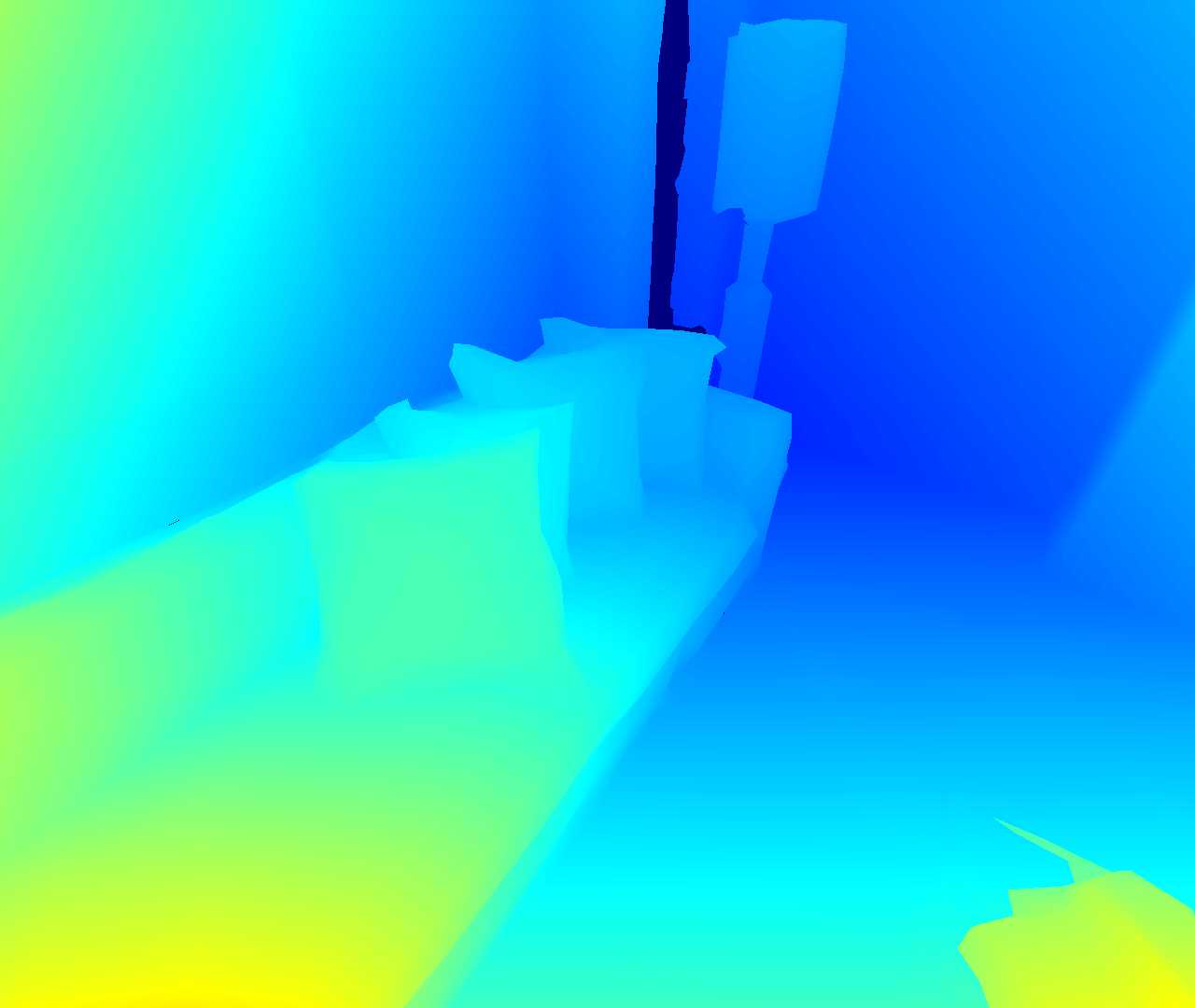}
\end{subfigure}
\begin{subfigure}{0.48\columnwidth}
  \centering
  \includegraphics[width=1\columnwidth, trim={0cm 0cm 0cm 0cm}, clip]{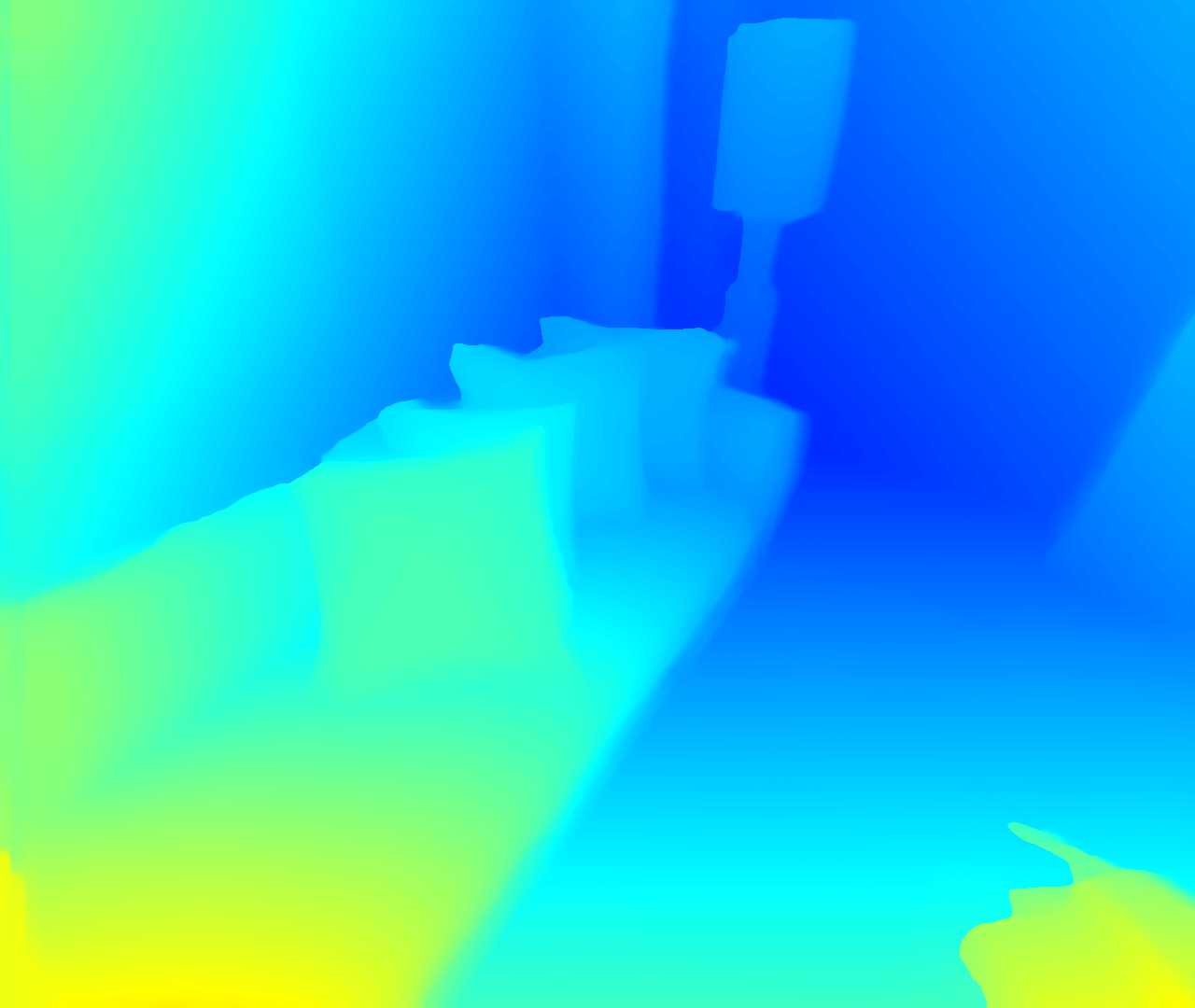}
\end{subfigure}
\begin{subfigure}{0.48\columnwidth}
  \centering
  \includegraphics[width=1\columnwidth, trim={0cm 0cm 0cm 0cm}, clip]{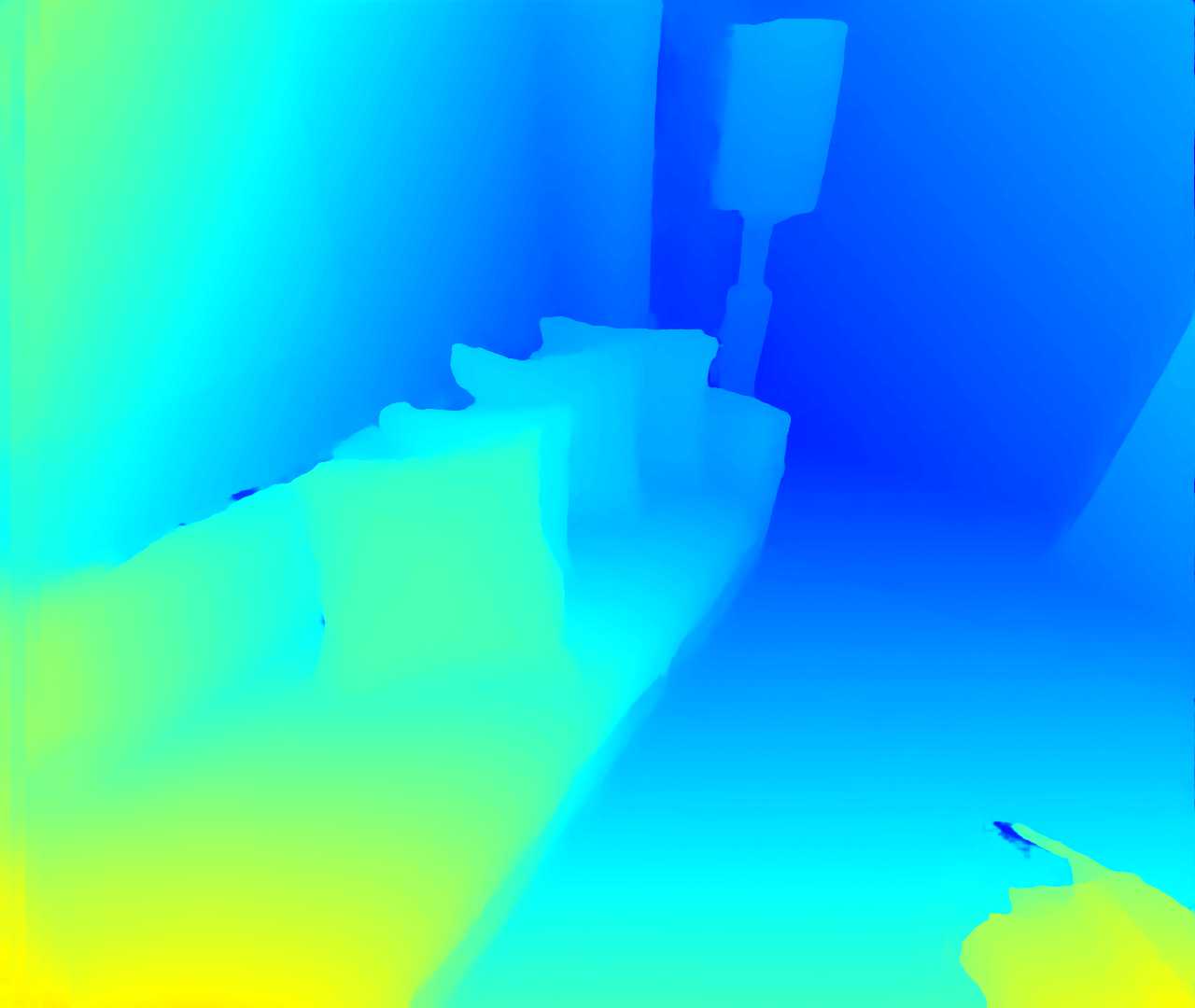}
\end{subfigure}

\begin{subfigure}{0.48\columnwidth}
  \centering
  \includegraphics[width=1\columnwidth, trim={0cm 0cm 0cm 0cm}, clip]{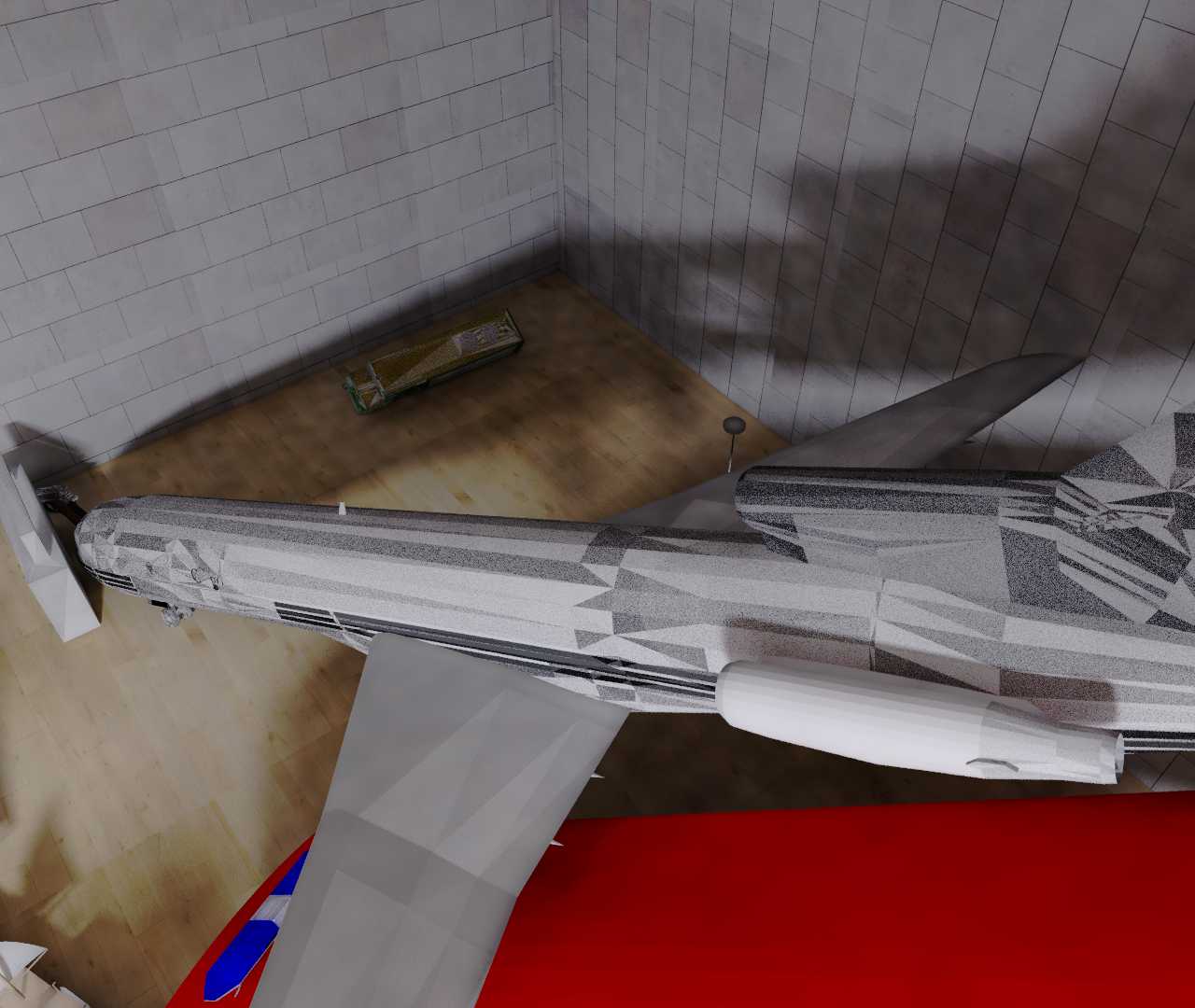}
\end{subfigure}
\begin{subfigure}{0.48\columnwidth}
  \centering
  \includegraphics[width=1\columnwidth, trim={0cm 0cm 0cm 0cm}, clip]{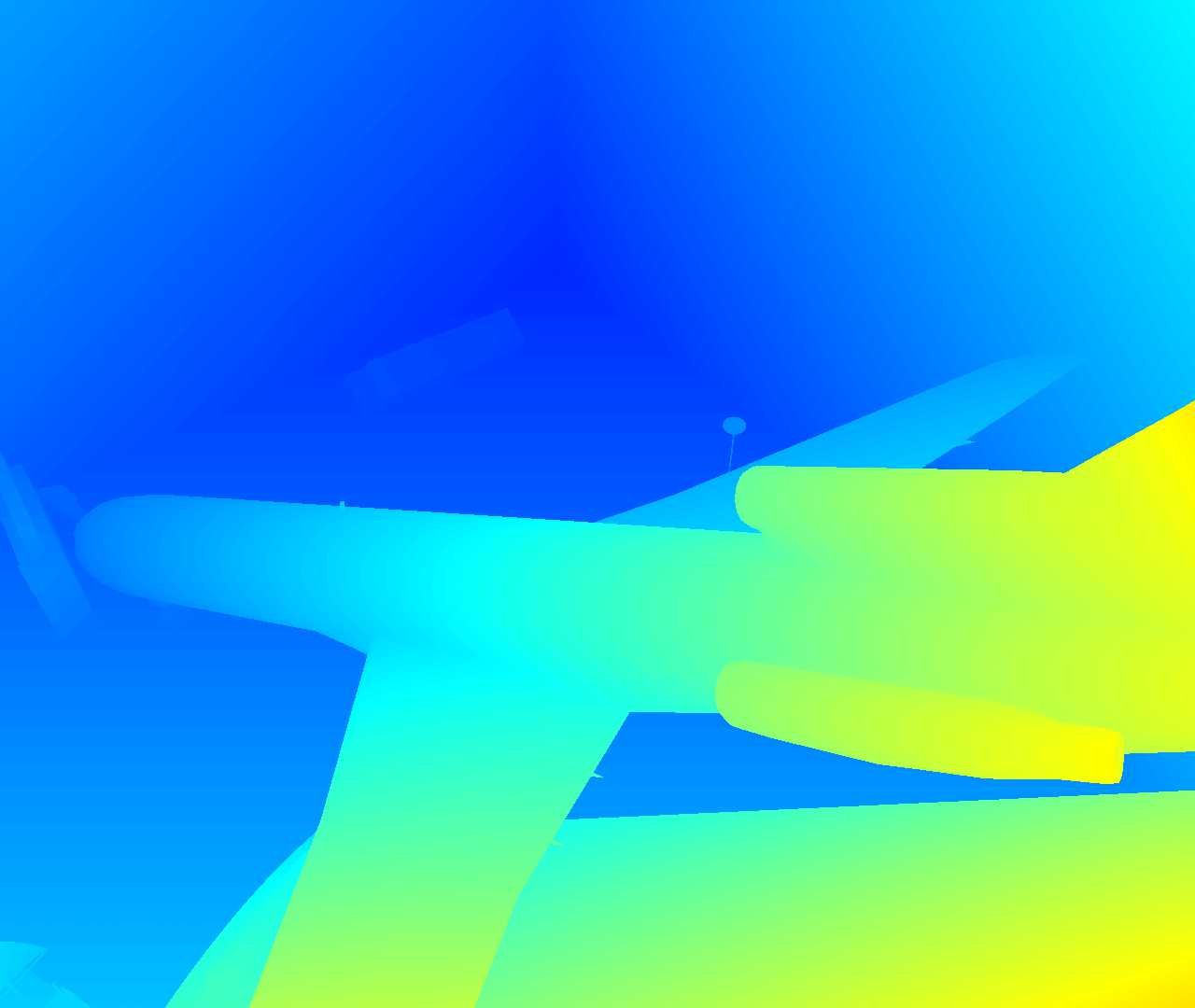}
\end{subfigure}
\begin{subfigure}{0.48\columnwidth}
  \centering
  \includegraphics[width=1\columnwidth, trim={0cm 0cm 0cm 0cm}, clip]{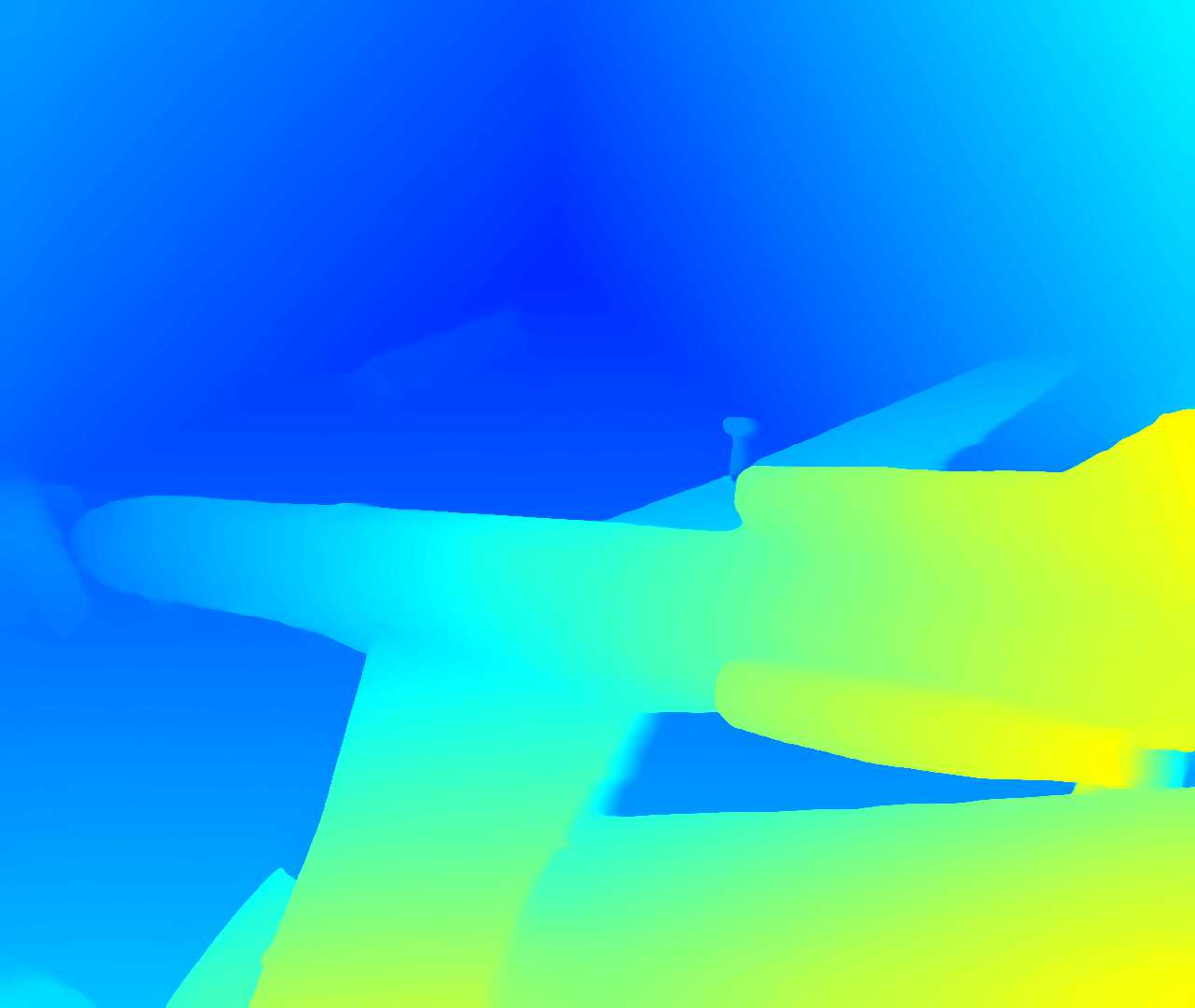}
\end{subfigure}
\begin{subfigure}{0.48\columnwidth}
  \centering
  \includegraphics[width=1\columnwidth, trim={0cm 0cm 0cm 0cm}, clip]{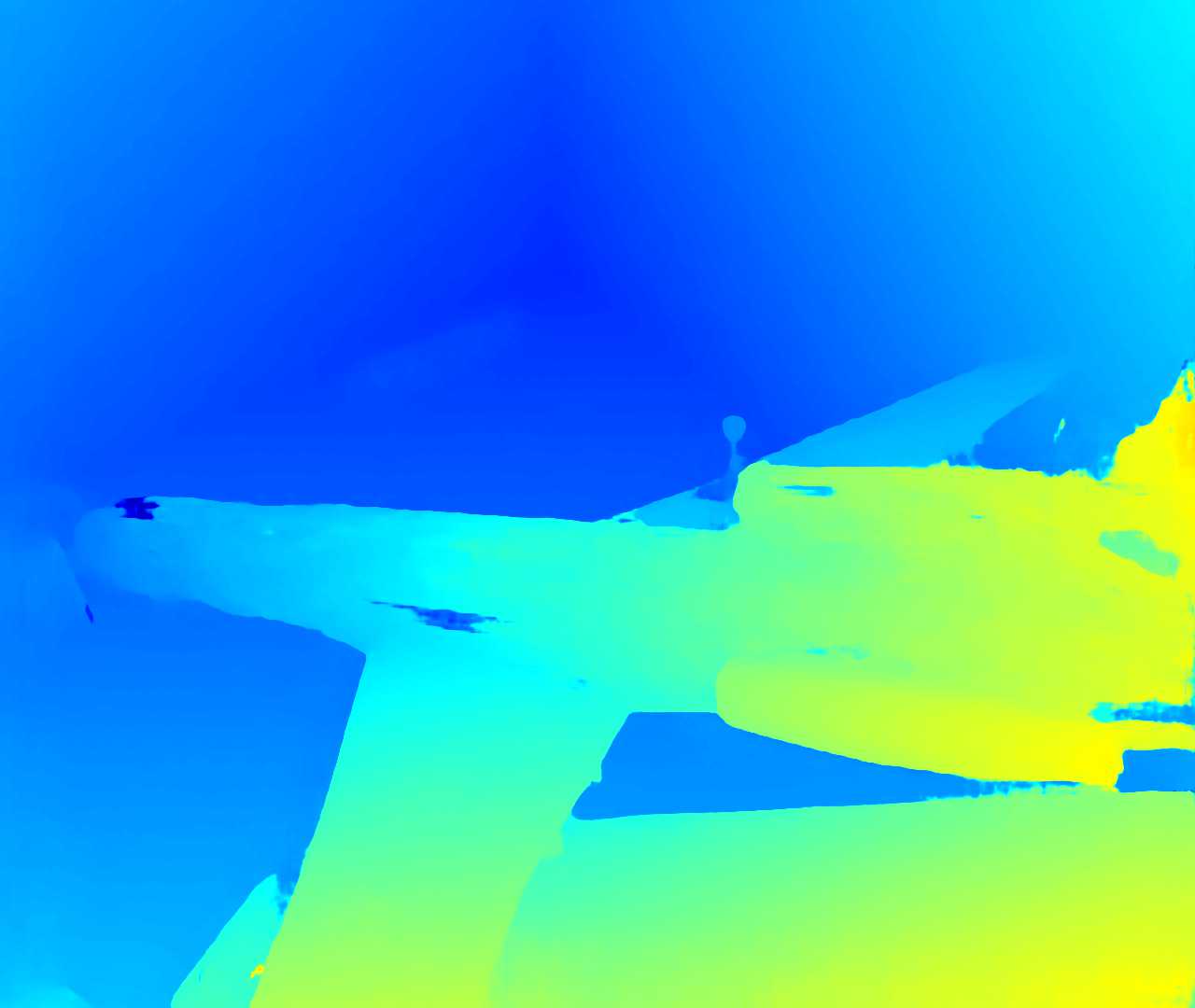}
\end{subfigure}

\begin{subfigure}{0.48\columnwidth}
  \centering
  \includegraphics[width=1\columnwidth, trim={0cm 0cm 0cm 0cm}, clip]{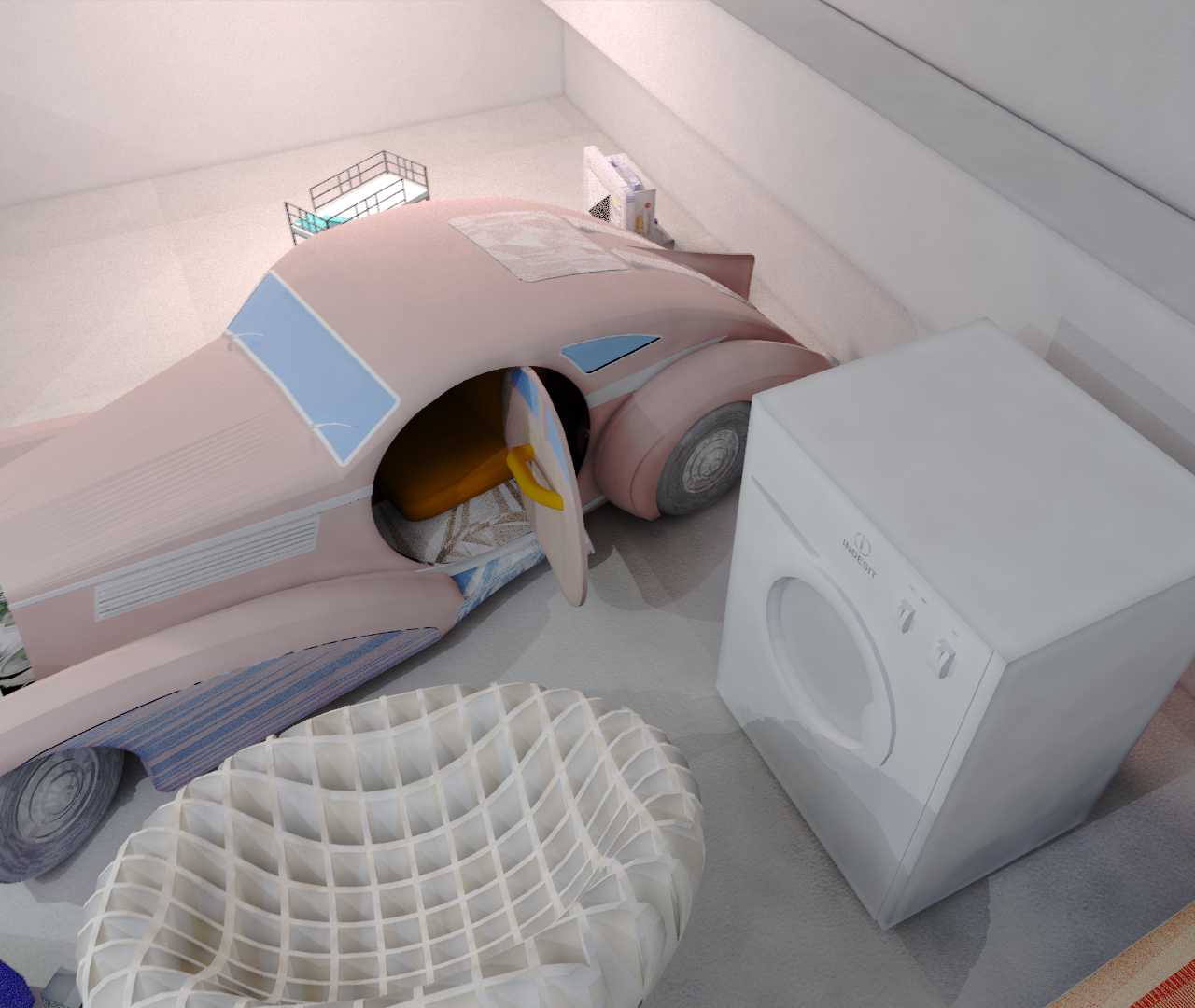}
\end{subfigure}
\begin{subfigure}{0.48\columnwidth}
  \centering
  \includegraphics[width=1\columnwidth, trim={0cm 0cm 0cm 0cm}, clip]{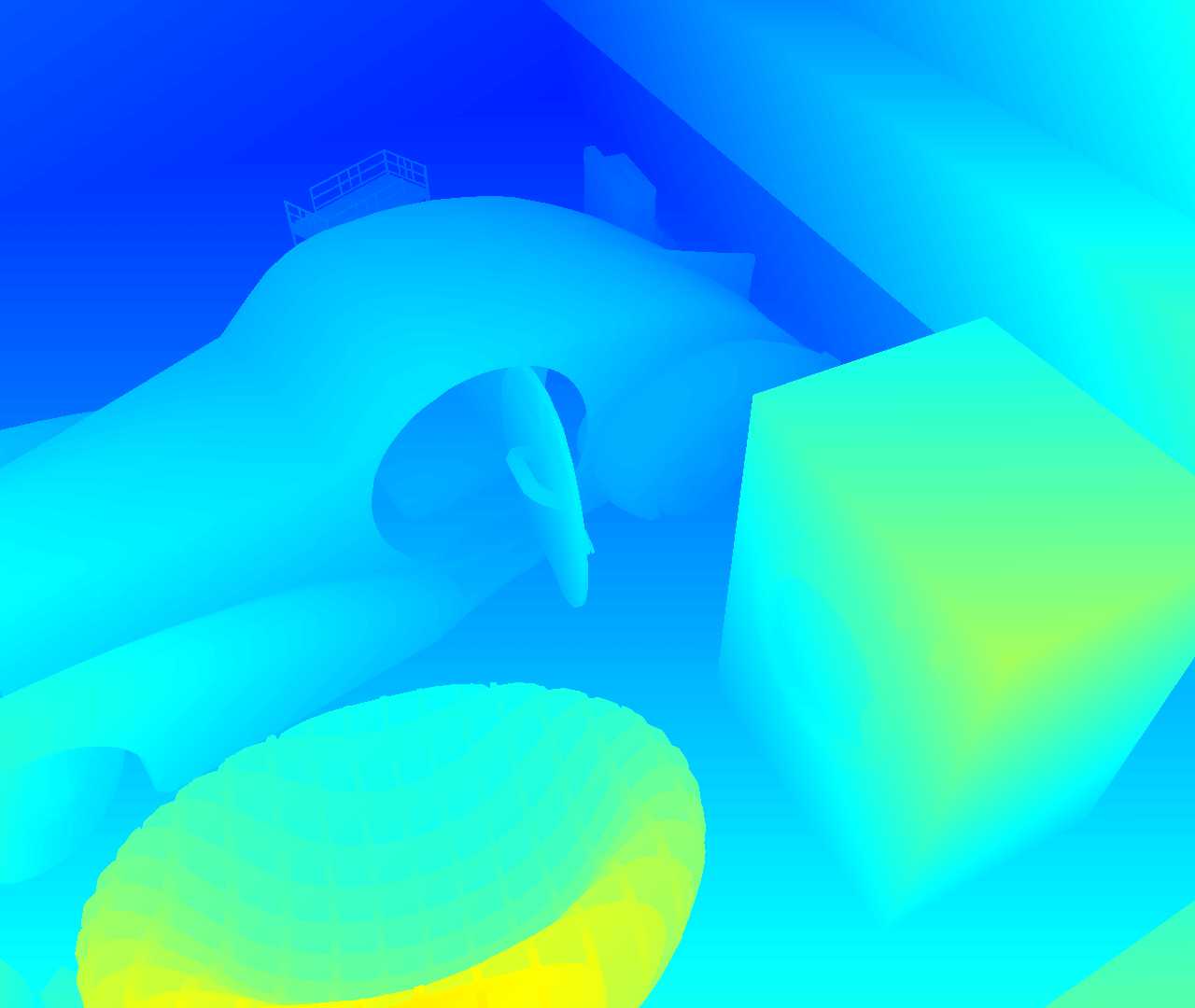}
\end{subfigure}
\begin{subfigure}{0.48\columnwidth}
  \centering
  \includegraphics[width=1\columnwidth, trim={0cm 0cm 0cm 0cm}, clip]{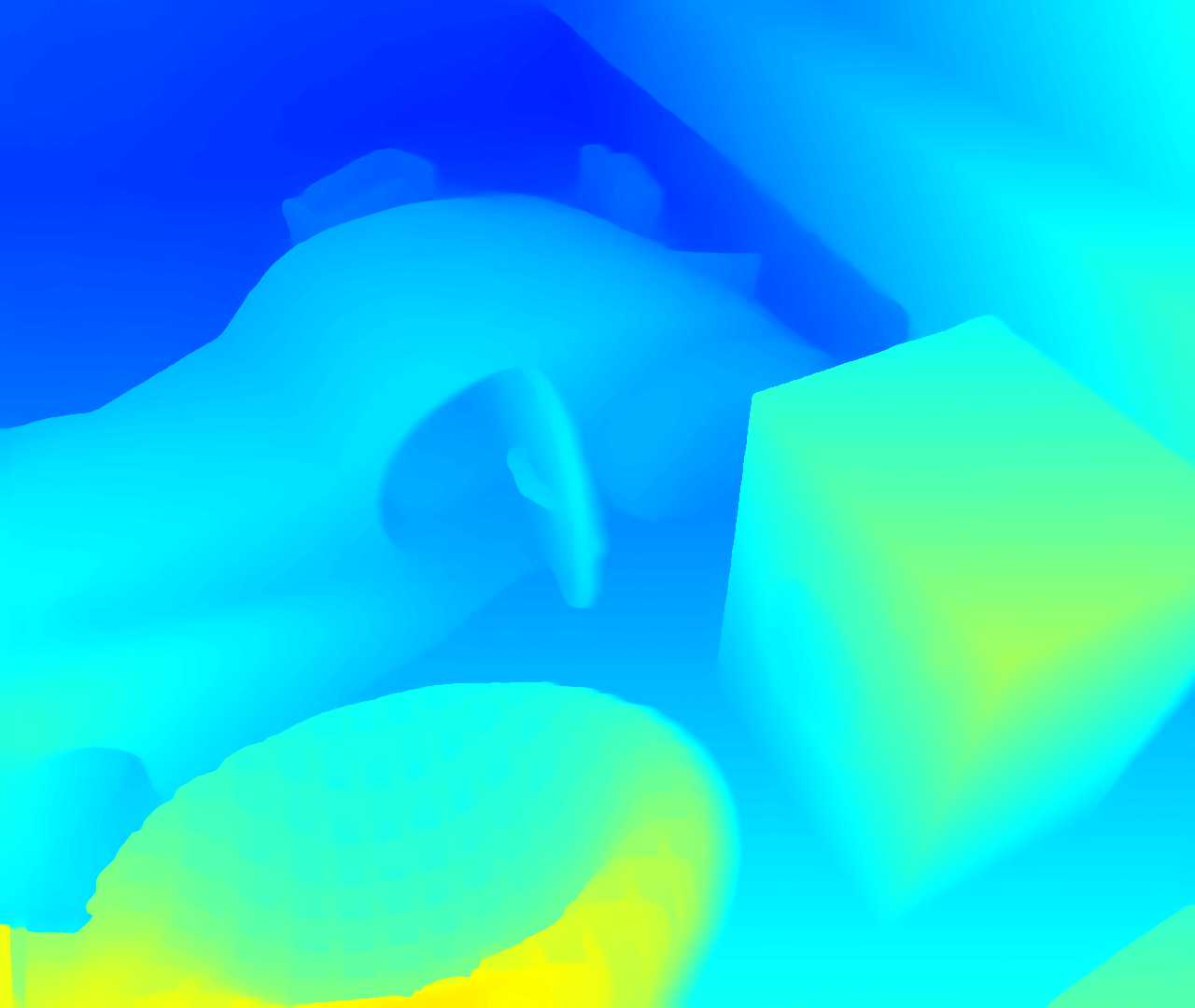}
\end{subfigure}
\begin{subfigure}{0.48\columnwidth}
  \centering
  \includegraphics[width=1\columnwidth, trim={0cm 0cm 0cm 0cm}, clip]{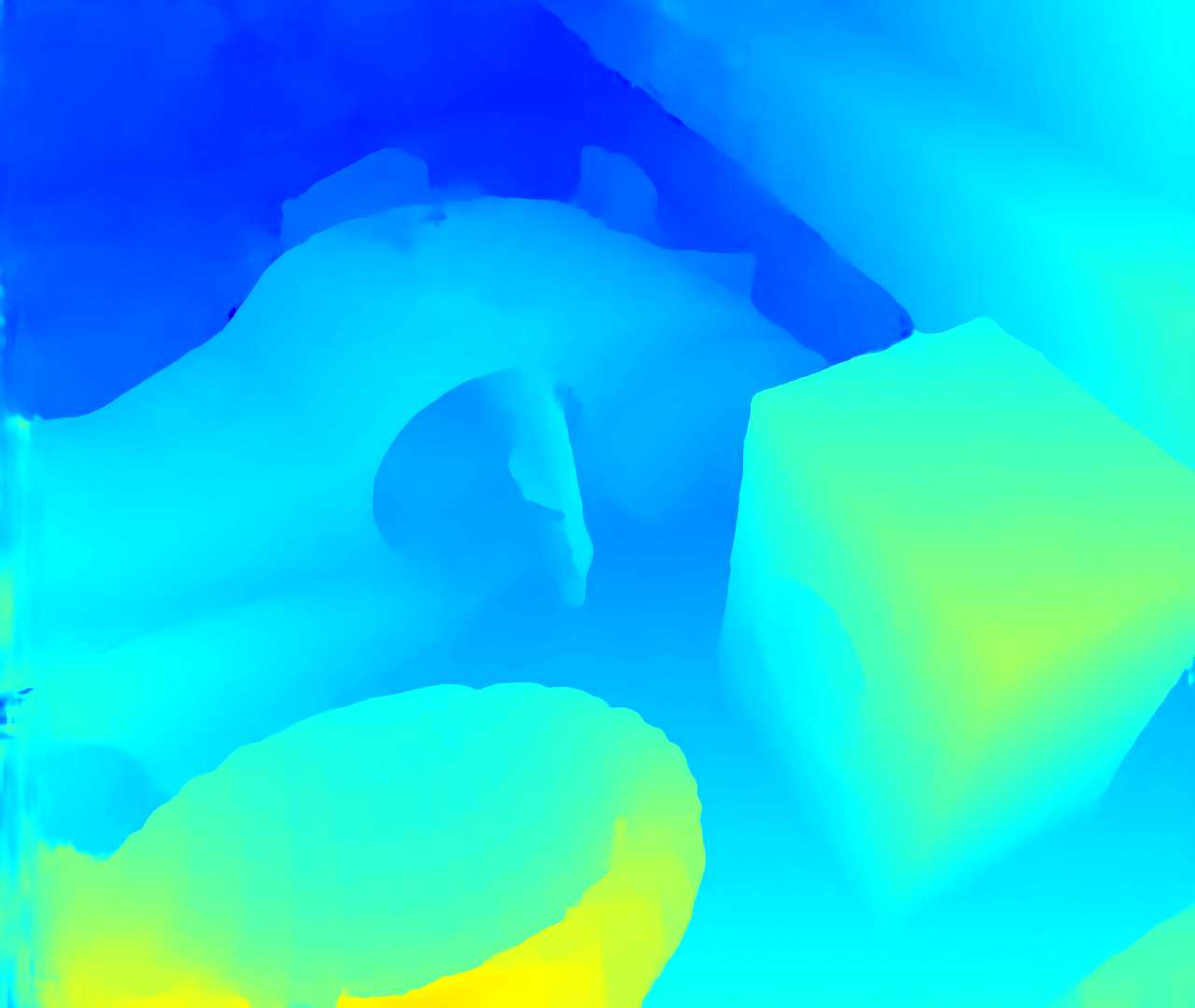}
\end{subfigure}


\begin{subfigure}{0.48\columnwidth}
  \centering
  \includegraphics[width=1\columnwidth, trim={0cm 0cm 0cm 0cm}, clip]{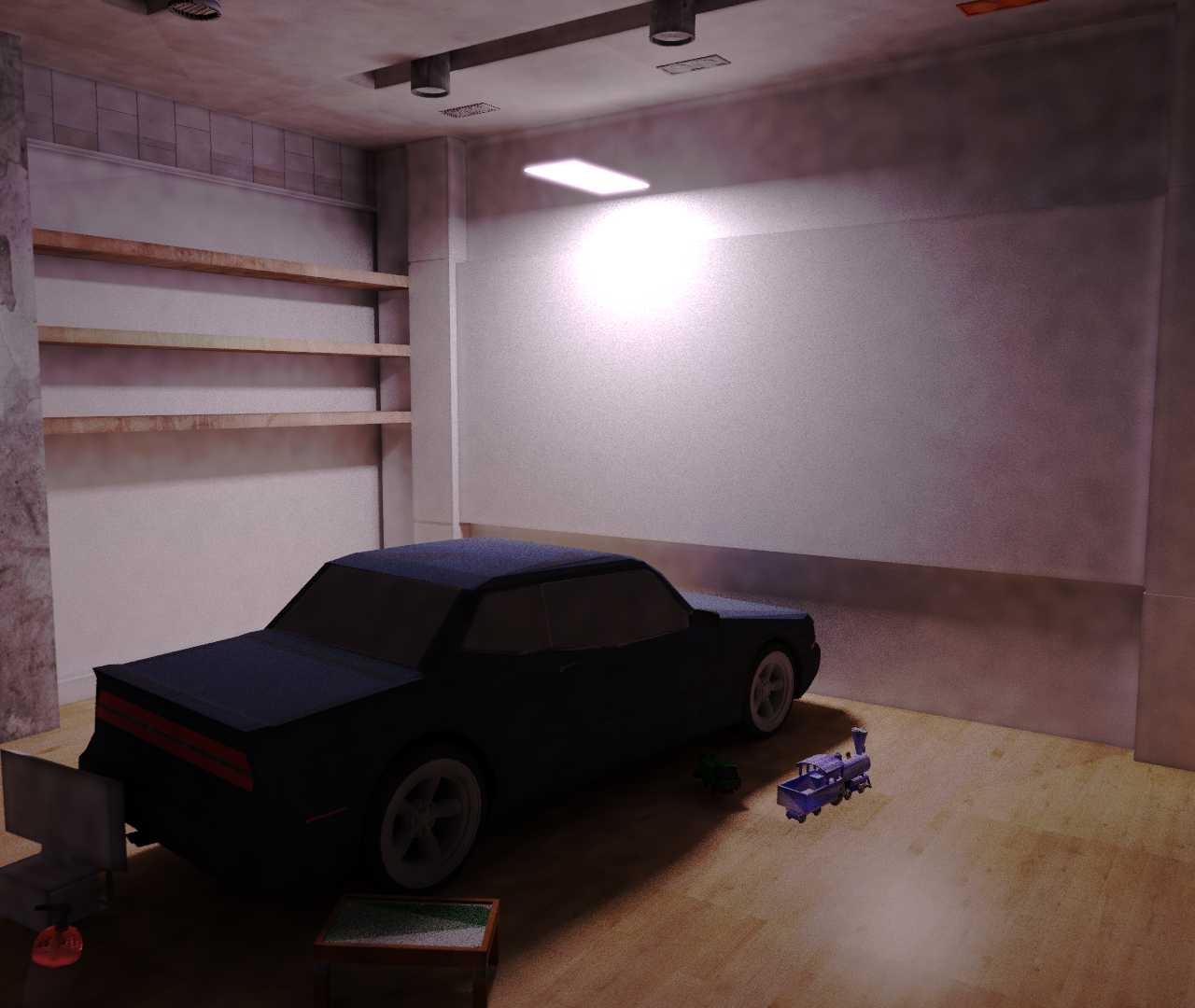}
  \caption*{Center view}
\end{subfigure}
\begin{subfigure}{0.48\columnwidth}
  \centering
  \includegraphics[width=1\columnwidth, trim={0cm 0cm 0cm 0cm}, clip]{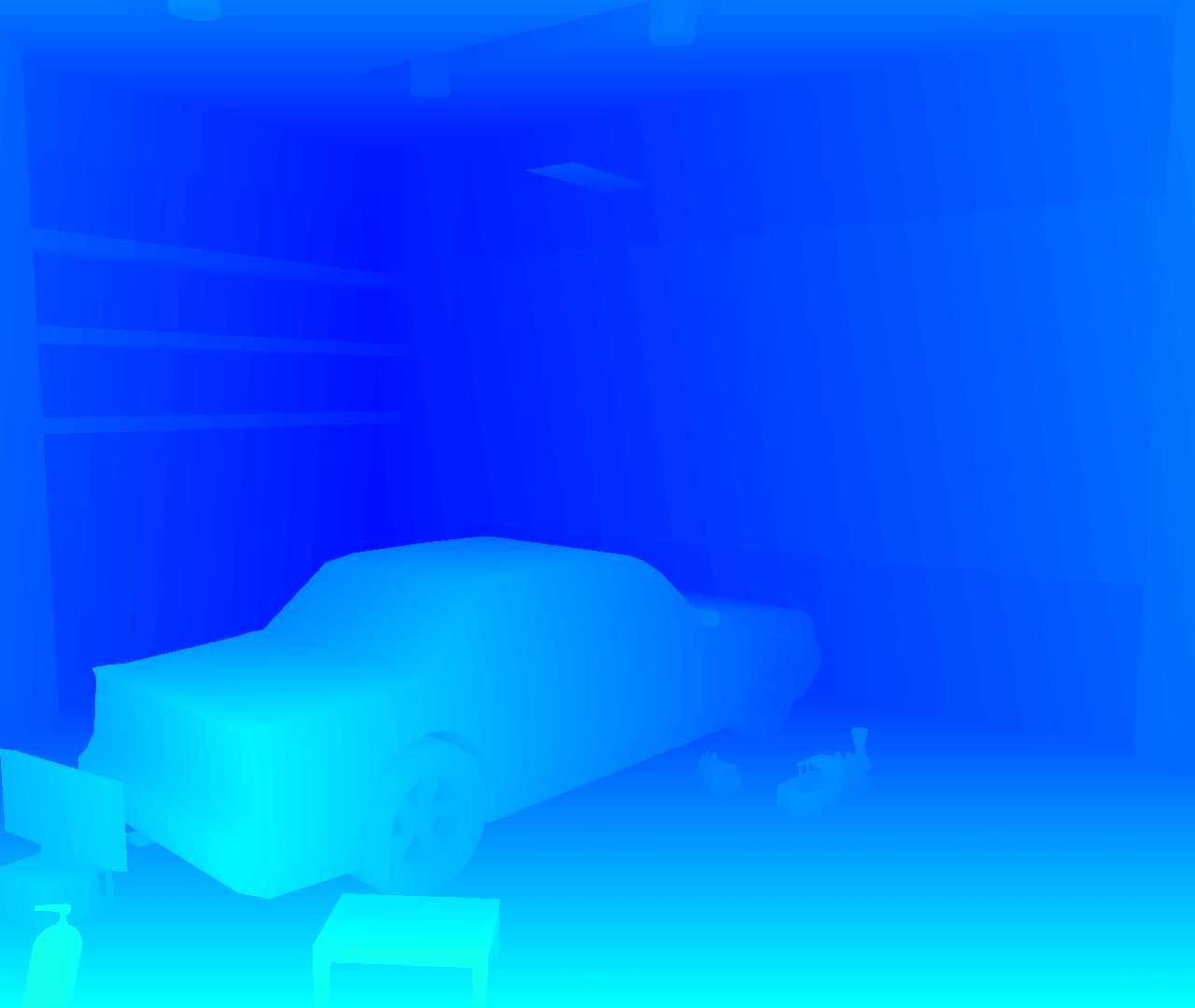}
  \caption*{Ground truth}
\end{subfigure}
\begin{subfigure}{0.48\columnwidth}
  \centering
  \includegraphics[width=1\columnwidth, trim={0cm 0cm 0cm 0cm}, clip]{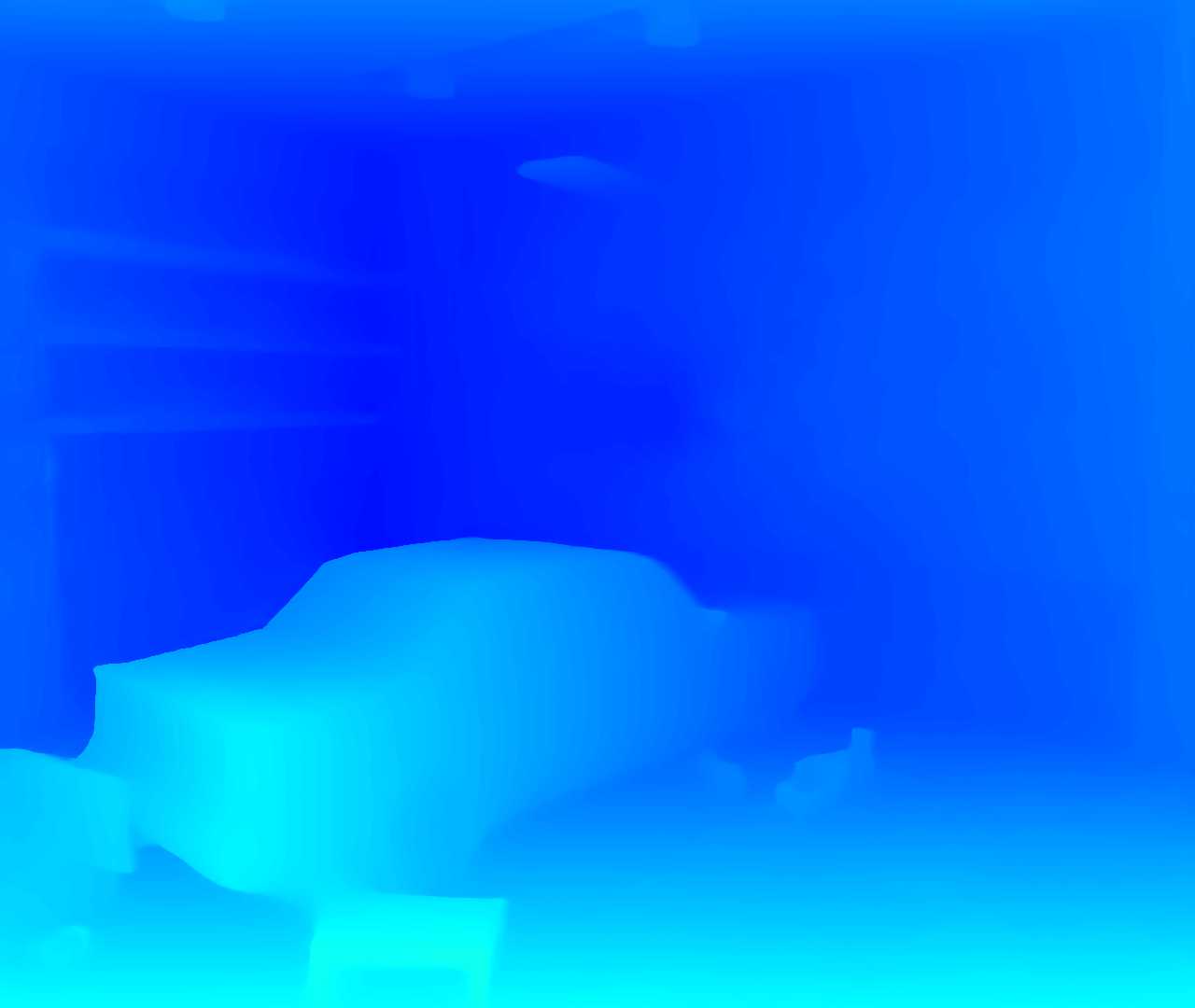}
  \caption*{Our model}
\end{subfigure}
\begin{subfigure}{0.48\columnwidth}
  \centering
  \includegraphics[width=1\columnwidth, trim={0cm 0cm 0cm 0cm}, clip]{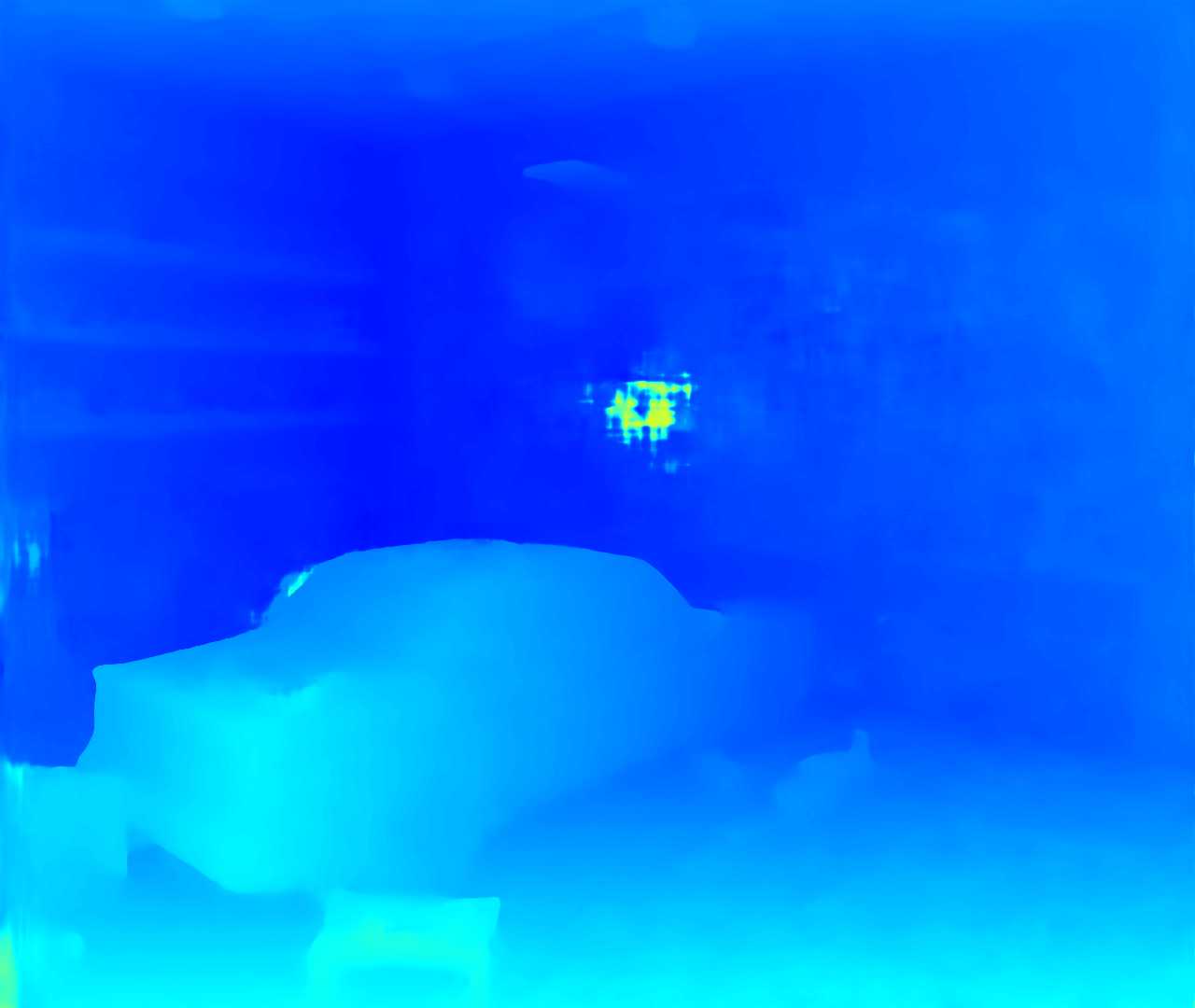}
  \caption*{PSMNet}
\end{subfigure}

\caption{Outputs on synthetic test images. The center-view color images, the ground truth disparity maps, the output disparity maps of our self-supervised model, and the output of supervised method PSMNet are shown.}
\label{fig:app-res}
\end{figure*}

\begin{figure*}[]
\centering
\begin{subfigure}{0.65\columnwidth}
  \centering
  \includegraphics[width=1\columnwidth, trim={0cm 0cm 0cm 0.8cm}, clip]{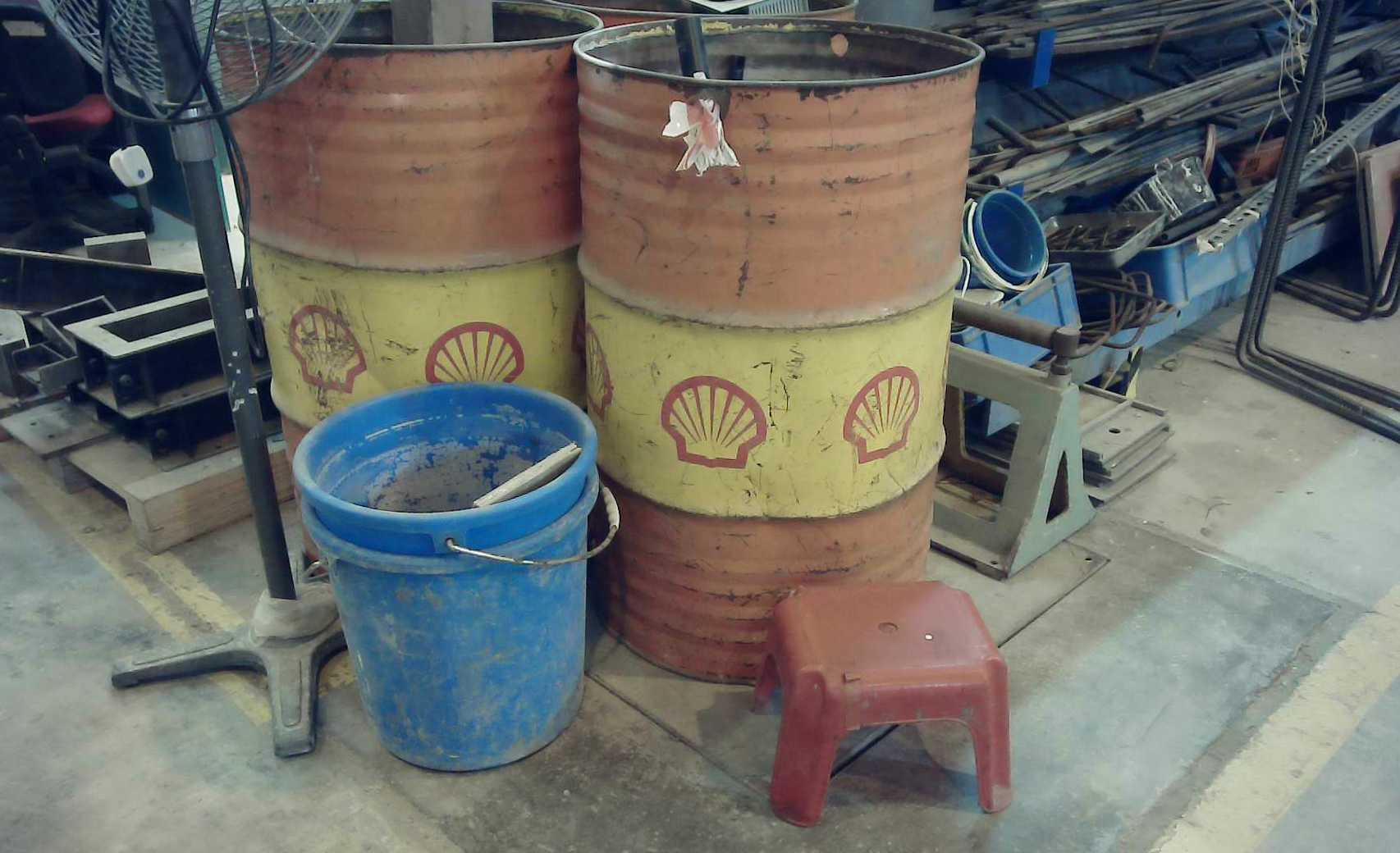}
\end{subfigure}
\begin{subfigure}{0.65\columnwidth}
  \centering
  \includegraphics[width=1\columnwidth, trim={0cm 0cm 0cm 0cm}, clip]{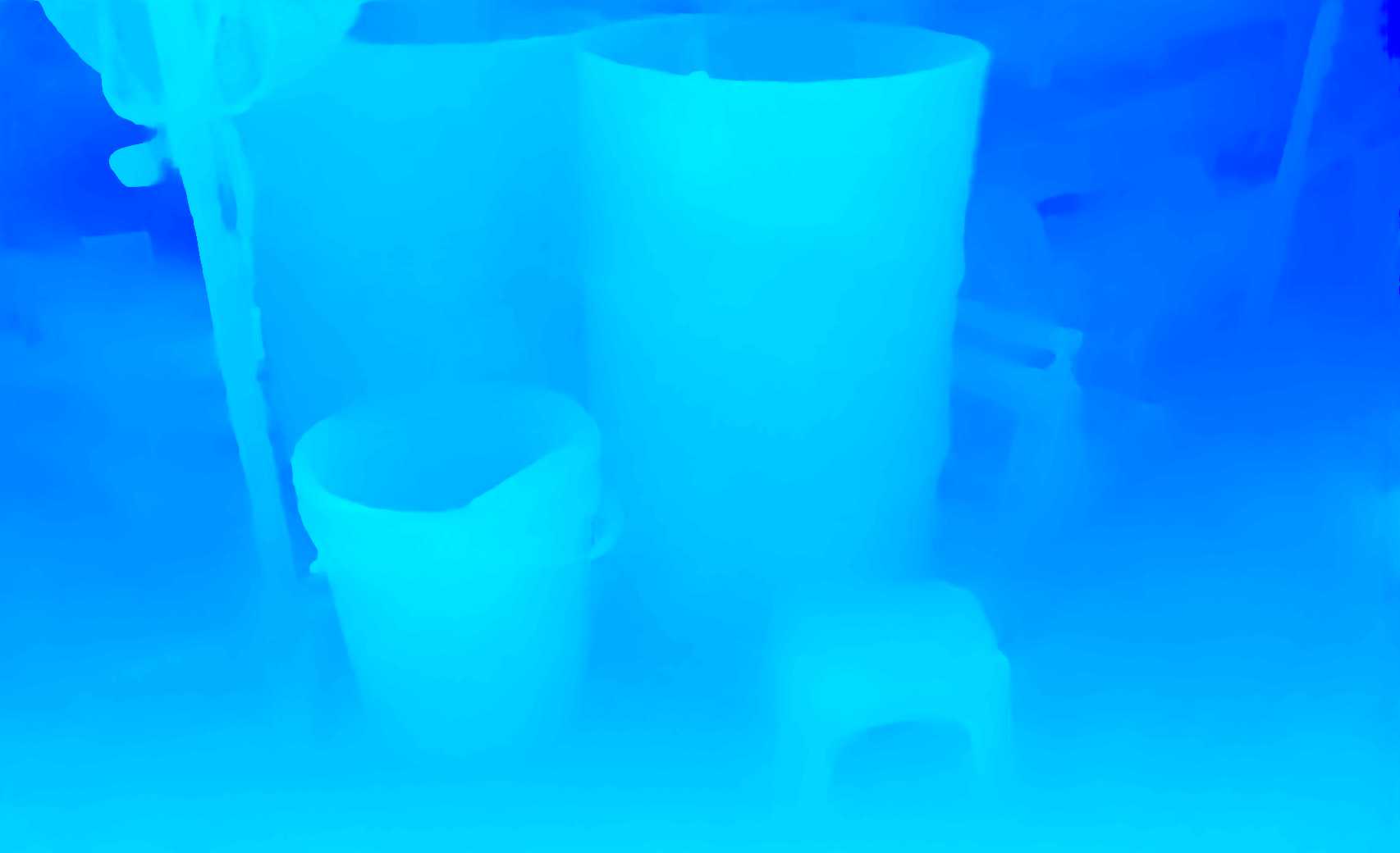}
\end{subfigure}
\begin{subfigure}{0.65\columnwidth}
  \centering
  \includegraphics[width=1\columnwidth, trim={0cm 0cm 0cm 0cm}, clip]{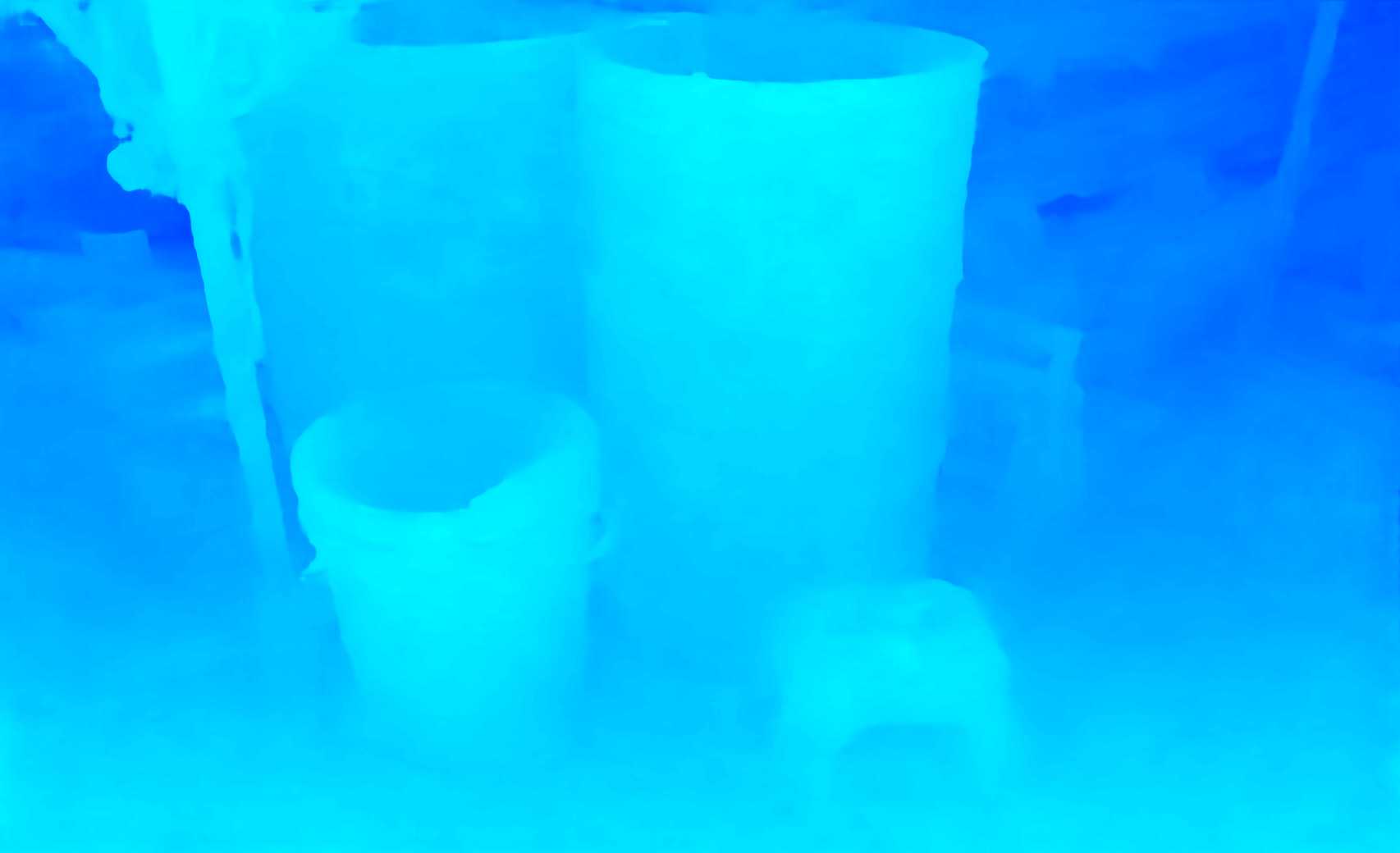}
\end{subfigure}

\begin{subfigure}{0.65\columnwidth}
  \centering
  \includegraphics[width=1\columnwidth, trim={0cm 0cm 0cm 0.8cm}, clip]{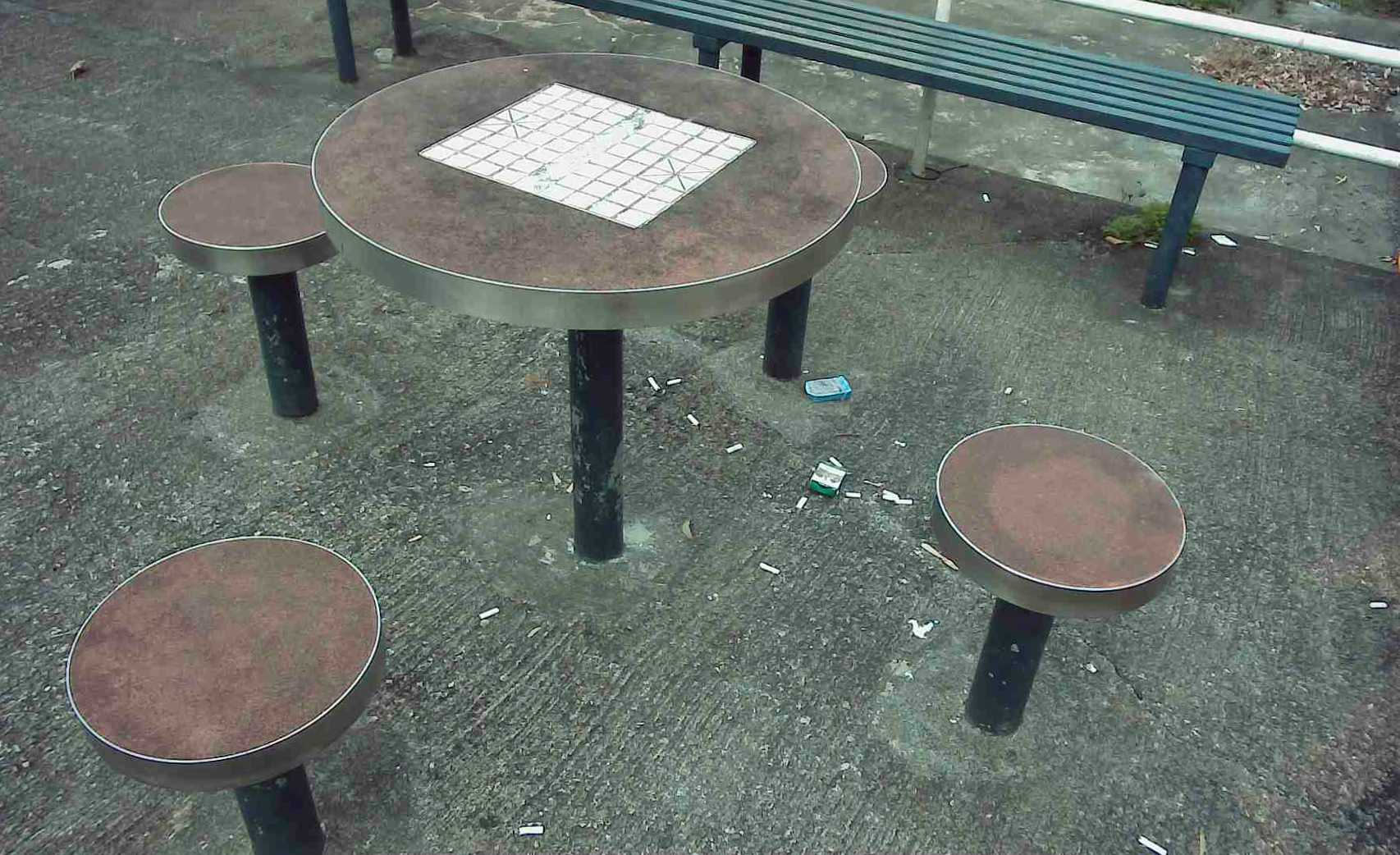}
\end{subfigure}
\begin{subfigure}{0.65\columnwidth}
  \centering
  \includegraphics[width=1\columnwidth, trim={0cm 0cm 0cm 0cm}, clip]{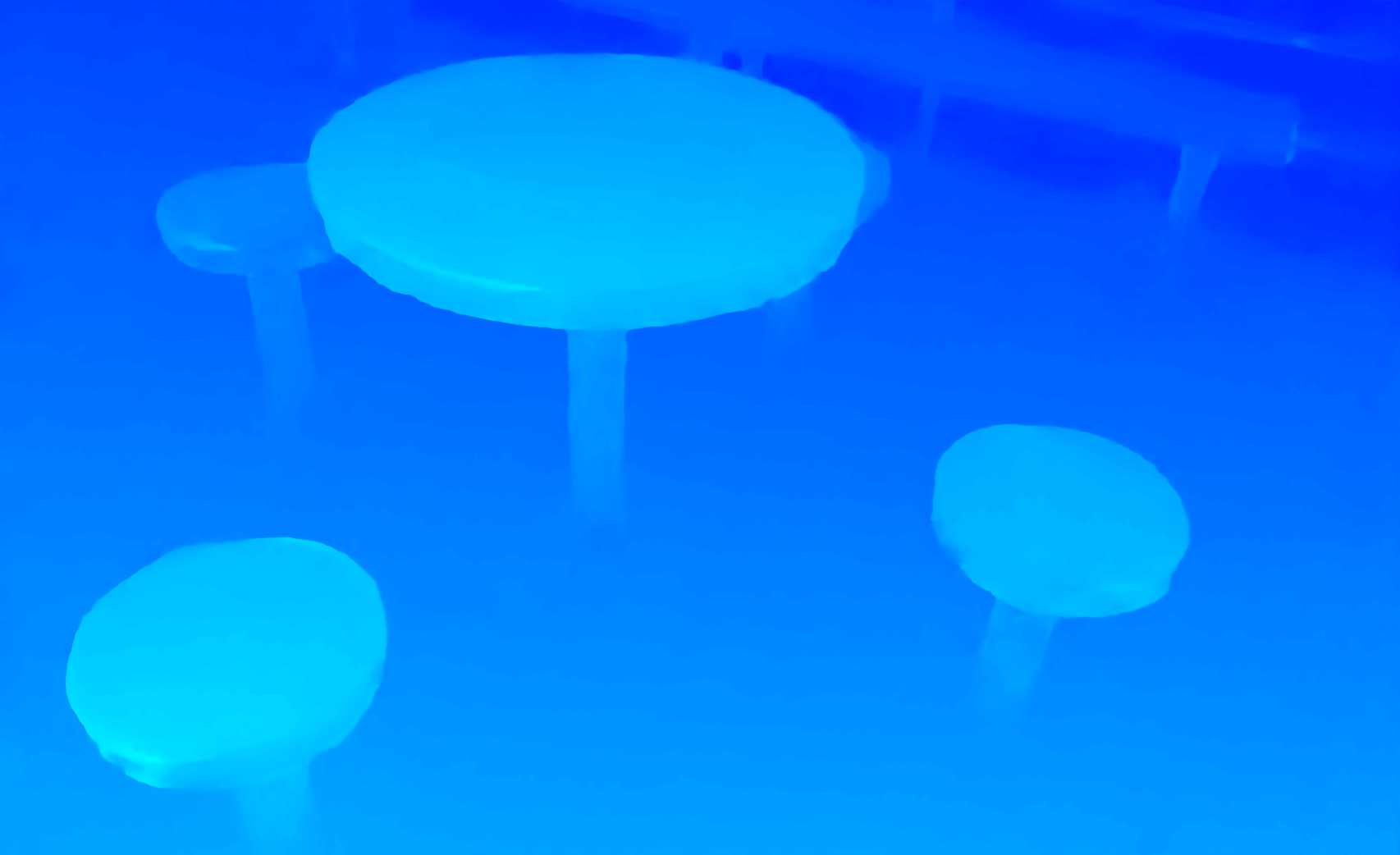}
\end{subfigure}
\begin{subfigure}{0.65\columnwidth}
  \centering
  \includegraphics[width=1\columnwidth, trim={0cm 0cm 0cm 0cm}, clip]{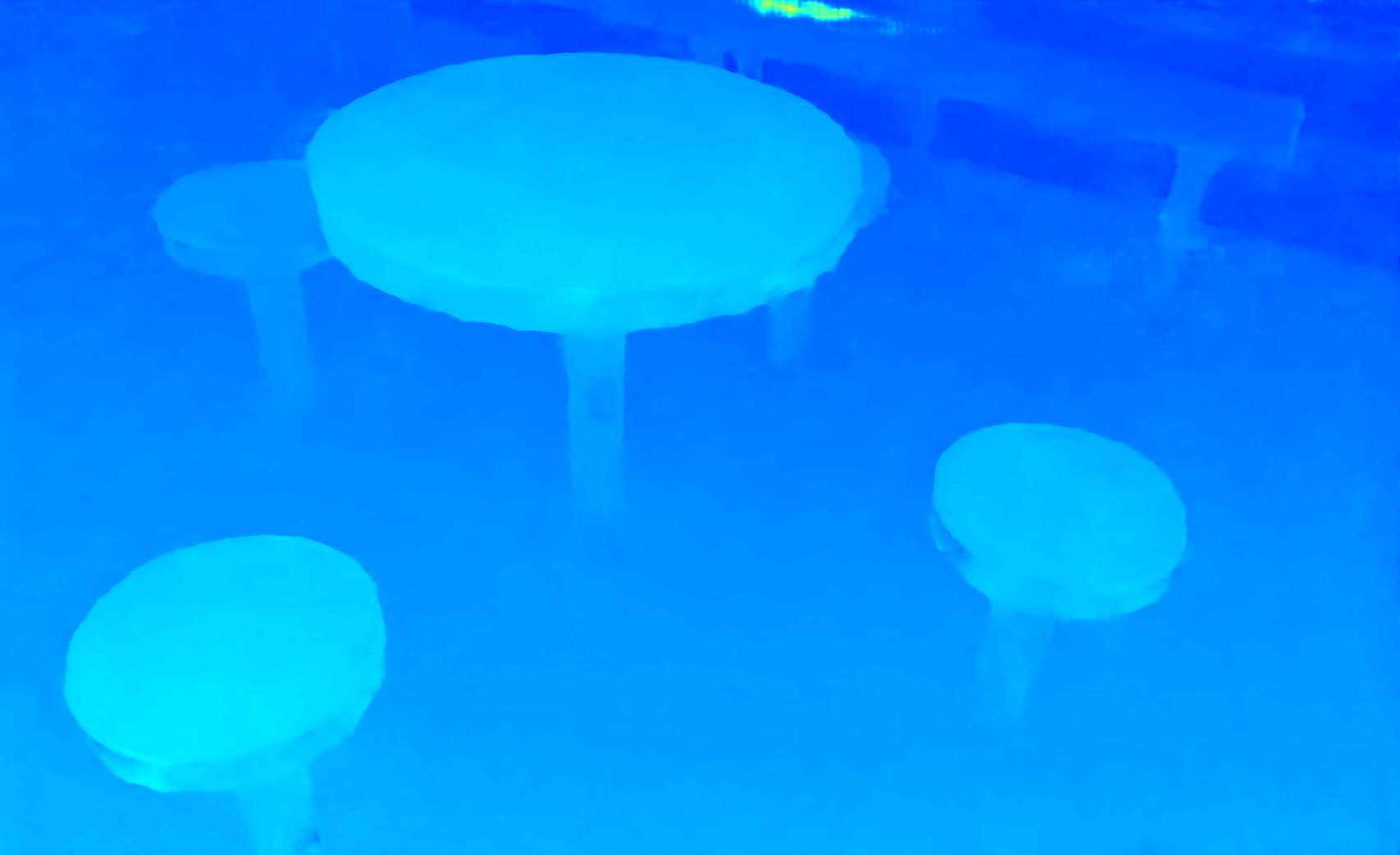}
\end{subfigure}

\begin{subfigure}{0.65\columnwidth}
  \centering
  \includegraphics[width=1\columnwidth, trim={0cm 0cm 0cm 0.8cm}, clip]{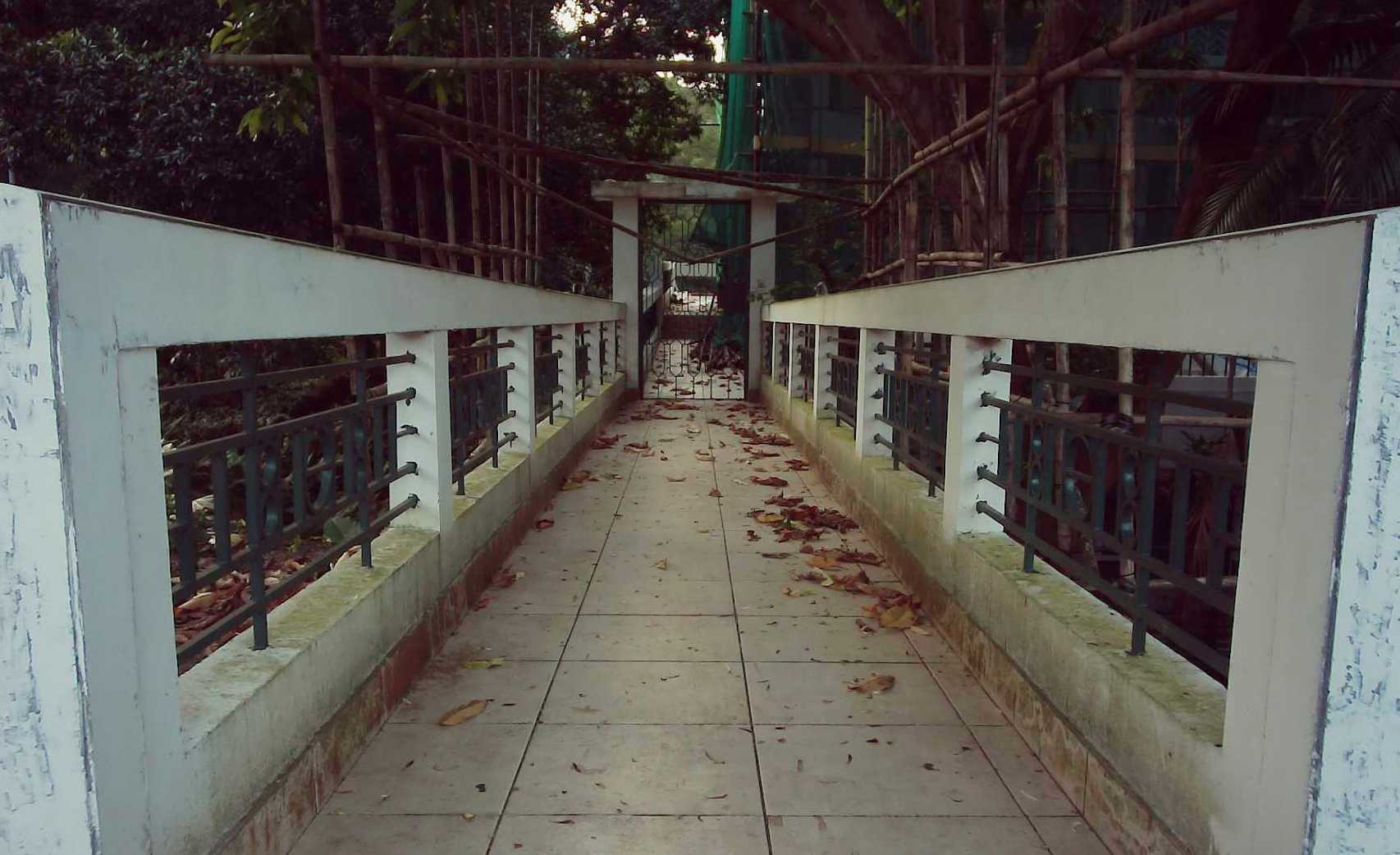}
  \caption*{Center view}
\end{subfigure}
\begin{subfigure}{0.65\columnwidth}
  \centering
  \includegraphics[width=1\columnwidth, trim={0cm 0cm 0cm 0cm}, clip]{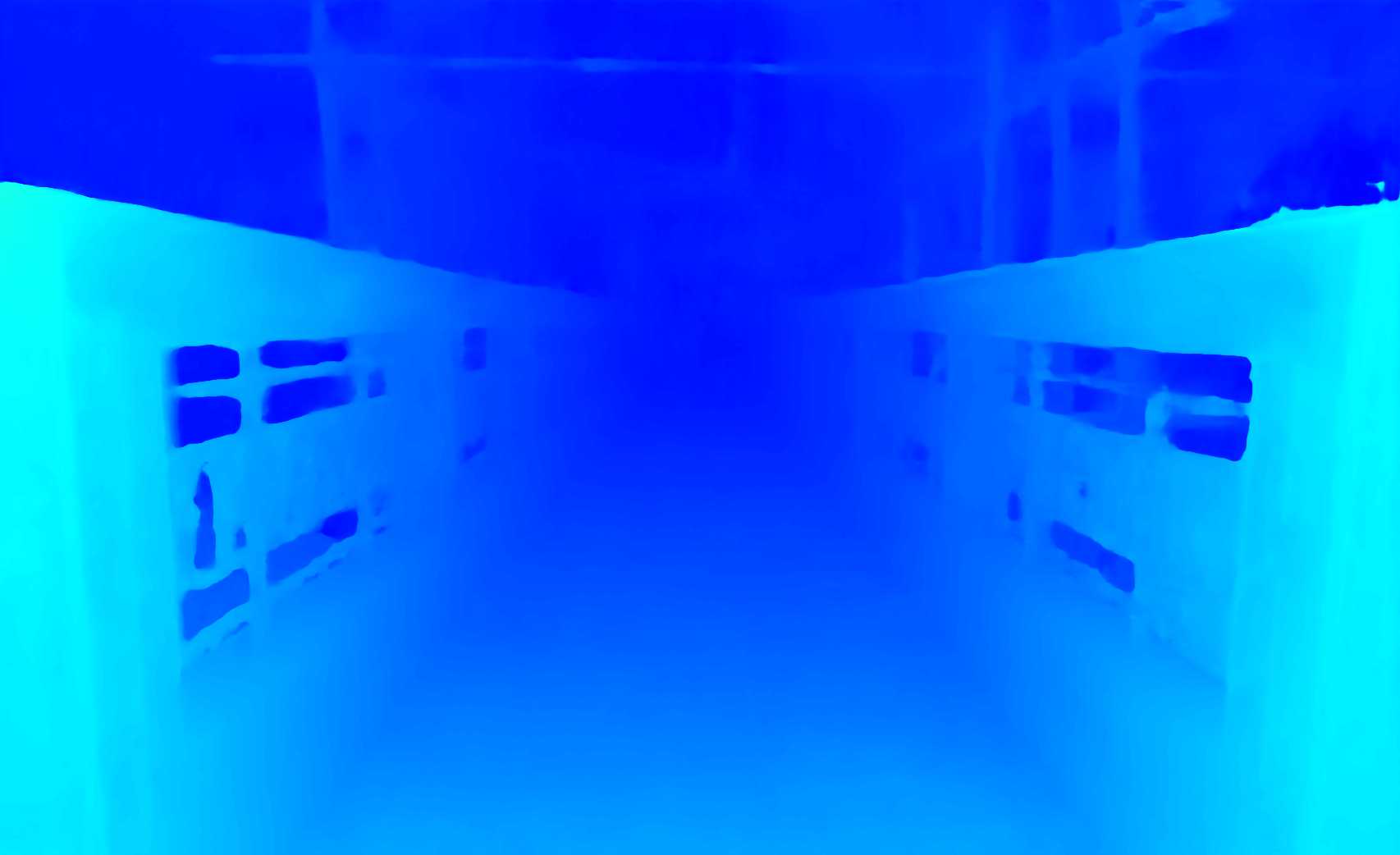}
  \caption*{Our model}
\end{subfigure}
\begin{subfigure}{0.65\columnwidth}
  \centering
  \includegraphics[width=1\columnwidth, trim={0cm 0cm 0cm 0cm}, clip]{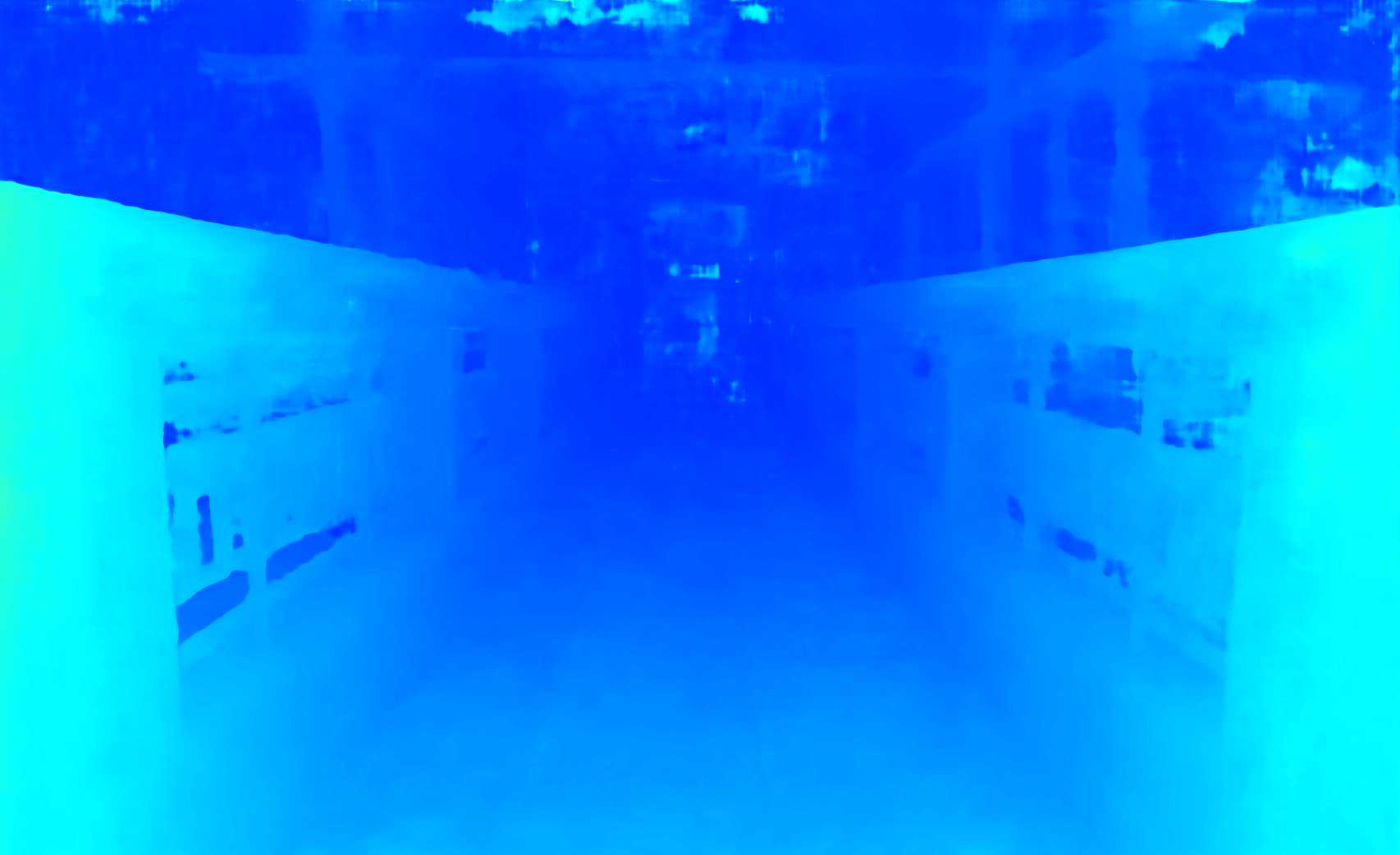}
  \caption*{PSMNet}
\end{subfigure}
\caption{Outputs on real images. The center-view color images, the output disparity maps of our self-supervised model, and the output of supervised method PSMNet are shown.}
\label{fig:app-real}
\end{figure*}

\begin{figure*}[]
\centering
\begin{subfigure}{0.65\columnwidth}
  \centering
  \includegraphics[width=1\columnwidth, trim={0cm 4cm 0cm 0.8cm}, clip]{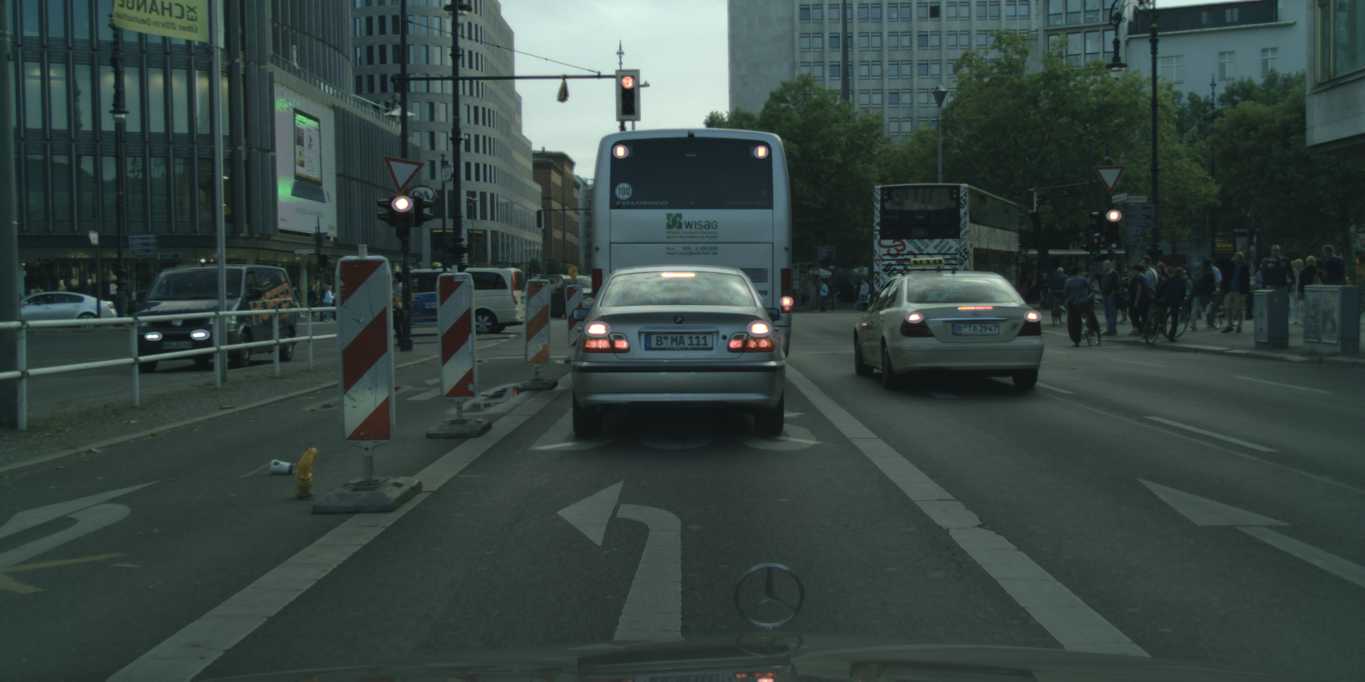}
\end{subfigure}
\begin{subfigure}{0.65\columnwidth}
  \centering
  \includegraphics[width=1\columnwidth, trim={0cm 4cm 0cm 0cm}, clip]{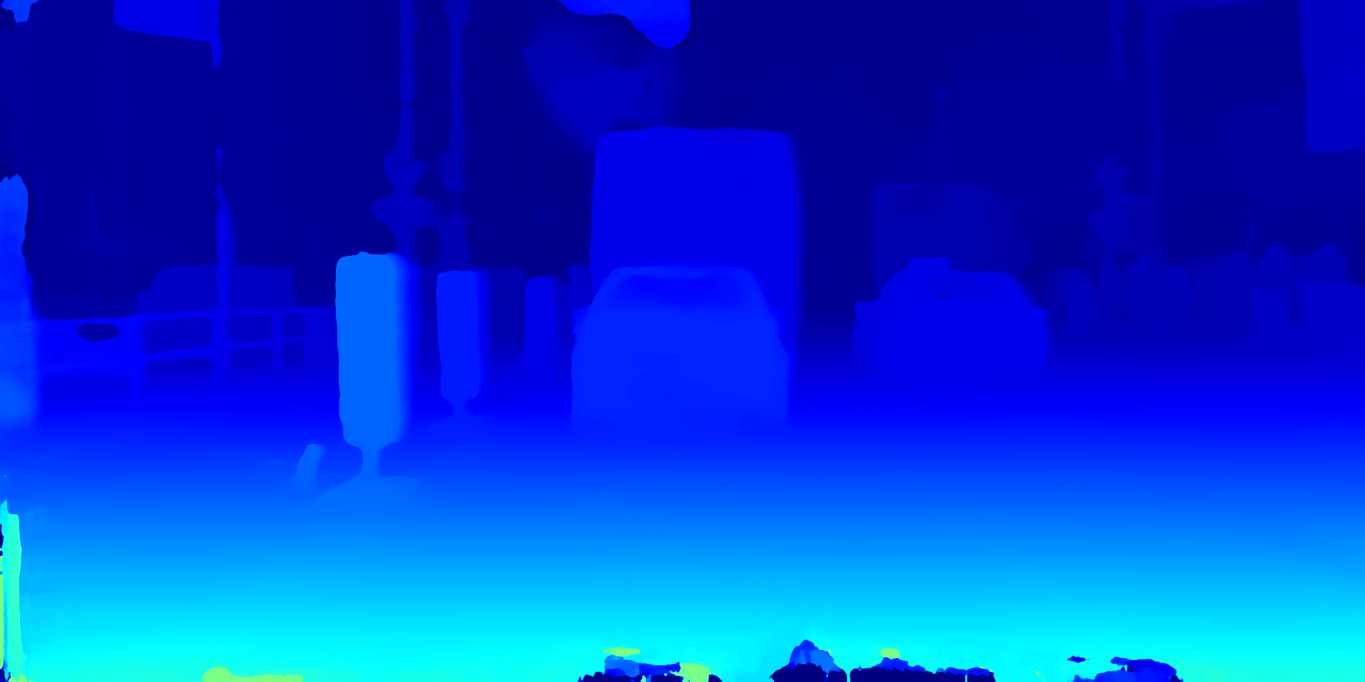}
\end{subfigure}
\begin{subfigure}{0.65\columnwidth}
  \centering
  \includegraphics[width=1\columnwidth, trim={0cm 4cm 0cm 0cm}, clip]{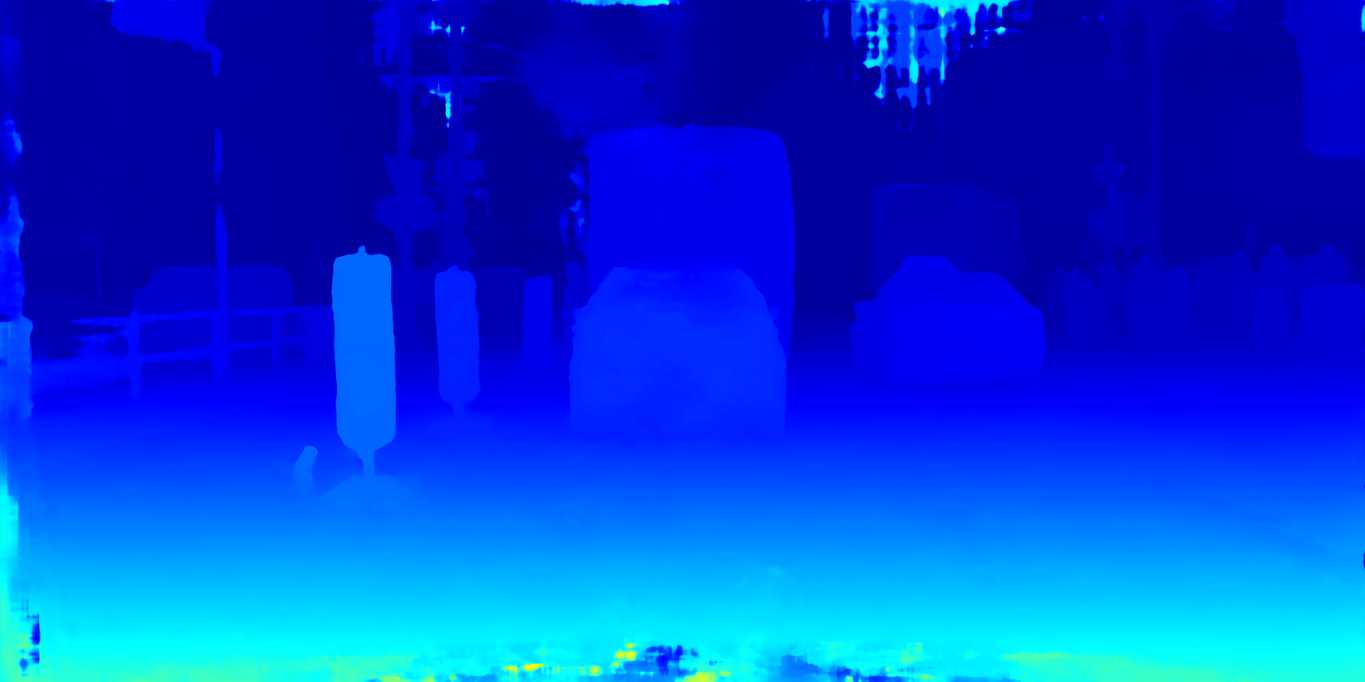}
\end{subfigure}

\begin{subfigure}{0.65\columnwidth}
  \centering
  \includegraphics[width=1\columnwidth, trim={0cm 4cm 0cm 0.8cm}, clip]{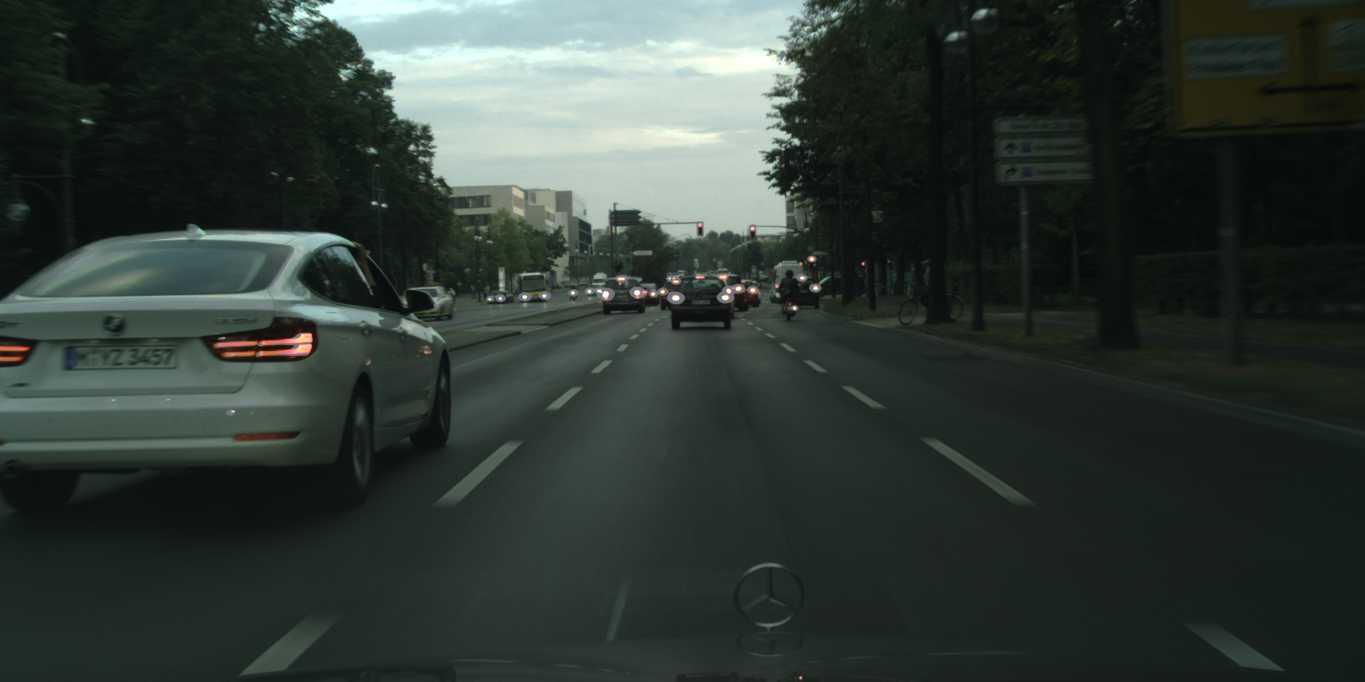}
\end{subfigure}
\begin{subfigure}{0.65\columnwidth}
  \centering
  \includegraphics[width=1\columnwidth, trim={0cm 4cm 0cm 0cm}, clip]{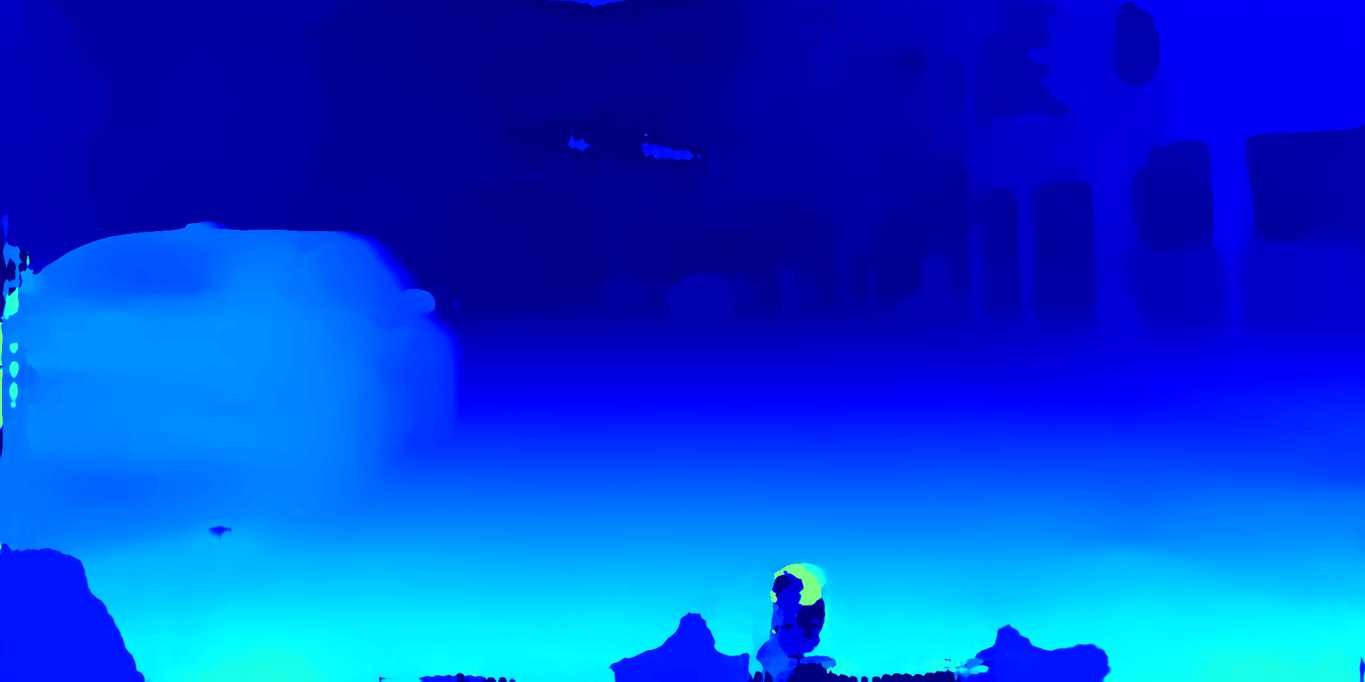}
\end{subfigure}
\begin{subfigure}{0.65\columnwidth}
  \centering
  \includegraphics[width=1\columnwidth, trim={0cm 4cm 0cm 0cm}, clip]{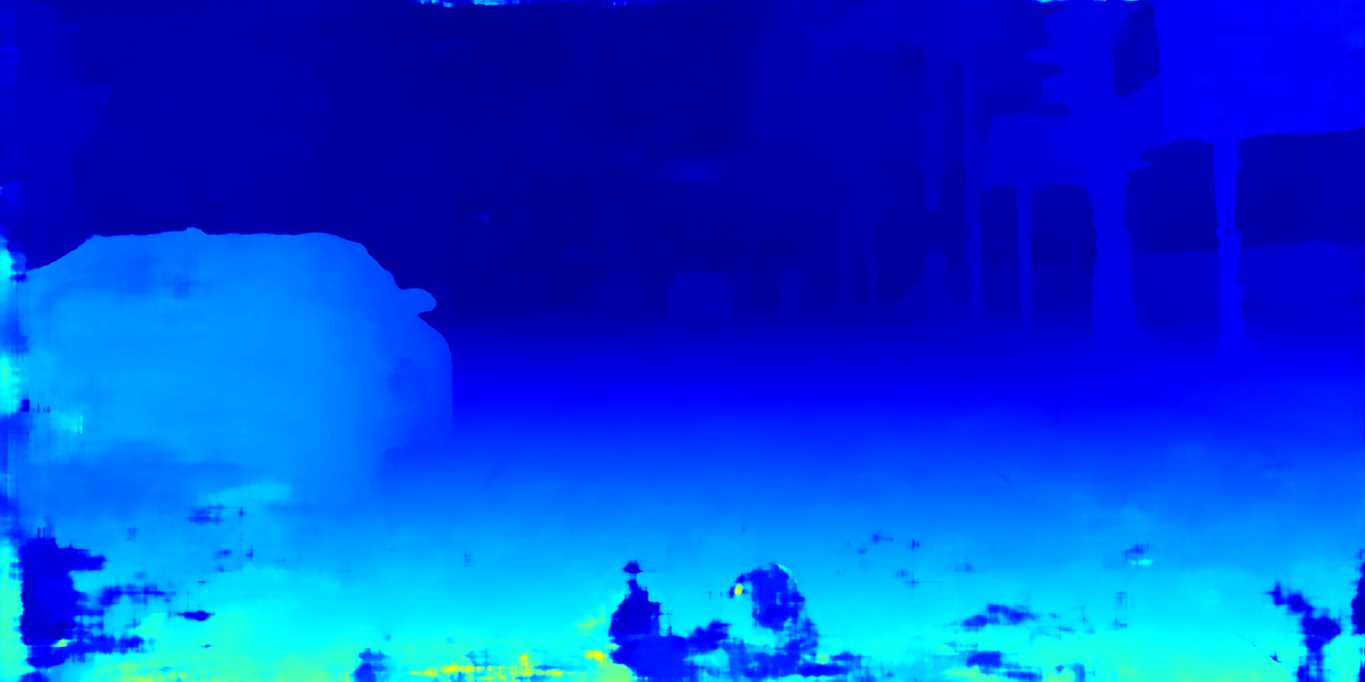}
\end{subfigure}

\begin{subfigure}{0.65\columnwidth}
  \centering
  \includegraphics[width=1\columnwidth, trim={0cm 4cm 0cm 0.8cm}, clip]{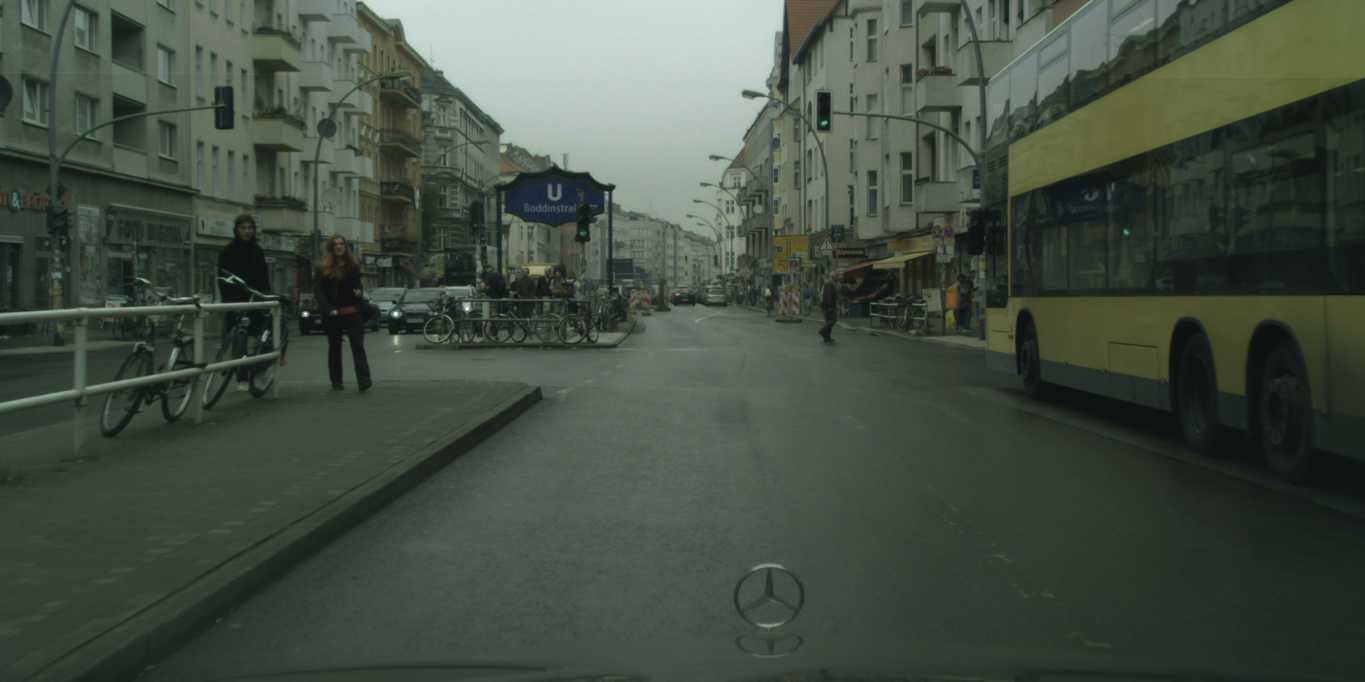}
  \caption*{Left image}
\end{subfigure}
\begin{subfigure}{0.65\columnwidth}
  \centering
  \includegraphics[width=1\columnwidth, trim={0cm 4cm 0cm 0cm}, clip]{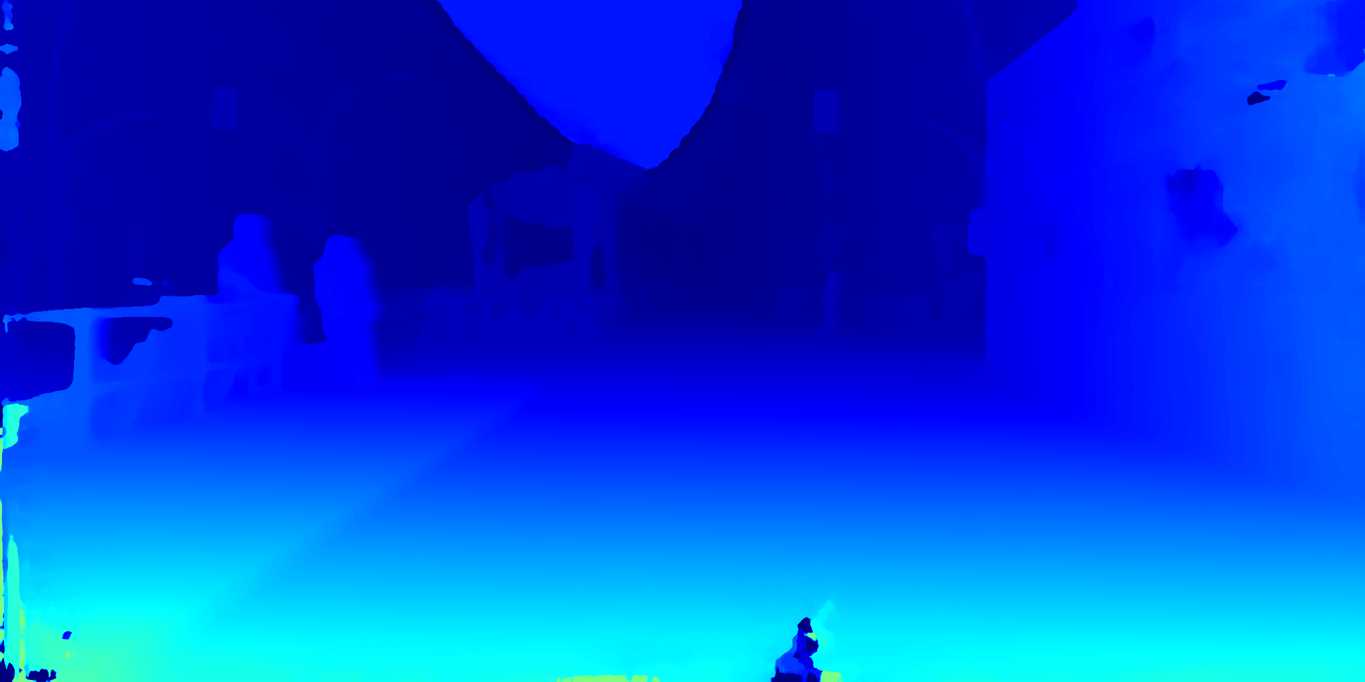}
  \caption*{Our model}
\end{subfigure}
\begin{subfigure}{0.65\columnwidth}
  \centering
  \includegraphics[width=1\columnwidth, trim={0cm 4cm 0cm 0cm}, clip]{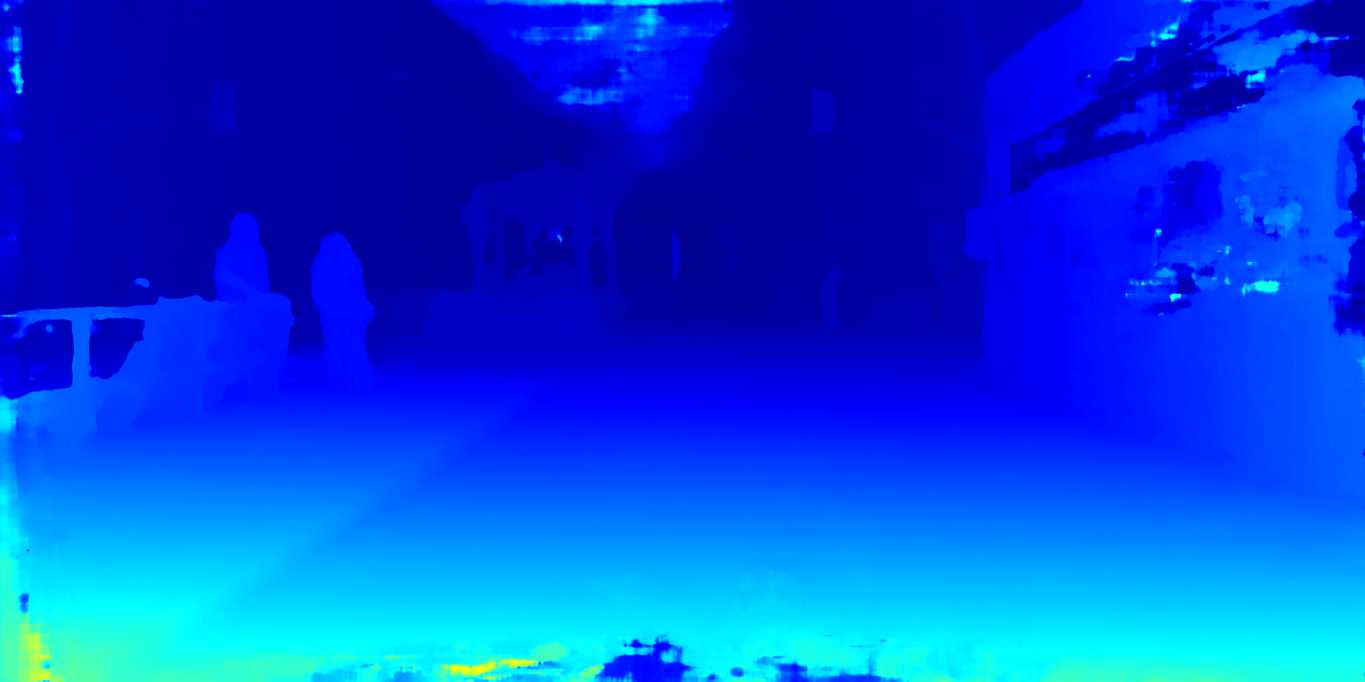}
  \caption*{PSMNet}
\end{subfigure}
\caption{Outputs on Cityscapes dataset. The center-view color images, the output disparity maps of our self-supervised model, and the output of supervised method PSMNet are shown.}
\label{fig:app-city}
\end{figure*}

We show more qualitative results of our self-supervised model and PSMNet on the synthetic test set in Fig.~\ref{fig:app-res}, after trained on our synthetic data. Also, we present more results on real images: Fig.~\ref{fig:app-real} displays the outputs on the real images obtained by our multiscopic system, in which images our self-supervised model always performs better than the supervised method PSMNet. Fig.~\ref{fig:app-city} displays the outputs on the Cityscapes dataset \cite{Cordts2016Cityscapes}. Since this dataset is similar to the original training set of PSMNet, here we display the result of the original PSMNet model, which performs better than the one trained on our synthetic data. In this case, the performance on Cityscapes of our model and PSMNet are comparable.

\section{Online Training}

\begin{figure*}[]
\centering
\begin{subfigure}{0.65\columnwidth}
  \centering
  \includegraphics[width=1\columnwidth, trim={0cm 0cm 0cm 0.8cm}, clip]{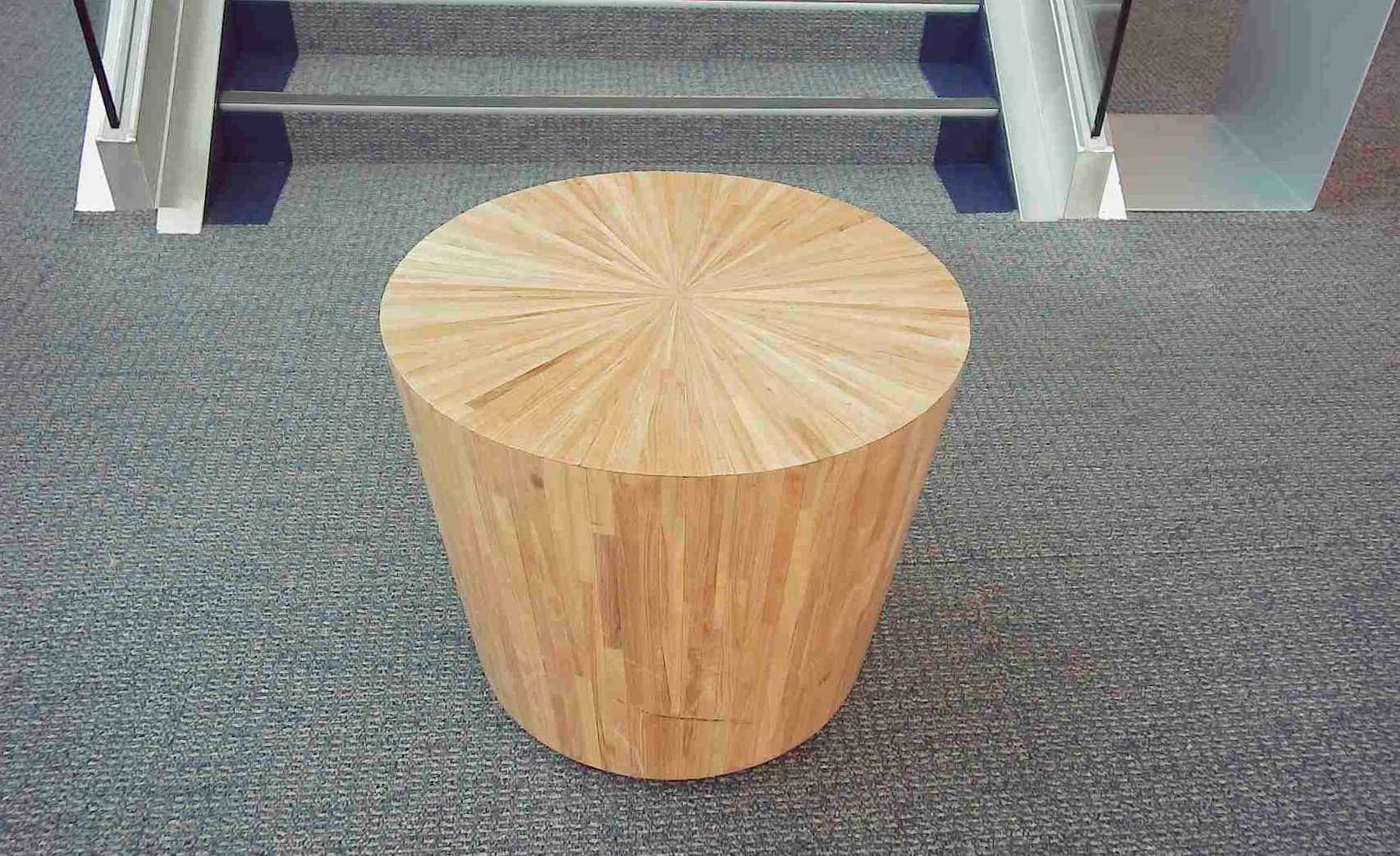}
\end{subfigure}
\begin{subfigure}{0.65\columnwidth}
  \centering
  \includegraphics[width=1\columnwidth, trim={0cm 0cm 0cm 0cm}, clip]{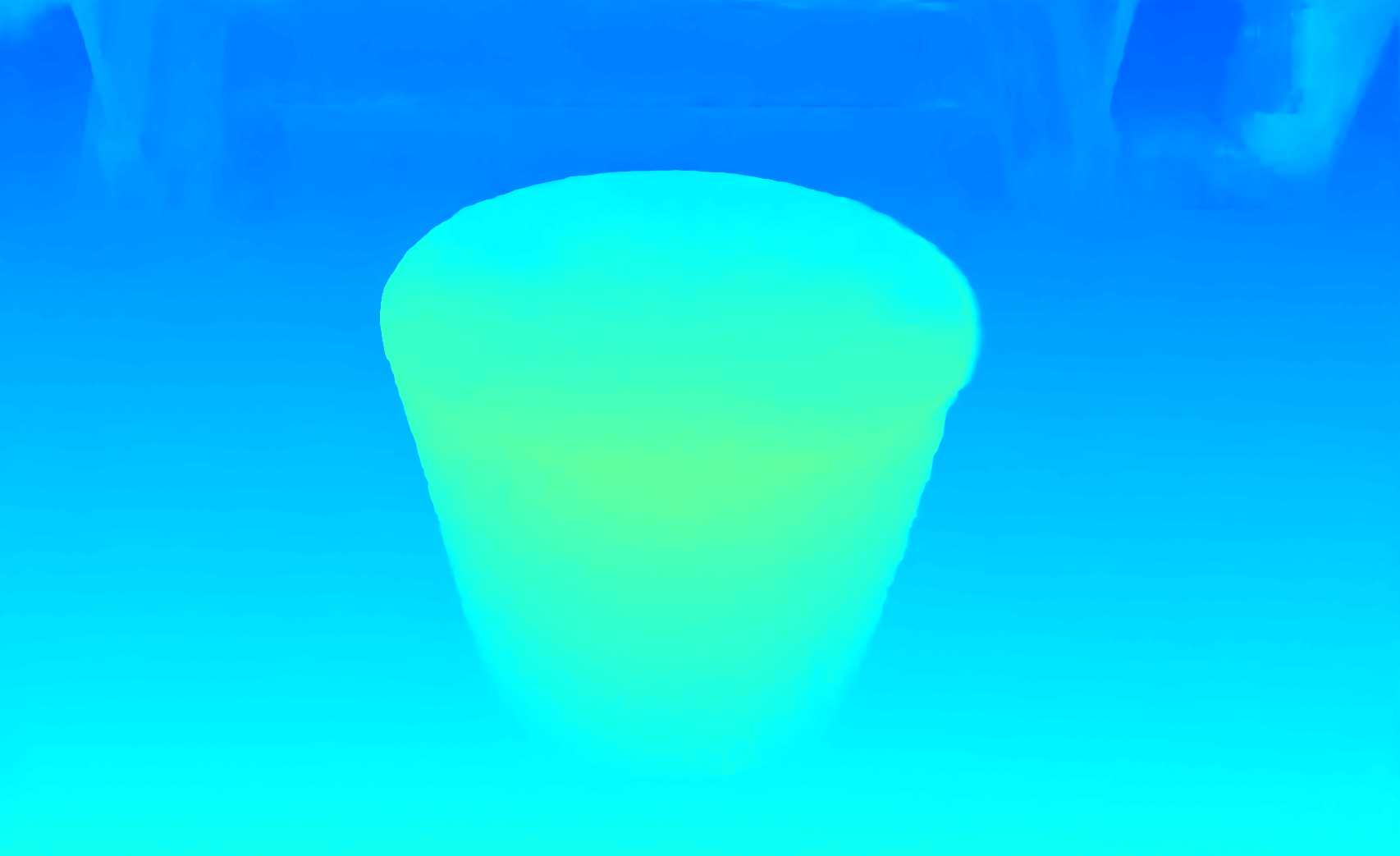}
\end{subfigure}
\begin{subfigure}{0.65\columnwidth}
  \centering
  \includegraphics[width=1\columnwidth, trim={0cm 0cm 0cm 0cm}, clip]{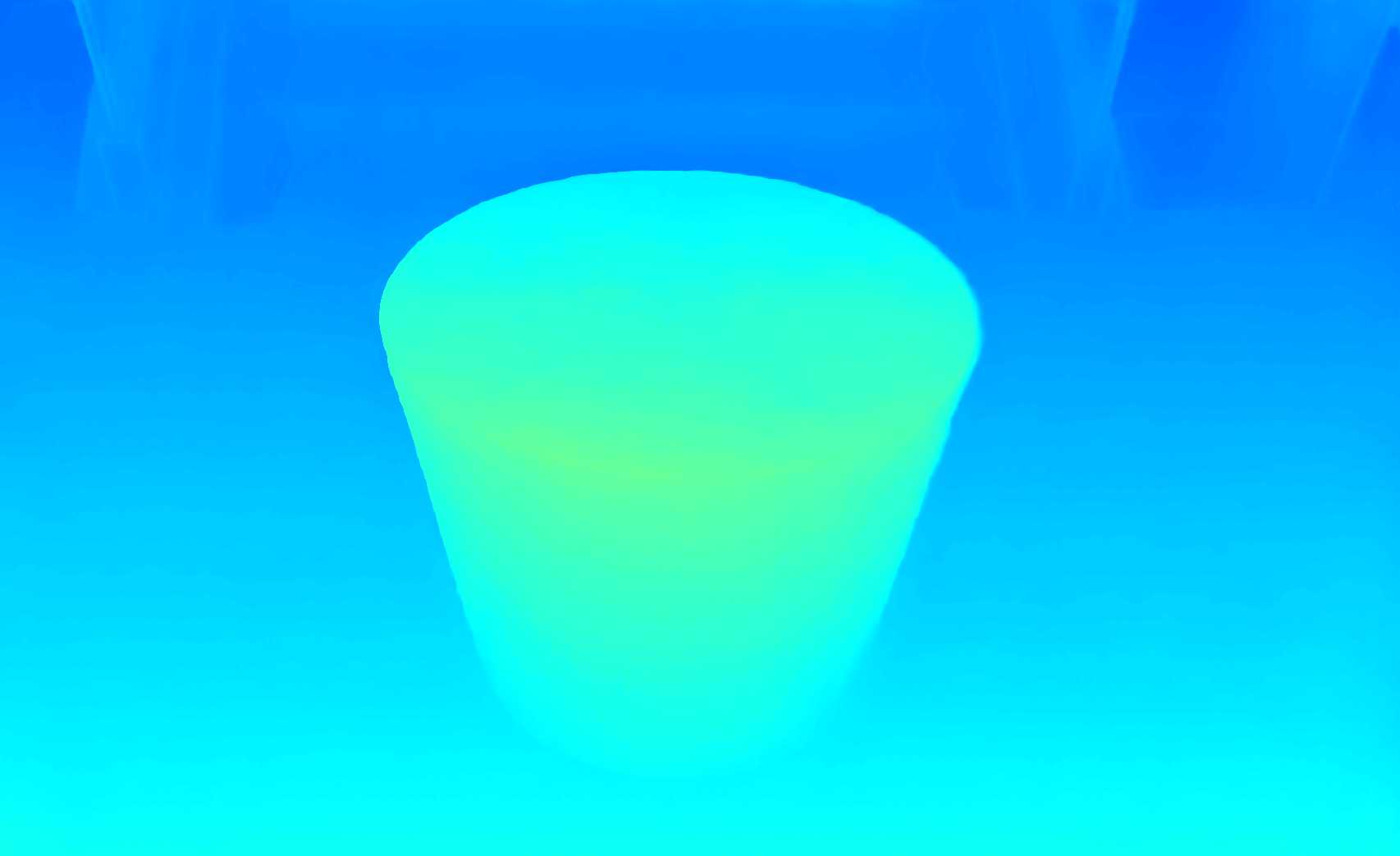}
\end{subfigure}

\begin{subfigure}{0.65\columnwidth}
  \centering
  \includegraphics[width=1\columnwidth, trim={0cm 0cm 0cm 0.8cm}, clip]{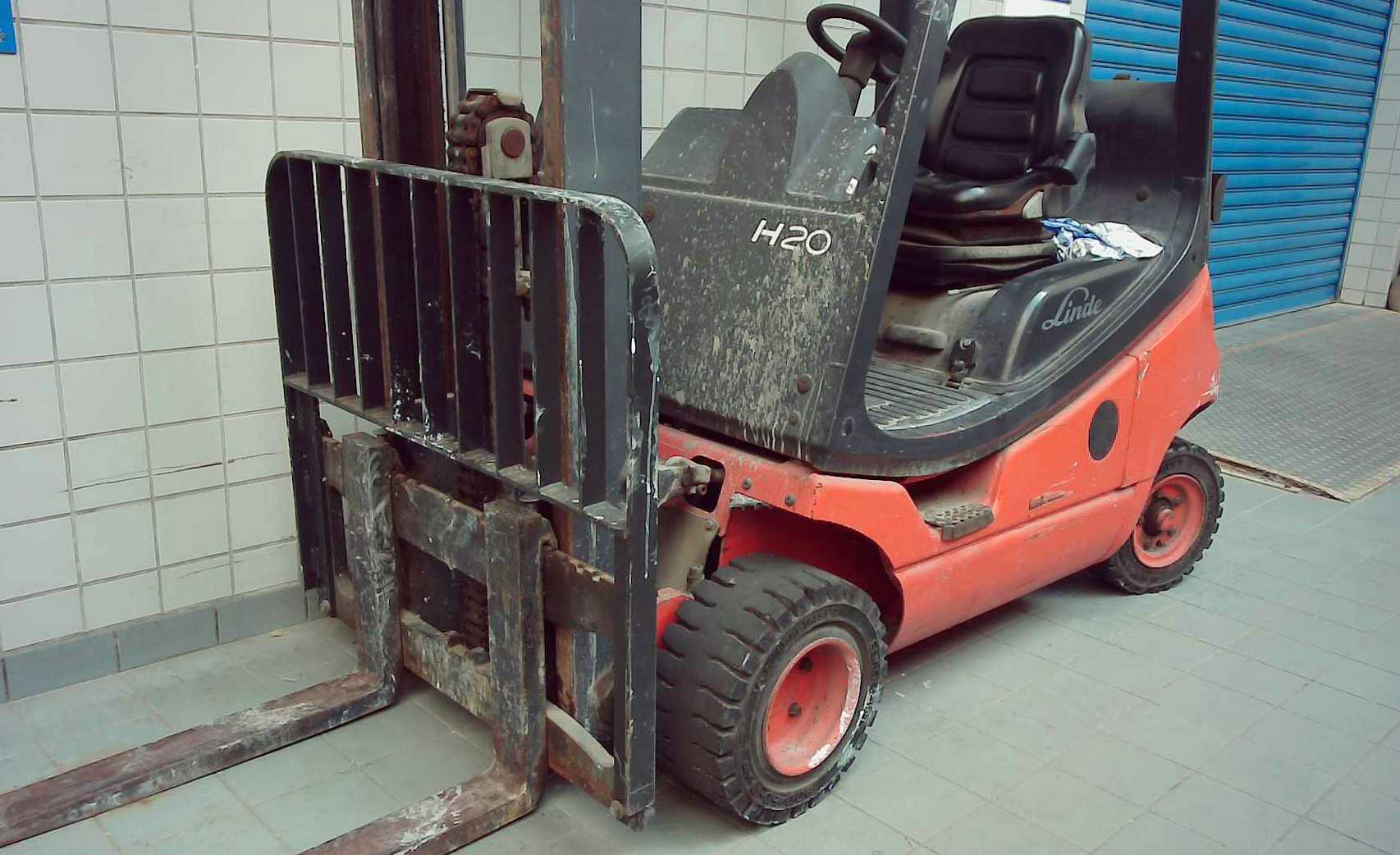}
\end{subfigure}
\begin{subfigure}{0.65\columnwidth}
  \centering
  \includegraphics[width=1\columnwidth, trim={0cm 0cm 0cm 0cm}, clip]{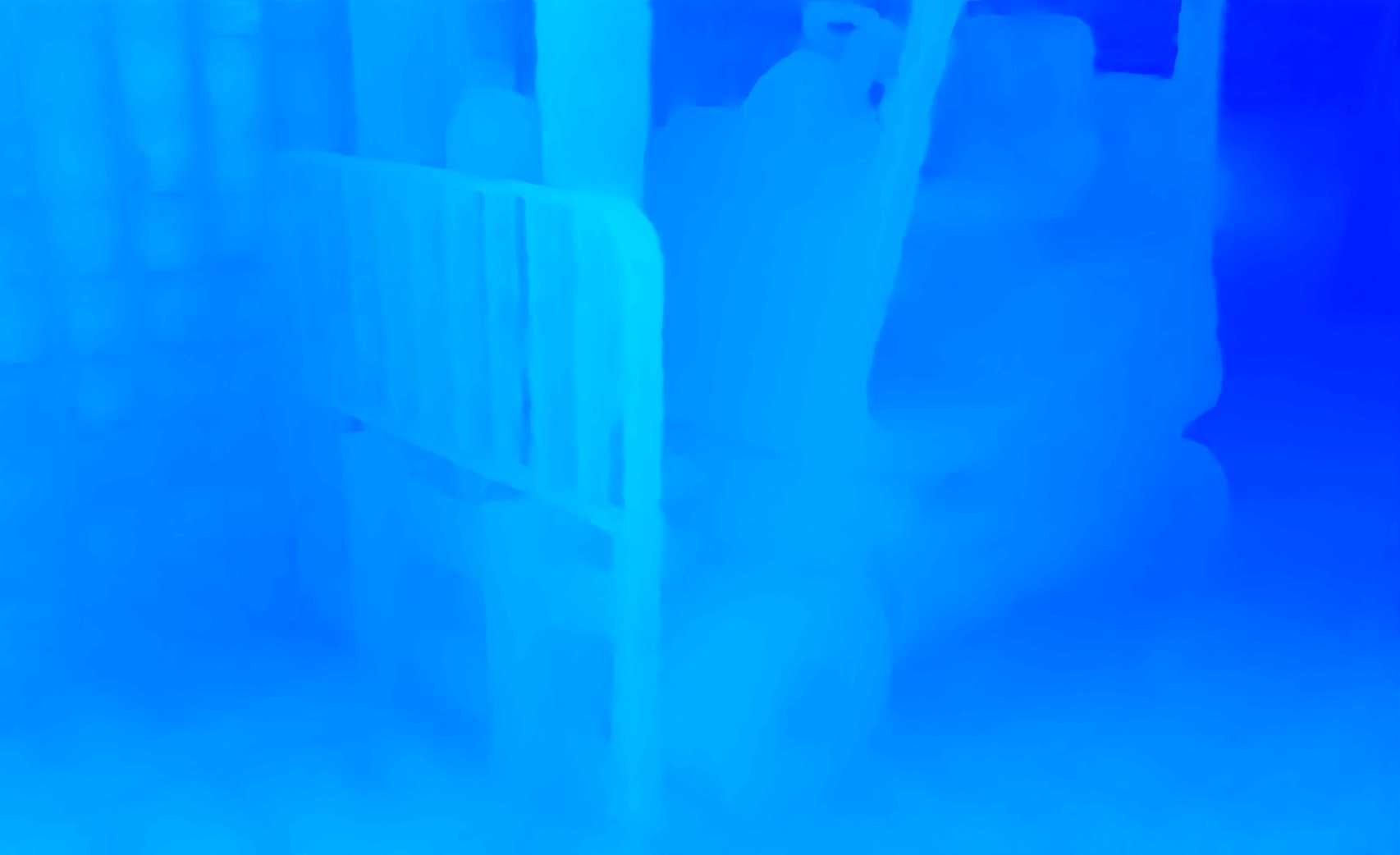}
\end{subfigure}
\begin{subfigure}{0.65\columnwidth}
  \centering
  \includegraphics[width=1\columnwidth, trim={0cm 0cm 0cm 0cm}, clip]{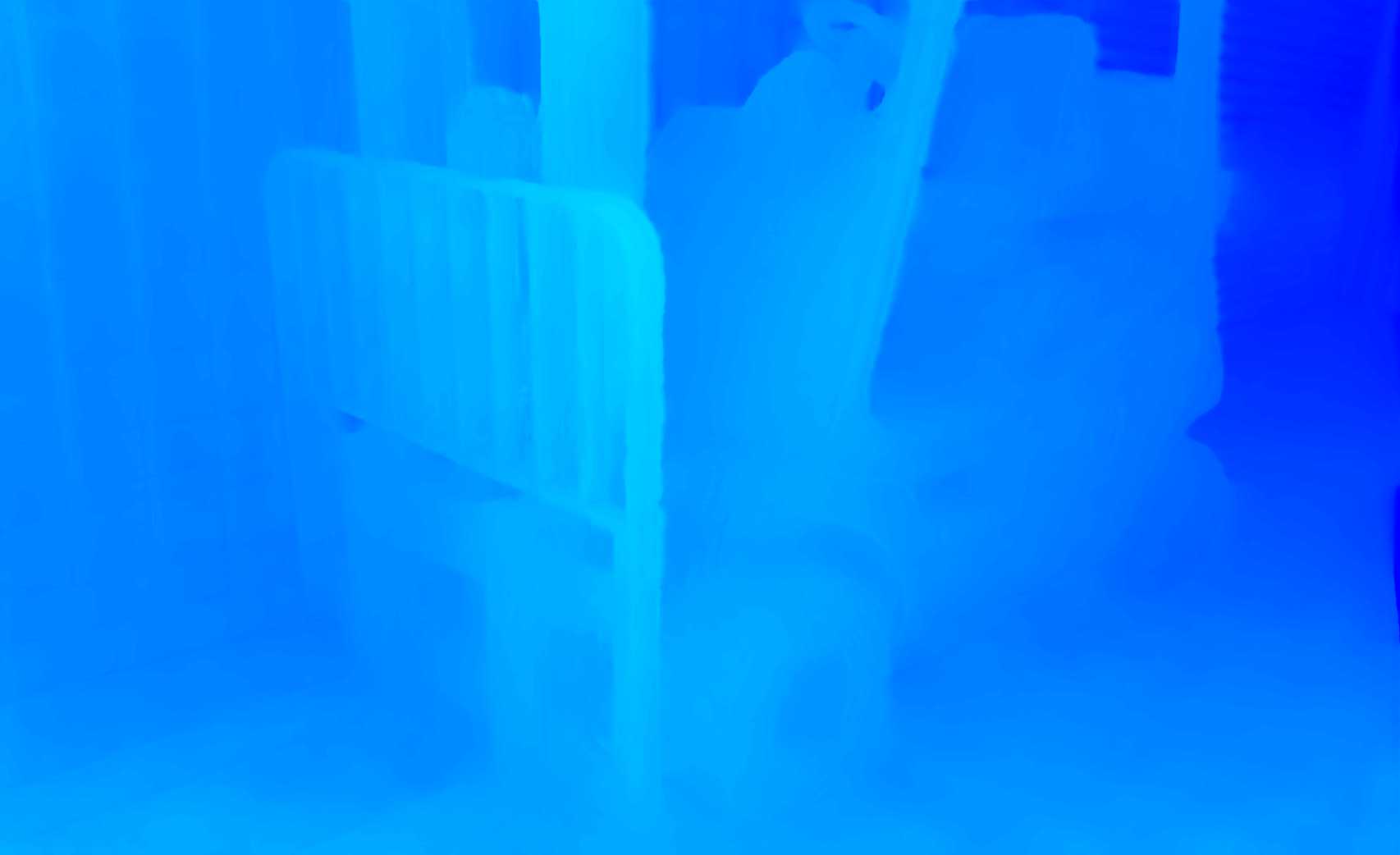}
\end{subfigure}

\begin{subfigure}{0.65\columnwidth}
  \centering
  \includegraphics[width=1\columnwidth, trim={0cm 0cm 0cm 0.8cm}, clip]{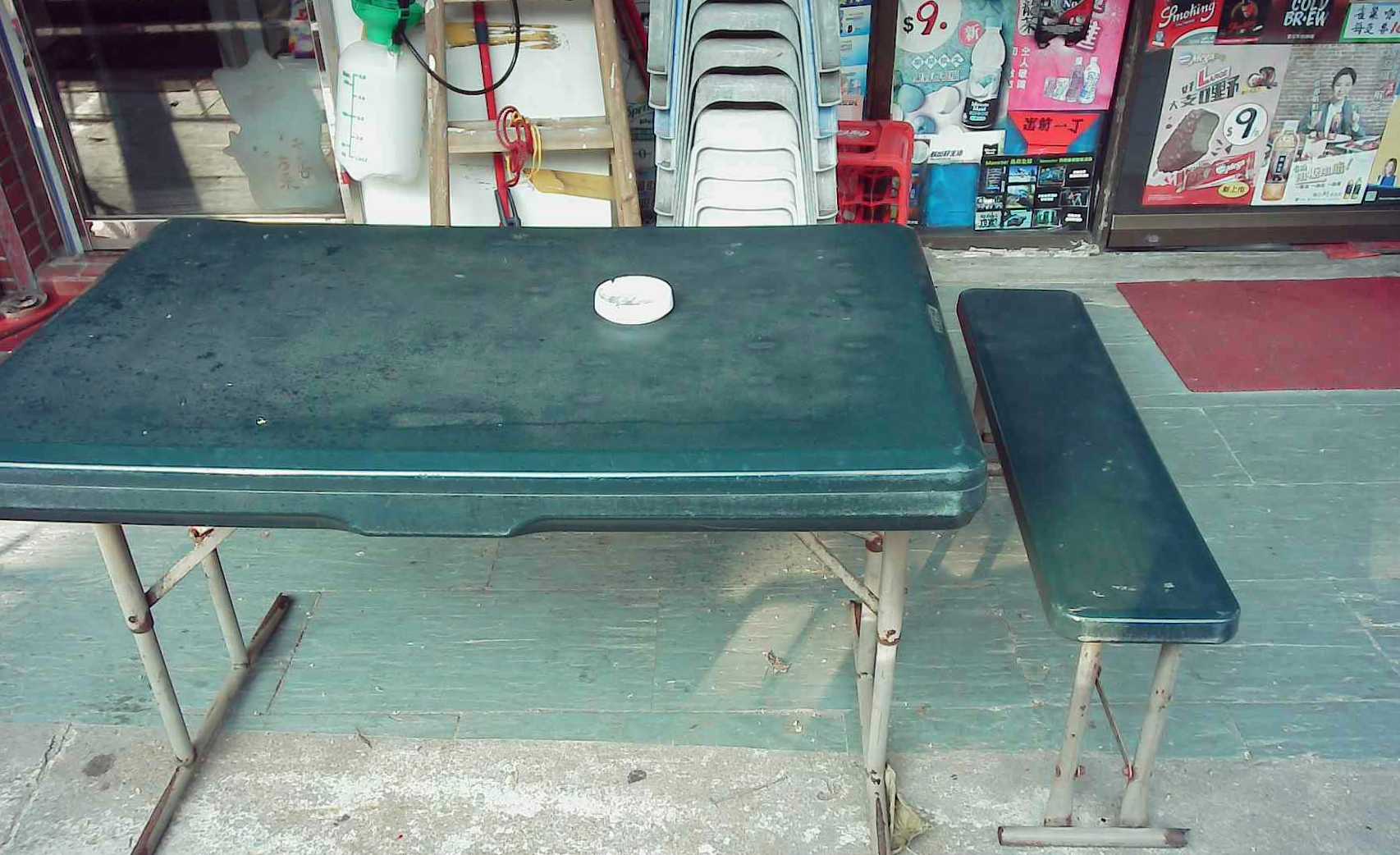}
\end{subfigure}
\begin{subfigure}{0.65\columnwidth}
  \centering
  \includegraphics[width=1\columnwidth, trim={0cm 0cm 0cm 0cm}, clip]{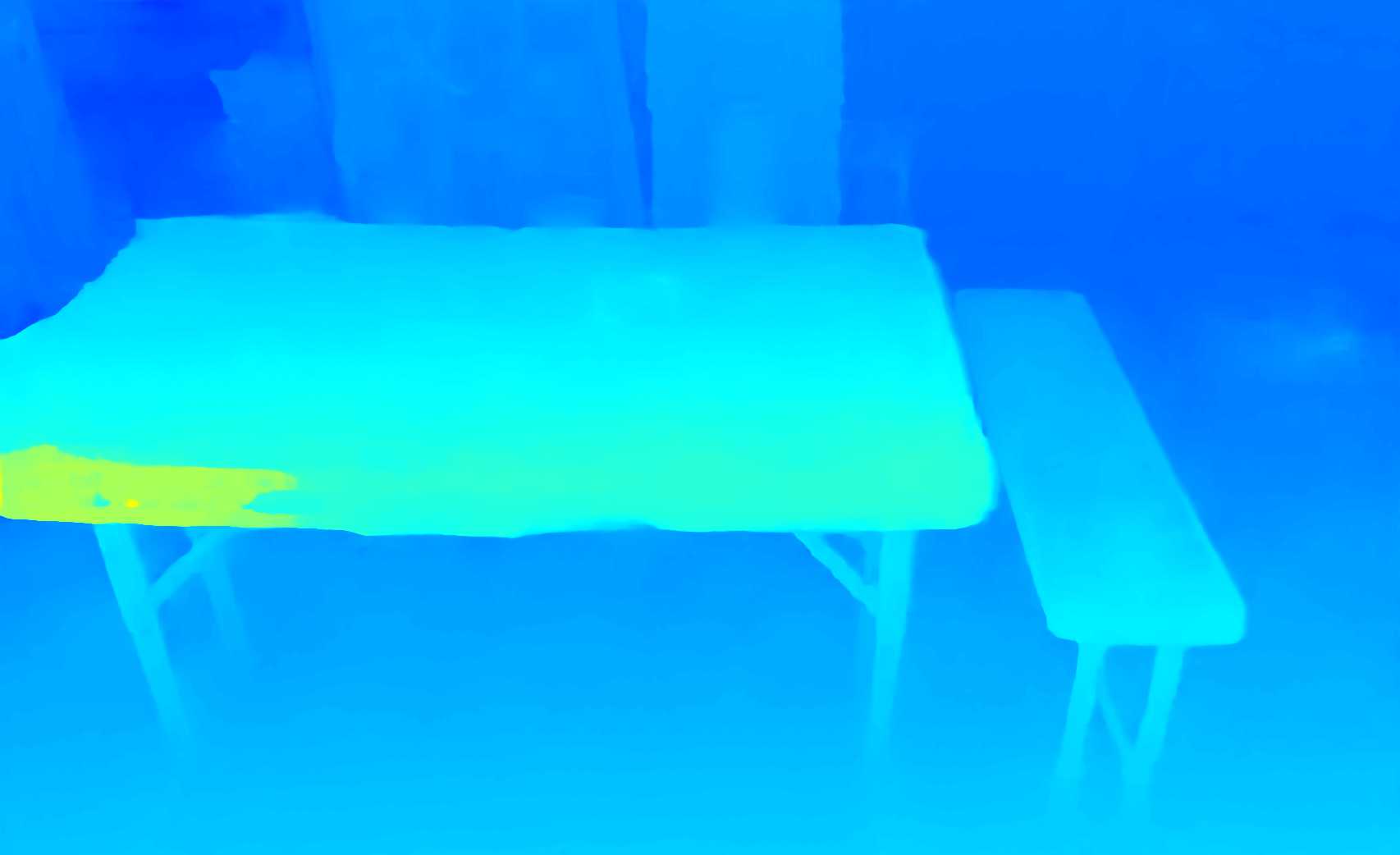}
\end{subfigure}
\begin{subfigure}{0.65\columnwidth}
  \centering
  \includegraphics[width=1\columnwidth, trim={0cm 0cm 0cm 0cm}, clip]{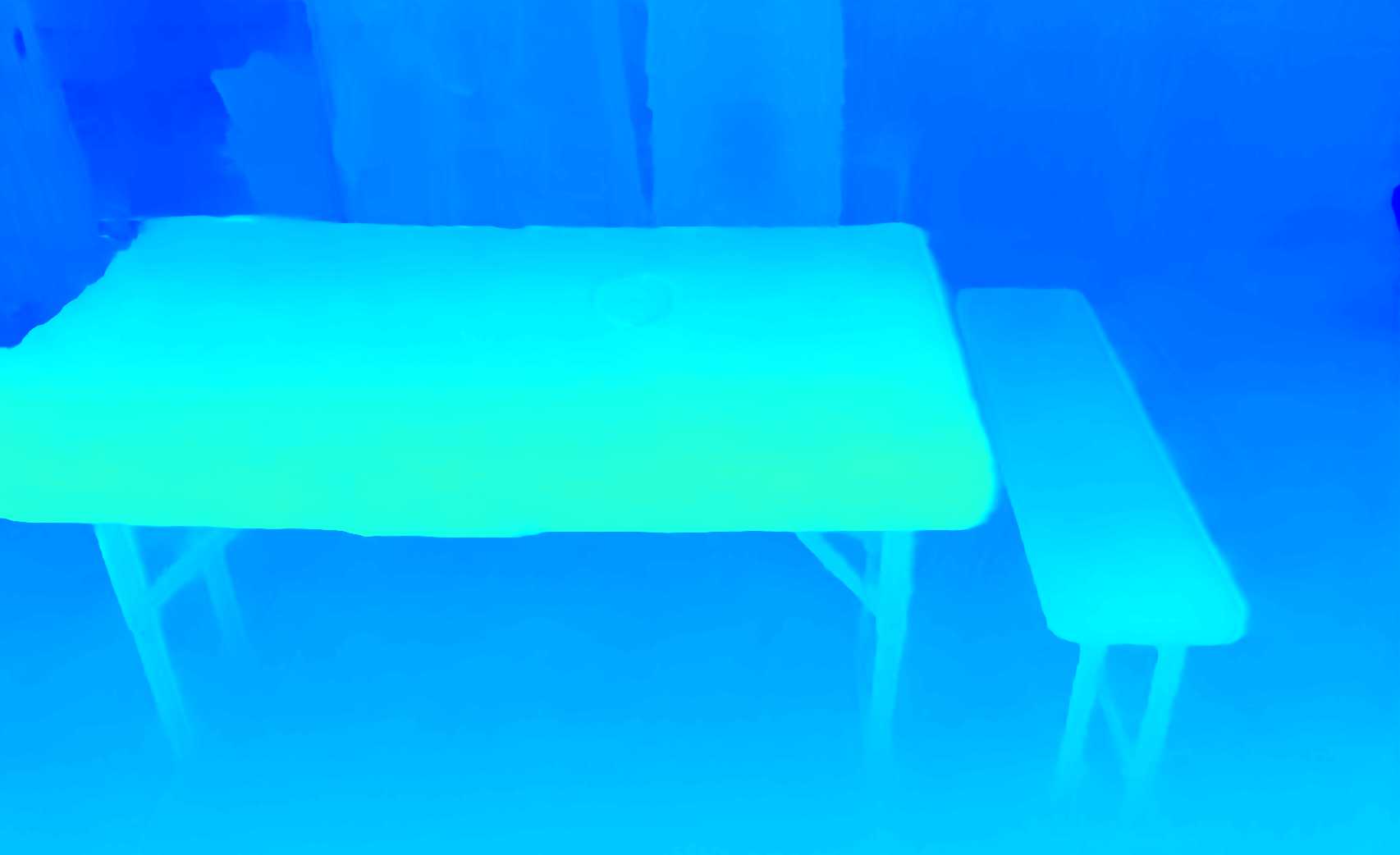}
\end{subfigure}

\begin{subfigure}{0.65\columnwidth}
  \centering
  \includegraphics[width=1\columnwidth, trim={0cm 0cm 0cm 0.8cm}, clip]{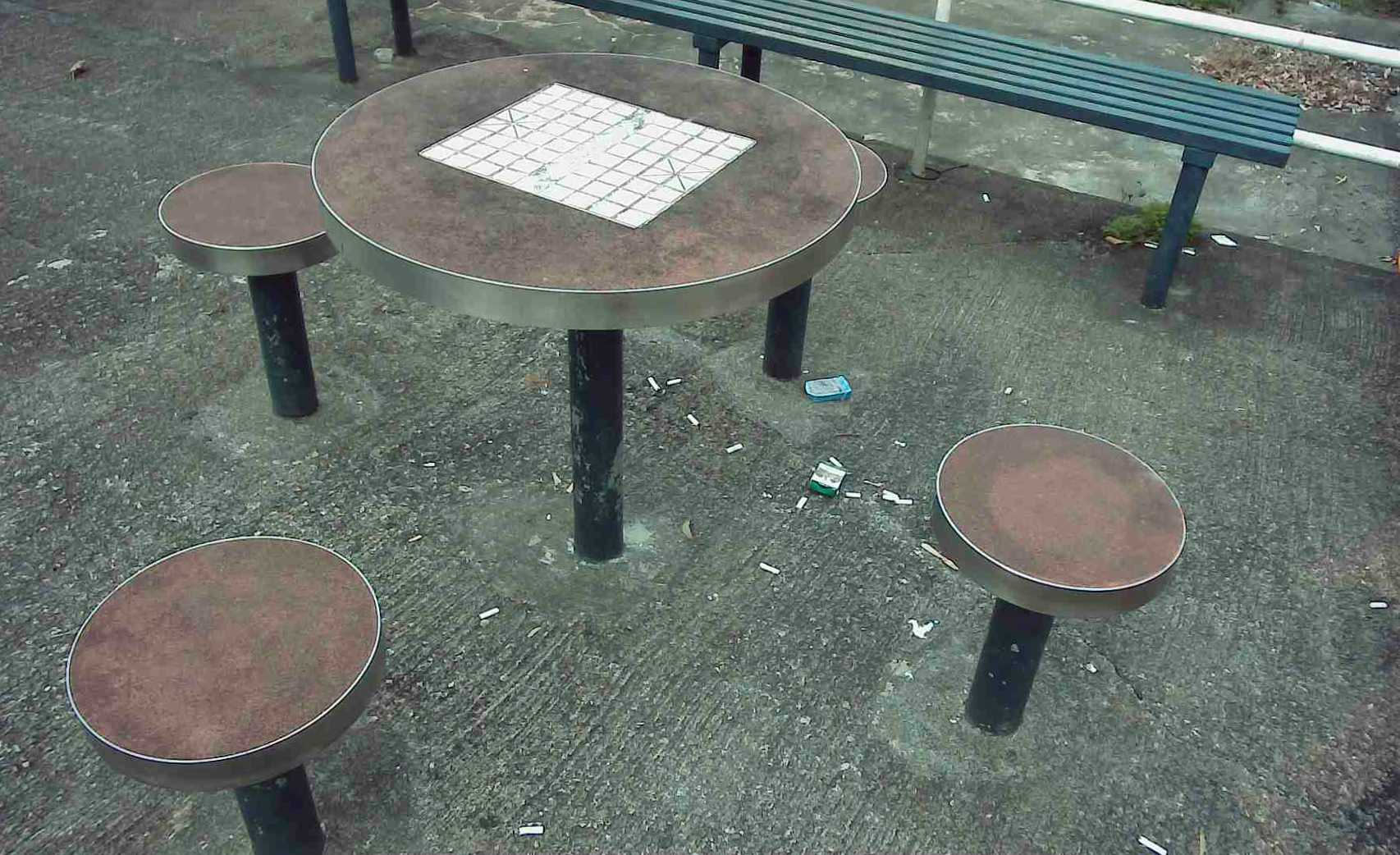}
\end{subfigure}
\begin{subfigure}{0.65\columnwidth}
  \centering
  \includegraphics[width=1\columnwidth, trim={0cm 0cm 0cm 0cm}, clip]{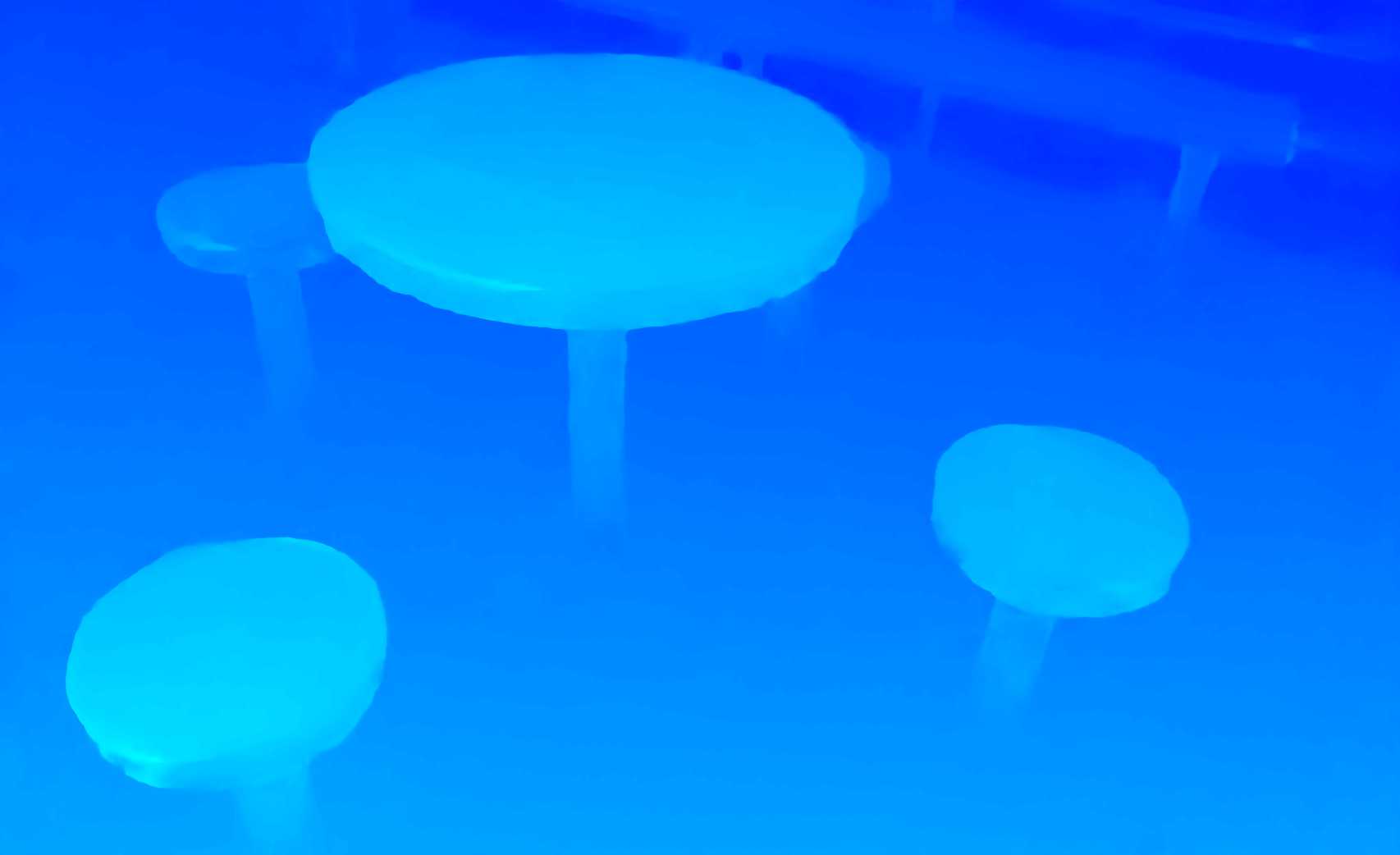}
\end{subfigure}
\begin{subfigure}{0.65\columnwidth}
  \centering
  \includegraphics[width=1\columnwidth, trim={0cm 0cm 0cm 0cm}, clip]{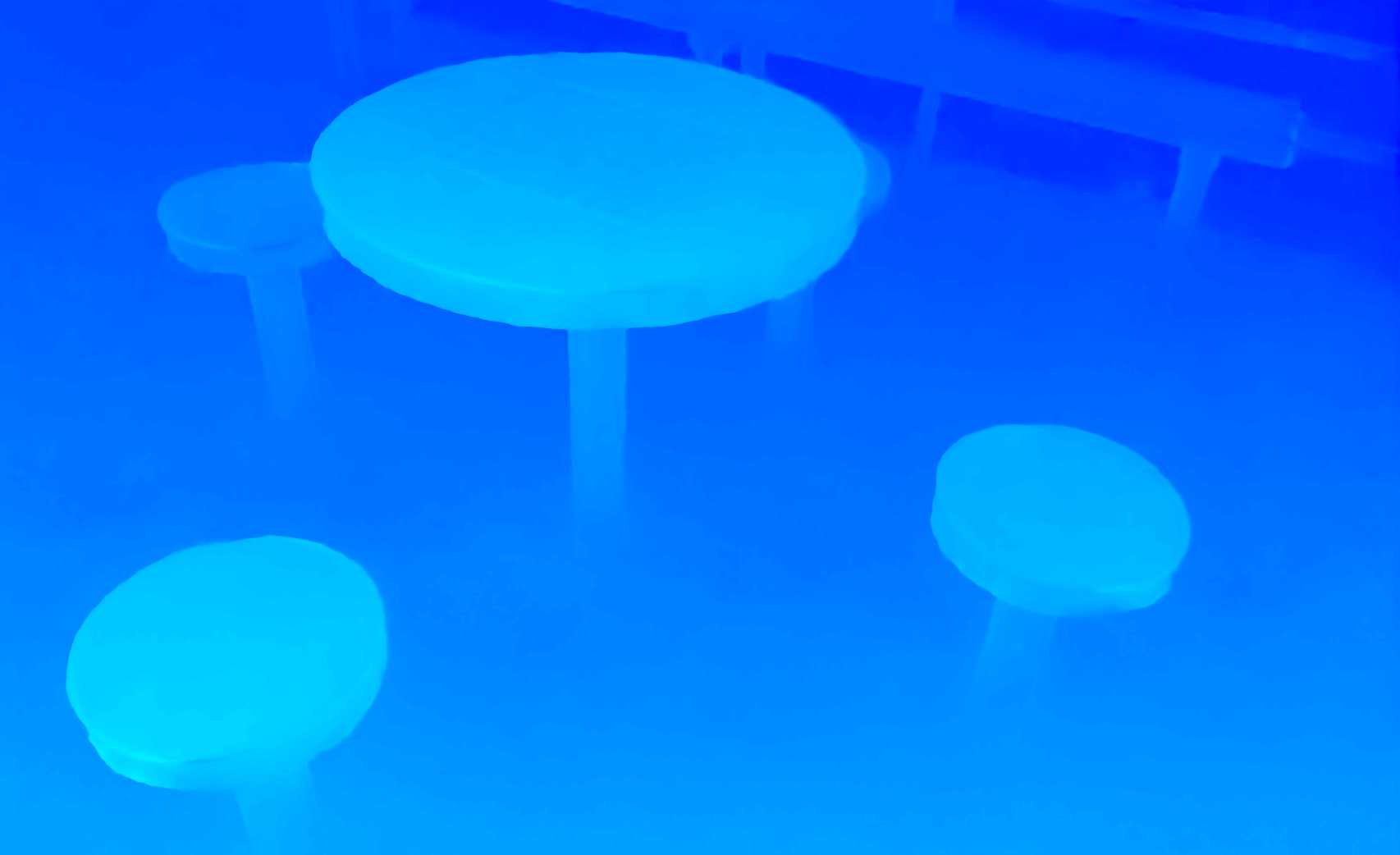}
\end{subfigure}

\begin{subfigure}{0.65\columnwidth}
  \centering
  \includegraphics[width=1\columnwidth, trim={0cm 0cm 0cm 0.8cm}, clip]{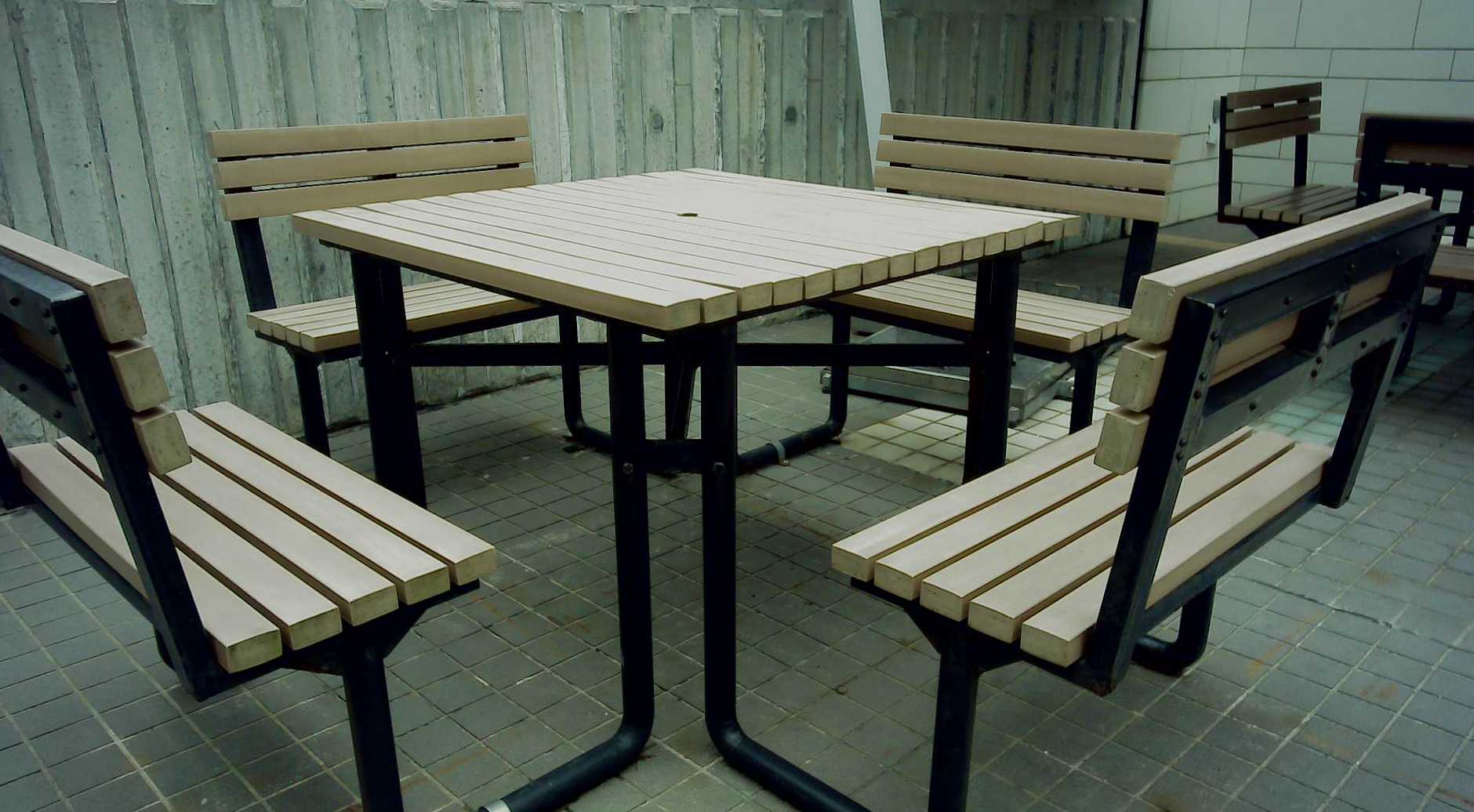}
  \caption*{Center view}
\end{subfigure}
\begin{subfigure}{0.65\columnwidth}
  \centering
  \includegraphics[width=1\columnwidth, trim={0cm 0cm 0cm 0cm}, clip]{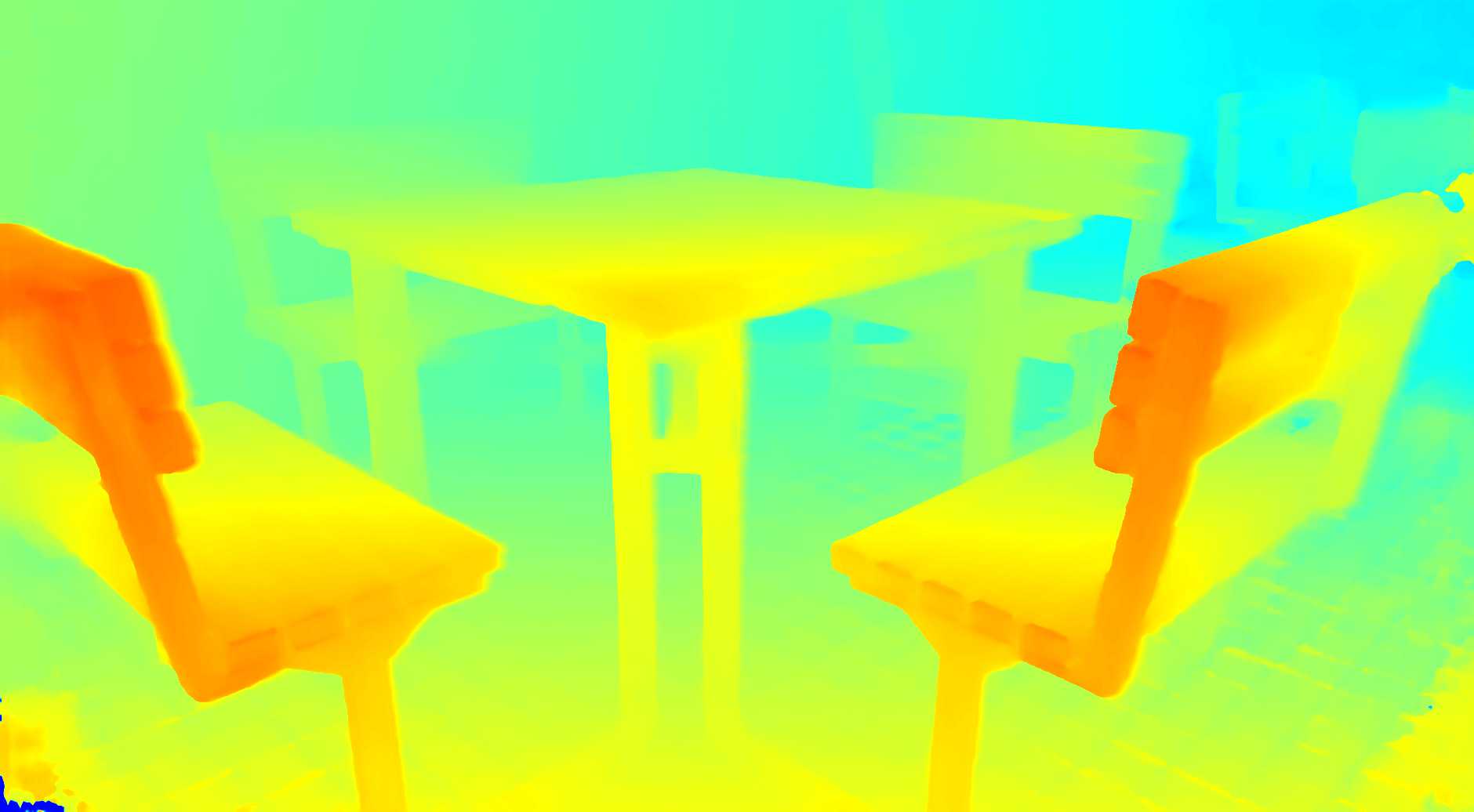}
  \caption*{Original model}
\end{subfigure}
\begin{subfigure}{0.65\columnwidth}
  \centering
  \includegraphics[width=1\columnwidth, trim={0cm 0cm 0cm 0cm}, clip]{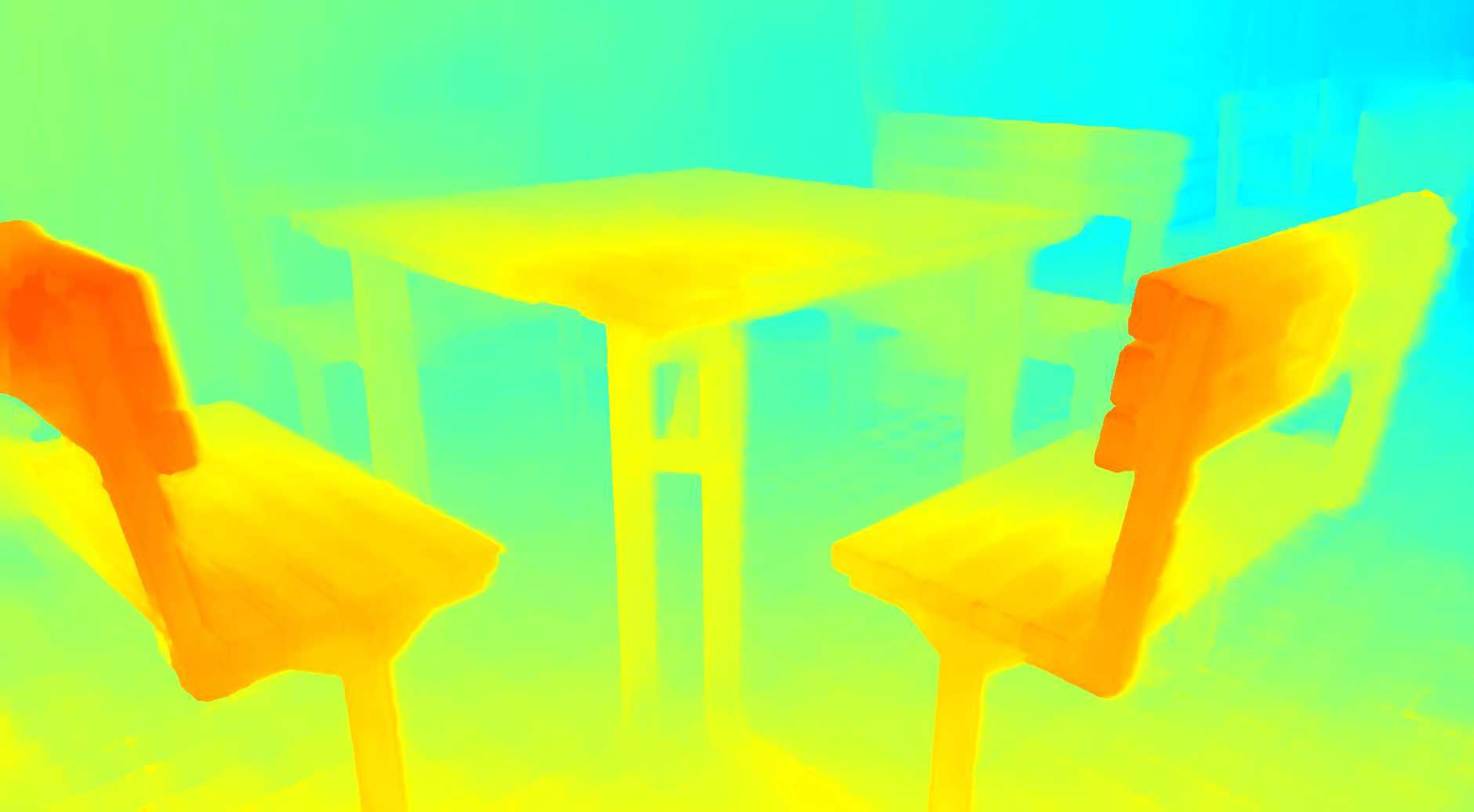}
  \caption*{Fine-tune}
\end{subfigure}
\caption{Online fine-tuning. The center-view color images, the output disparity maps of the original model, and the output after fine-tuning are shown.}
\label{fig:app-finetune}
\end{figure*}

One advantage of the self-supervised method is that online fine-tuning becomes feasible. Therefore, we fine-tune the network using the newly captured images when being employed in the real world. The comparison between the estimation after the fine-tuning and without the fine-tuning is displayed in Fig.~\ref{fig:app-finetune}, from which we can see the estimated disparity maps after fine-tuning are better. This demonstrates the practical usefulness of our self-supervised multiscopic framework.

\balance



{\small
\bibliographystyle{IEEEtranN}
\bibliography{ref}
}